\documentclass[12pt]{article}
\usepackage[utf8]{inputenc}
\usepackage{amsmath,amssymb}
\usepackage{graphicx,epsf, color}
\usepackage{enumerate}
\usepackage{natbib}
\usepackage{forest}
\usepackage{float}
\usepackage{url} 
\usepackage{times}
\usepackage[ruled,noend]{algorithm2e}
\usepackage{graphics}
\usepackage{booktabs}
\usepackage{xr}
\usepackage[colorlinks=true, linkcolor=blue, citecolor=blue, urlcolor = blue, filecolor=blue]{hyperref}
\usepackage{setspace}
\usepackage{caption}
\usepackage{subcaption}
\usepackage{lineno}

\newcommand{\blind}{1}

\usepackage[margin=1in]{geometry}

\catcode`\^ = 13 \def^#1{\sp{#1}{}}

\definecolor{Pink}{rgb}{1.0, 0.5, 0.5}
\definecolor{Maroon}{rgb}{0.8, 0.0, 0.0}

\def\boxit#1{\vbox{\hrule\hbox{\vrule\kern6pt\vbox{\kern6pt#1\kern6pt}\kern6pt\vrule}\hrule}}

\newcommand{\cre}{}

\newtheorem{theorem}{Theorem}[section]
\newtheorem{lemma}[theorem]{Lemma}

\newtheorem{corollary}[theorem]{Corollary}

\newenvironment{definition}[1][Definition]{\begin{trivlist}
\item[\hskip \labelsep {\bfseries #1}]}{\end{trivlist}}

\newcommand{\bX}{\mbox{\bf X}}

\newcommand{\bvare}{\mbox{\boldmath $\varepsilon$}}

\newcommand{\bbeta}{\mbox{\boldmath $\beta$}}
\newcommand{\beeta}{\mbox{\boldmath $\eta$}}

\newcommand{\bgamma}{\mbox{\boldmath $\gamma$}}

\newcommand{\bSigma}{\mbox{\boldmath $\Sigma$}}

\newcommand{\var}{\mathrm{var}}

\newcommand{\tr}{\mathrm{tr}}

\def\t{^\top}

\def\beqn{\begin{eqnarray}}
\def\eeqn{\end{eqnarray}}
\def\beqns{\begin{eqnarray*}}
\def\eeqns{\end{eqnarray*}}

\def\0{{\bf 0}}
\def\A{{\bf A}}

\def\C{{\bf C}}

\def\D{{\bf D}}

\def\I{{\bf I}}

\def\t{{\bf t}}

\def\T{{\bf T}}
\def\Z{{\bf Z}}

\def\u{{\bf u}}

\def\w{{\bf w}}
\def\X{{\bf X}}
\def\x{{\bf x}}
\def\T{{\bf T}}

\def\y{{\bf y}}
\def\Z{{\bf Z}}

\def\1{{\bf 1}}

\def\trans{^{\rm T}}

\def\t1trans{^{t+1\rm T}}

\begin{document}

\def\spacingset#1{\renewcommand{\baselinestretch}
{#1}\small\normalsize}
\spacingset{1}

\title{\bf Tree-Guided Rare Feature Selection and Logic Aggregation with Electronic Health Records Data}

\if0\blind
{
  \author{}\date{}
  \maketitle
} \fi

\if1\blind
{
  \author{Jianmin Chen$^1$, Robert H. Aseltine$^{23}$, Fei Wang$^4$, Kun Chen$^{13}$\thanks{Corresponding author; kun.chen@uconn.edu}\\    
$^1$\textit{Department of Statistics, University of Connecticut}\\
$^2$\textit{Center for Population Health, University of Connecticut Health Center (UCHC)}\\
$^3$\textit{Division of Behavioral Sciences and Community Health, UCHC}\\
$^4$\textit{Weill Cornell Medical College, Cornell University}
}
\maketitle
}
\fi

\begin{abstract}
  Statistical learning with a large number of rare binary features is commonly encountered in analyzing electronic health records (EHR) data, especially in the modeling of disease onset with prior medical diagnoses and procedures. 
  Dealing with the resulting highly sparse and large-scale binary feature matrix is notoriously challenging as conventional methods may suffer from a lack of power in testing and inconsistency in model fitting, while machine learning methods may suffer from the inability of producing interpretable results or clinically-meaningful risk factors. To improve EHR-based modeling and utilize the natural hierarchical structure of disease classification, we propose a tree-guided feature selection and logic aggregation approach for large-scale regression with rare binary features, in which dimension reduction is achieved through not only a sparsity pursuit but also an aggregation promoter with the logic operator of ``or''. We convert the combinatorial problem into a convex linearly-constrained regularized estimation, which enables scalable computation with theoretical guarantees. In a suicide risk study with EHR data, our approach is able to select and aggregate prior mental health diagnoses as guided by the diagnosis hierarchy of the International Classification of Diseases. By balancing the rarity and specificity of the EHR diagnosis records, our strategy improves both prediction and interpretation. We identify important higher-level categories and subcategories of mental health conditions and simultaneously determine the level of specificity needed for each of them in associating with suicide risk.

\end{abstract}

\noindent%

\doublespacing

\clearpage


\section{Introduction}

Statistical learning with a large number of rare binary features is commonly
encountered in modern applications. One prominent example arises in analyzing electronic
health records (EHR) data, where the main objective is to model/predict certain
disease onset with the medical history of diagnoses recorded as ICD (\emph{International Classification of Diseases}) codes.
The comprehensive disease classification and high specificity of the full-digit ICD codes
often lead to an extremely sparse and large-scale binary design matrix. Such rare feature problems are prevalent in a variety of fields. Some other examples include prediction of user ratings with absence/presence indicators of hundreds of keywords extracted from customer reviews~\citep{haque2018sentiment}, and studies of the gut-brain axis with absence/presence data of a large number of microbes~\citep{schloss2009introducing}. 

Rare features impose many challenges in statistical modeling with EHR data, making several conventional methods inadequate or even inapplicable. It is well known that naively using a large set of rare features in a regression could lead to inconsistency and divergence in model fitting and parameter estimation~\citep{albert1984existence,yan2021rare}. Even with the help of sparse regularization or other routine regularization techniques, the resulting models may still be lack of predictability and interpretability, as these methods tend to keep individual features with high prevalence and often fail to recognize ``weak'' features collectively. For many practitioners, feature deletion based on prevalence is a convenient approach to result in a reduced set of denser features \citep{forman2003extensive, huang2008similarity}. Its drawbacks are many: the selection of the prevalence threshold could be arbitrary, and it risks completely discarding potential information in many rare features. Another common approach is to perform marginal screening through univariate modeling and hypothesis testing~\citep{su2020machine, saeys2007review}. However, statistical test may be lack of power when dealing with rare features \citep{koehler1990chi,mukherjee2015hypothesis}. Moreover, due to the univariate nature of screening, a feature may be easily missed when its effect on the response is through interactions among multiple features  \citep{lucek1997neural}, which is quite likely for rare features. Dimension reduction methods or machine learning methods attempt to distill latent features or spaces (e.g., principal components or embeddings~\citep{levy2014neural,kozma2009binary}) from rare features. Unfortunately, these methods often facilitate prediction at a high cost of sacrificing model interpretability, which could be unbearable in medical studies with EHR data.

\begin{figure}[htp]
\centering
\includegraphics[width=0.5\textwidth]{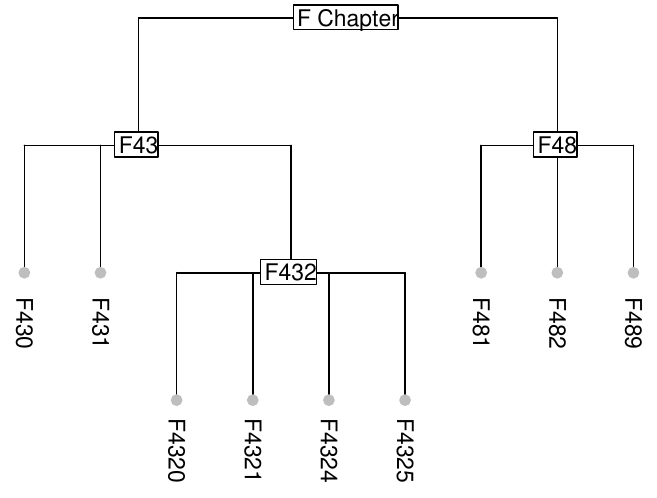}
\caption{Example: Hierarchical structure of ICD-10-CM codes.}
\label{fig:fig1}
\end{figure}

From a statistical perspective, rare features are unlikely to act alone, rather, they tend to impact the outcome in a collective and interactive way. Therefore, properly balancing the rarity and the specificity of the rare features holds the key in harvesting useful information from them. Fortunately, there often exists an inherited or derived 
hierarchical tree structure among the rare (binary) features, making it possible to perform interpretable \textit{rare feature aggregation}; see, e.g., the pioneer work by \citet{yan2021rare} and the references therein. Particularly, in EHR data analysis, ICD codes are organized in a hierarchical structure, where the higher-level codes represent more generic disease categories and the lower-level codes represent more specific diseases and conditions.
Figure~\ref{fig:fig1} displays an example with ICD-10-CM (\emph{International Classification of Diseases, Tenth Revision, Clinical Modification}) codes, where each binary feature, as a leaf node in the tree, indicates the absence/presence of a very specific disease condition. The F4320--F4325 codes indicate different types of ``adjustment disorder'', which can be collapsed into F432, their parent node; F432 is itself a sub-category of F43, ``reaction to severe stress, and adjustment disorders''. We remark that the ICD-10-CM system is very comprehensive with over 69,000 diagnosis codes; the example here is only a small subset of the ``F'' chapter code set, which is about mental, behavioral and neurodevelopmental disorders. There are other examples of tree-structured rare features: in microbiome studies, the evolutionary history of the taxa is charted through a taxonomic tree with up to seven taxonomic ranks: kingdom, phylum, class, order, family, genus, and species \citep{bien2021tree,li2020s}; in studies of medications, the active ingredients of drugs are classified in groups at five different levels according to the \emph{Anatomical Therapeutic Chemical} (ATC) classification system 
~\citep{cruciol2006prevalence, bjorkman2002drug}; in text mining and classification, a hierarchical structure of the keywords can be derived and utilized from pre-trained word embeddings such as GloVe and fastText~\citep{yan2021rare, naumov2021objective}.

With the available hierarchical structure, one naive approach of reducing feature rarity (and dimension) is to collapse all the features to a higher level in the hierarchy. Indeed, in EHR applications it is common to use only the first few digits of ICD codes for high-level disease categories, and in microbiome studies it is common to perform the analysis with aggregated data at the genus or order level. 
Although this simple approach maintains conformity to the tree structure~\citep{Cartwright2013,deschepper2019hospital}, it results in an indifferent loss of specificity across all feature categories/branches. 
Some recent works have considered supervised and data-driven rare feature aggregation. \citet{she2010sparse} considered clustered lasso without any side information. \citet{liu2010moreau} proposed tree structured group lasso and introduced an efficient algorithm; their method was deployed in \citet{jovanovic2016building} to build a predictive model for hospital readmission based on ICD codes, resulting in a more interpretable model with fewer high-level diagnoses and comparable prediction accuracy.
\citet{yan2021rare} proposed a feature aggregation method in high-dimensional regression through a tree-guided equi-sparsity regularization.
It has been applied on count data and further specialized to handle compositional data \citep{li2020s,liu2021surf,bien2021tree}. The framework is also extended in \citet{liu2022tree} to detect predictive interaction effects between rare features and covariates.
All the above approaches are designed for numerical features, where the features are aggregated through summation. As such, these methods do not encode potential nonlinear effects or high-order interactions among the rare features and may not be ideal for handling rare binary features from EHR data. 

Motivated by the summation-based feature aggregation methods \citep{yan2021rare} and the logic regression approach \citep{Ruczinski2003}, we propose a novel {\bf t}ree-guided feature {\bf s}election and {\bf l}ogic {\bf a}ggregation (TSLA) approach for large-scale regression with rare binary features, in which dimension reduction is achieved through not only a sparsity pursuit, but also an aggregation promoter with the logic operation of ``or'', as guided by any given hierarchical structure of the features. 
The utilization of the ``or'' operation is novel yet very natural for binary features, and this distinguishes our work from existing summation-based aggregation methods. Since the ``or'' operation involves interaction terms
between binary features, our approach automatically encodes nonlinear yet interpretable effects of the features on the response.
Furthermore, our TSLA approach can be regarded as a tailored logic regression \citep{Ruczinski2003,ruczinski2004exploring, kooperberg2001sequence, schwender2008identification}, which, in its general form, aims to learn Boolean combinations among predictors and is highly non-convex and computationally intensive.
In contrast, by formulating the task as a convex linearly-constrained regularized
estimation problem,
our approach enables efficient and scalable computation. Our logic aggregation approach is appealing in EHR applications; in predictive modeling with ICD diagnosis features, it holds promise to identify the important disease categories and subcategories and simultaneously determine the right levels of specificity in different disease categories.

The rest of the paper is organized as follows. In Section~\ref{sec:frame}, we introduce our logic aggregation framework for binary features in the context of EHR data. 
In Section~\ref{sec:alg}, a smoothing proximal gradient algorithm is introduced for efficient computation and some main theoretical results are discussed. 
Simulation studies are carried out in Section~\ref{sec:simu}, where both regression and classification scenarios are considered. 
The suicide risk modeling with EHR data is carried out in Section~\ref{sec:real:suicide}. Our results shed lights on distinguishing the important sub-categories of mental, behavioral, and neurodevelopmental disorders that are the most susceptible to suicide risk and thus are the most relevant to informing suicide prevention.
Some concluding remarks are provided in Section~\ref{sec:dis}.

\section{Logic Aggregation Framework with EHR Data}
\label{sec:frame}

Let $y \in \mathbb{R}$ be a response/outcome variable and {\cre $\x = (x_1,\ldots,x_{p_0})\trans \in \{0, 1\}^{p_0}$} be a $p_0$-dimensional predictor vector. Suppose $n$ copies of independent data on $(y,\x)$ are available, and denote $\y = (y_1,\ldots, y_n)\trans \in \mathbb{R}^{n}$ as the response vector and {\cre$\X_0 = (\x_1,\ldots, \x_n)\trans \in \{0, 1\}^{n\times p_0}$} the binary predictor matrix.

We are interested in the scenarios in EHR data analysis that the $p_0$ features are binary, sparse, high dimensional, and associated with a hierarchical grouping structure. See Figure~\ref{fig:fig1} for an example in EHR data analysis with ICD-10-CM codes. With the given feature hierarchy, there is a trade-off between data resolution/specificity and data sparsity. The main problem we are going to tackle is to efficiently identify the optimal levels of specificity in the feature hierarchy in supervised learning tasks. In the context of EHR data analysis, this amounts to selecting and identifying properly aggregated absence/presence patterns of disease conditions for best associating with or predicting a clinical outcome of interesting. 

Building upon summation-based aggregation methods and logic regression \citep{yan2021rare, Ruczinski2003}, the novelty of our approach is to reformulate the combinatorial problem of binary feature aggregation into a convex scalable learning problem through three major steps: (1) feature expansion, (2) reparameterization, and (3) equi-sparsity regularization.

\subsection{Tree-Guided Feature Expansion}\label{sec:method:expansion}
A simple yet critical observation is that under the linear regression setup, the aggregation of two binary features through the ``or'' operation amounts to requiring a special linear constraint of the parameters in a regression model that includes their two-way interaction term. Specifically, for two binary features, say, $x_1$ and $x_2$, we can write
\begin{align}
  x_{1}\ \lor \ x_{2}\  =\, & \ x_{1}\ + \ x_{2}\ -\ x_{1}\wedge x_{2}\notag\\
   =\, & \ x_{1}\ +\ x_{2}\ -\ x_{1}x_{2},
\end{align}
where the symbols ``$\lor$'' and ``$\wedge$" denote the logic ``or'' and ``and'' operators, respectively. Therefore, in a linear regression model with $x_1$, $x_2$ and their two-way interaction $x_1x_2$, the two predictors can be aggregated if and only if
\begin{align}
        \beta_1\ =\ \beta_2\ =\ -\beta_{12}, \label{eq:lc}
\end{align}
where $\beta_1$, $\beta_2$, and $\beta_{12}$ are the regression coefficients corresponding to $x_1$, $x_2$, and $x_1x_2$, respectively. More generally, the aggregation of multiple binary predictors corresponds to a parameter constraint that involves higher-order interaction terms. For $r$ ($r\geq 2$)  binary features, we can write
\begin{align}
     x_{1}\ \lor \ x_{2}\  \lor\ \cdots\ \lor\ x_{r}
     = & \sum_{i=1}^{r}\ x_{i}\ -\ \sum_{i\neq j}\ x_{i}x_{j}\ +\
     \sum_{i\neq j\neq k}x_{i}x_{j}x_{k}\ +\ \cdots\ +\ (-1)^{r-1}\prod_{i=1}^{r}x_{i},
\end{align}
and they can be aggregated if and only if their corresponding regression coefficients satisfy
\begin{align}
    \beta_i\ =\ -\beta_{ij}\ =\ \beta_{ijk}\ =\ \cdots\ =\ (-1)^{r-1}\beta_{12\cdots r}.
\end{align}

The above equivalency reveals that the binary feature aggregation through the ``or'' operation can be formulated as promoting certain linear constraints of the regression coefficients in a model with both main effects terms and appropriate interaction terms.

Consequently, the first step of our framework is to expand the feature set and the corresponding tree structure by including the interaction terms involved in the possible ``or'' operations in conformity to the feature hierarchy. On the one hand, the conventional main-effects regression is no longer adequate, as we now need a more comprehensive model with interactions. On the other hand, unlike the fully flexible logic regression, in our framework the possible aggregations are guided and restricted by the hierarchical tree structure. Besides, in practice many high-order interaction terms of the rare features could become zero vectors, so they can be directly dropped from the model. In EHR data, the simultaneous presence of a set of rare and related medical conditions is certainly possible but also is even more rare. So the potential computational burden associated with this tree-guided feature expansion is more bearable than it appears.

To proceed, we introduce some standard terminologies of the tree structure. For a tree $\T$, let $I(\T)$ and $L(\T)$ denote the set of internal nodes and the set of leaf nodes, respectively. For a node $u$, denote $C(u)$, $A(u)$, and $D(u)$ as the set of child nodes of $u$, the set of ancestor nodes of $u$, and the set of descendant nodes of $u$, respectively.

\begin{figure}[htp]
     \centering
     \begin{subfigure}[b]{0.49\textwidth}
         \centering
         \includegraphics[width=\textwidth]{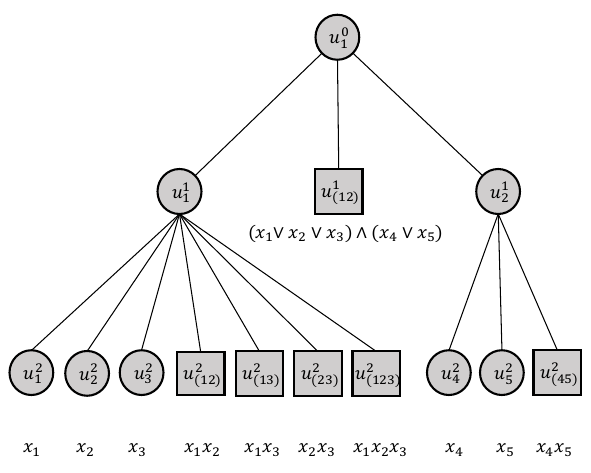}
         \caption{Feature expansion}
     \end{subfigure}
     \hfill
     \begin{subfigure}[b]{0.49\textwidth}
         \centering
         \includegraphics[width=\textwidth]{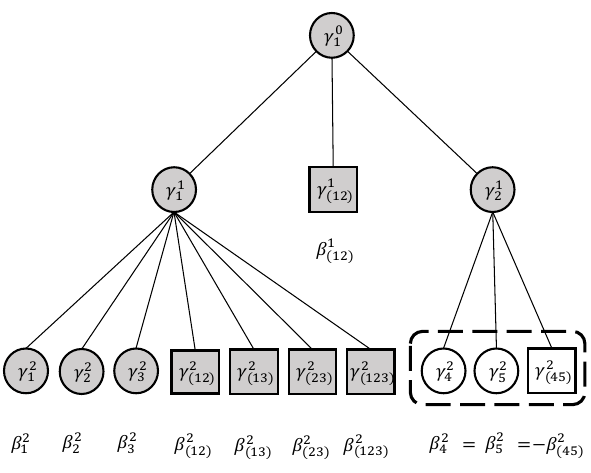}
         \caption{Reparameterization}
       \end{subfigure}
  \caption{Tree-guided feature expansion and reparameterization. Circles and squares indicate native and derived nodes, respectively. For example, on the left panel, node $u_{(12)}^1$ is derived from $u_1^1$ and $u_2^1$ from their two-way interaction. On the right panel, solid and blank nodes indicate non-zero and zero coefficients, respectively.}
  \label{fig:fig2}
\end{figure}

In our problem, the original $p_0$ nodes and nodes generated for their interactions are referred as ``native'' nodes and ``derived'' nodes, respectively. We denote each derived node in the expanded tree as $u_{(\mathcal{S})}^h$, where the superscript $h$ indicates the depth of the node on the tree, and $\mathcal{S}$, the aggregation set, gives the corresponding nodes at the same depth that are involved in the construction of the node. Figure~\ref{fig:fig2}(a) provides an illustration of the feature expansion. The original tree is with $p_0=5$ leaf nodes and of depth $h=2$. In this example, $u_{j}^2$, $j = 1,\ldots, 5$, are the native leaf nodes corresponding to $x_j$, $j = 1,\ldots, 5$.
The node $u_{(12)}^2$ is a derived node from $\mathcal{S} = \{u_1^2, u_2^2\}$, which corresponds to the two-way interaction of $x_1$ and $x_2$. Similarly, the node $u_{(12)}^{1}$ is a derived node from $\mathcal{S} = \{u_{1}^{1}, u_{2}^{1}\}$, which corresponds to the interaction of $(x_1 \lor x_2 \lor x_3)$ and $(x_4 \lor x_5)$. Under our framework, potential aggregation only happens to the set of features corresponding to the child nodes $C(u)$ of each internal node $u$.

Through the aforedescribed process, the original design matrix {\cre $\X_0\in \{0, 1\}^{n\times p_0}$} is expanded to be the new design matrix {\cre$\X \in \{0, 1\}^{n\times p}$}, which consists of both the original binary predictors and the generated (non-zero) interaction columns. This enables us to pursue tree-guided binary feature aggregation under the linear regression model
\begin{align}
    \y =\, \mu\1\ +\ \X\bbeta\ +\ \bvare,
    \label{eq:lr}
\end{align}
where $\mu \in \mathbb{R}$ is the intercept, $\1 \in \mathbb{R}^{n}$, $\bbeta \in \mathbb{R}^{p}$ is the regression coefficient vector, and $\bvare \in \mathbb{R}^{n}$ is the random error vector. Correspondingly, let $\T$ be the expanded tree of the feature grouping structure; it is with 
in total $p$ leaf nodes (among which $p_0$ nodes are native) and is of depth $h$. 

\subsection{Tree-Guided Reparameterization}
With the expanded linear interaction model in \eqref{eq:lr}, the main problem now is how to parameterize it to facilitate feature aggregation through the ``or'' operation, i.e., to promote the linear constraints of the form in \eqref{eq:lc}. Motivated by \citet{yan2021rare} and \citet{li2020s}, we propose a tree-guided reparameterization of the model in \eqref{eq:lr}.

An intermediate coefficient $\gamma_u$ is assigned to each node $u$ of the expanded tree $\T$ with $p_0$ native leaf nodes and of depth $h$.
For each native leaf node $u_{j}^{l}$, for $l = \{0, \ldots, h\}$, its corresponding regression coefficient, denoted as $\beta_{j}^{l}$, is parameterized as
\begin{align}
    \beta_{j}^{l}\ =\, \sum_{u\in A(u_{j}^{l})\cup\{u_{j}^{l}\}}\gamma_u.\label{eq:para1}
\end{align}
For each derived leaf node, $u_{(\mathcal{S})}^{l}$, for $l = \{0, \ldots, h\}$, its parameterization is according to the sign of the interaction terms from the potential ``or'' operations,
\begin{align}
    \beta_{(\mathcal{S})}^{l}\ =\ (-1)^{|\mathcal{S}|-1}\times \sum_{u\in A(u_{(\mathcal{S})}^{l})\cup\{u_{(\mathcal{S})}^{l}\}}\gamma_u,\label{eq:para2}
\end{align}
where $|\mathcal{S}|$ is the cardinality of the aggregation set $\mathcal{S}$. 

Through this reparameterization, any potential ``or'' operation conforming to the tree structure can be expressed as a set of sparsity constraints on the $\gamma_u$ coefficients. As shown in Figure~\ref{fig:fig2}(b)
we have that $\beta_4^2 = \  \gamma_{4}^2 + \gamma_{2}^1 + \gamma_{1}^{0}$, $\beta_5^2 = \  \gamma_{5}^2 + \gamma_{2}^1 + \gamma_{1}^{0}$, and $\beta_{(45)}^2 =  -\gamma_{(45)}^2 - \gamma_{2}^1 - \gamma_{1}^{0}$. Therefore, the aggregation of $x_4$ and $x_5$ into $x_4 \lor x_5$ is equivalent to $\beta_4^2=\beta_5^2=-\beta_{(45)}^2$ or $\gamma_{4}^2=\gamma_{5}^2=\gamma_{(45)}^2=0$.

With any given tree $\T$, denote the number of nodes in $\T$ as $|\T|$, let $\bgamma = (\gamma_u) \in \mathbb{R}^{|\T|}$ be the vector collecting all the intermediate coefficients on the tree. The transformation from $\bbeta$ to $\bgamma$ is linear and deterministic, which can be expressed as $\bbeta=\A\bgamma$ with $\A$ of dimension $p\times |\T|$ derived from the expanded tree structure as \eqref{eq:para1} and \eqref{eq:para2}. 

\subsection{Tree-Guided Regularization}
\label{sec:frame:reg}

From the previous discussion, aggregating a set of features under the same ancestor node $u$ is equivalent to setting all intermediate coefficients $\gamma_u$ in $D(u)$ to be 0.
As a consequence, equi-sparsity in $\bbeta$ can be achieved by inducing zero-sparsity in $\bgamma$ with structured regularization.
With a given tree $\T$, we construct $\mathcal{G}$ as a set of non-overlapped node groups, where
\begin{align}
    \mathcal{G}\ =\ \{u_1^0\}\cup\{C(u),\ u\in I(\T)\}.
\end{align}
Then the groups in $\mathcal{G}$ form a partition of nodes in tree $\T$ and $D(u)$ can be expressed as the union of node groups in $\mathcal{G}$ for all $\u\in I(\T)$.
We consider the \emph{Child-$\ell_2$} penalty in $\bgamma$ based on the set $\mathcal{G}$, which is constructed as
\begin{align}
    P_T(\bgamma)=\sum_{g\in \mathcal{G}}w_g\|\bgamma_g\|, \qquad     w_g \ =\ \sqrt{\frac{p_g}{2^k-1}},\label{eq:weight}
\end{align}
where $\bgamma_g\in \mathbb{R}^{p_g}$ is a sub-vector of $\bgamma$ corresponding to nodes in group $g$, $p_g$ is the size of $g$, and $k$ is the maximum number of child nodes of the original tree before expansion.

The \emph{Child-$\ell_2$} penalty is in the form of a weighted group-lasso penalty with no overlapping between groups, where the root node is also regularized as a singleton. 
The weight parameter $w_g$ for each group are selected
to control relative importance between groups, depending on the specific
assumptions of the true tree structure.
The group weights in \eqref{eq:weight} are suggested by
our theoretical analysis in Section~\ref{sec::app-alm} of the Supplement and are used in all our numerical studies.

Finally, our proposed tree-guided feature selection and logic aggregation (TSLA) approach is expressed as the following optimization problem:
\begin{equation}
    \small
    \label{eq:optimprob}
    \begin{aligned}
    (\hat\mu,\ \hat\bbeta,\ \hat\bgamma)=\, &\arg\min_{\mu, \bbeta, \bgamma}\{\frac{1}{2n}\|\y-\mu\1-\X\bbeta\|^2+
    \lambda \Omega(\bbeta, \bgamma; \alpha)
    \},
    \ \mbox{s.t}\ \bbeta\ =\ \A \bgamma,
    \end{aligned}
\end{equation}
where
\begin{align}
\Omega(\bbeta, \bgamma; \alpha) = (1-\alpha)\sum_{j=1}^p\tilde{w}_j|\beta_j|+
    \alpha P_T(\bgamma), \label{eq:penalty}
\end{align}
with $\lambda \geq 0$ and $\alpha\in[0,\ 1]$. 
Here, an additional $\ell_1$ penalty term of $\bbeta$ is included to enable feature selection.
In our study, we use $\tilde{w}_j = \|\x_j\|/\sqrt{n}$ to take into account the scale of each feature. 
The tuning parameter $\alpha$ controls the balance between selection and aggregation while $\lambda$ controls the overall regularization level.

\section{Scalable Computation and Theoretical Guarantee}
\label{sec:alg}

To solve the optimization problem in \eqref{eq:optimprob}, we first rewrite it as an unconstrained problem of $\bgamma$ and $\mu$ by replacing $\bbeta$ with $\A\bgamma$, which gives
\begin{align}
    (\hat\mu,\ \hat\bgamma)=\, &\arg\min_{\mu, \bgamma}\{\frac{1}{2n}\|\y-\mu\1-\X\A\bgamma\|^2+
    \lambda(1-\alpha)\|\D\A\bgamma\|_1+
    \lambda\alpha P_T(\bgamma)\},
    \label{eq:optimprob2}
\end{align}
where $\D\in \mathbb{R}^{p\times p}$ is a diagonal matrix with the $j^{th}$ diagonal element as $\tilde{w}_j$. 
This is a convex problem, but the
penalty function is non-smooth. 
We adopt a \emph{Smoothing Proximal Gradient} (SPG) algorithm \citep{chen2012smoothing} for the optimization. The algorithm can be readily extended to the settings of logistic regression for binary response, by replacing the $\ell_2$ loss with the negative log-likelihood function. For large-scale data, the \emph{fast iterative shrinkage-thresholding algorithm} (FISTA)~\citep{beck2009fast} is applied in SPG for optimization. The FISTA combines the gradient descent method with the Nesterov acceleration technique~\citep{nesterov1983method}, which leads to faster convergence than standard gradient-based methods. The optimal combination of the tuning parameters $\alpha$ and $\lambda$ can be determined via cross validation. All the details are provided in Section~\ref{sec::app-alm} of the Supplement. 

After getting the estimator $\hat\bgamma$, the feature aggregation structure can be derived by selecting groups with nonzero coefficients. 
Due to the ``over-selection'' property of the convex group lasso penalty~\citep{wei2010consistent}, a
thresholding operation is preferred in practice to produce a more sparse group structure. 
Denote $\hat{m}_g=w_g\|\hat{\bgamma}_g\|$; we select group $g$ if
\begin{align}
  {\cre  \frac{\hat m_g}{\sum_{g^{'}\in \mathcal{G}} \hat m_{g^{'}}}> \frac{1}{|\mathcal{G}|},}
\label{eq:grouprule}
\end{align}
where $w_g$ is the weight for group $g$ specified in \eqref{eq:weight} and $|\mathcal{G}|$ is the cardinality of $\mathcal{G}$. 

We have studied the theoretical properties of TSLA under a general high-dimensional model setup, in which $p$ can potentially grow with $n$. We show that under the assumptions of Gaussian errors and \emph{k-ary} tree structure (a rooted tree where every internal node has at most $k$ child nodes), 
the following \textit{non-asymptotic prediction error bound} holds for any $\alpha \in [0,1]$ with probability at least $1-1/n-2/(pn)$,
\begin{align}
    \frac{1}{n}\|\X\hat\bbeta - \X\bbeta^*\| ^2 \preceq \sigma \sqrt{2^k-1} \sqrt{\frac{\log (p \vee n)}{n}}  
    \Omega(\bbeta^*,\bgamma^*; \alpha),
    \label{eq:bound}
\end{align}
where $p \vee n = \max(p,n)$, $\sigma$ is the error standard deviation, $\Omega(\bbeta^*,\bgamma^*; \alpha)$ is the penalty function in \eqref{eq:penalty} evaluated at the true coefficient vector $\bbeta^*$, and $\bgamma^*=\mathop{\arg\min}\limits_{\gamma: \A\gamma=\beta^*}\{P_T(\bgamma)\}$. 
The symbol $\preceq$ means that the inequality holds up to a multiplicative constant that is independent of $\sigma$, $p$, $n$, and $k$. The details are shown in Theorem~\ref{thm:1} in Section~\ref{sec:thm} of the Supplement.

Our analysis reveals that the dimension of the logic aggregation problem depends on the tree structure of the feature hierarchy mainly through the number of original features ($p_0$), the depth of the tree ($h$), and the maximum number of child nodes ($k$). This is not surprising, as the interactions of each set of child nodes are needed in order to pursue their potential logic aggregation. Based on the property of a \emph{k-ary} tree, it always holds that  $p_0\leq k^h$, and $p < k^h2^k/(k-1)$. In our EHR applications, we typically have a ``slim'' tree, where the number of child nodes of each internal node, i.e., the number of sub-categories of a particular disease condition, is limited and small, so that it is reasonable to consider $k$ as a constant.  

The term $\Omega(\bbeta^*,\bgamma^*; \alpha)$ measures the minimal penalty function evaluated at the truth, which is a natural measure of \textit{the complexity of the true model} under both zero-sparsity and equi-sparsity. Moreover, according to Lemma~1 in \citet{yan2021rare}, there is a unique coarsest ``true aggregation set'' corresponding to the true regression coefficient vector $\bbeta^*$. By constructing this coarsest aggregation set, we can show that  $\Omega(\bbeta^*,\bgamma^*;\alpha) \leq \|\bbeta^*\|_1$ for any $\alpha \in [0,1]$, which results in a simplified high-probability error bound: {\cre$\|\X\hat\bbeta-\X\bbeta^*\|^2/n\preceq
  \sigma\sqrt{2^k-1}\sqrt{\log (p\vee n)/n}\|\bbeta^*\|_1$. This shows that the prediction error bound of TSLA is at least comparable to that of the lasso \citep{buhlmann2011statistics} when $k$ is small or considered as a constant. See Corollary~\ref{coro:1} in Section~\ref{sec:thm} of the Supplement for details.}

It is possible to improve the error bound through taking into account the rarity of the features and the equi-sparsity of the true model; see details in Section~\ref{sec:thm} of the Supplement. Roughly, our analysis shows that if the original features are assumed to be independent and $P(x_1 \lor x_2 \lor \cdots \lor x_{p_0} = 1) \leq a/k$, where $0< a < k/3$ is a  smaller positive constant, then
the multiplicative factor $\sqrt{2^k-1}$ in \eqref{eq:bound} can be replaced with $\sqrt{e^a}$. This reveals that TSLA can still perform well when $k$ is large as long as the original features are sufficiently sparse; the key is to bound the magnitude of the high-order interaction terms that are needed for pursuing the ``or'' operation. Moreover, if we explore the equi-sparsity pattern by assuming that the original features can be aggregated to the depth $h_0$ along the tree in the true model, then it holds that $\Omega(\bbeta^*,\bgamma^*; \alpha) \leq \|\bbeta^*\|_1-\alpha(1-f(h_0; p_0,h))\|\bbeta_2^*\|_1$, where $\bbeta_2^*$ collects the coefficients with equi-sparsity in $\bbeta^*$ and $0<f(h_0; p_0,h) \leq 1$ is an increasing function of $h_0$. In particular, $f(0; p_0,h) = 1/p$ and $f(h; p_0,h) = 1$. Therefore, the potential gain of TSLA over the lasso in prediction indeed depends on how much the features can be aggregated along the tree. These results are discussed in Corollary~\ref{coro:2} and Corollary~\ref{coro:3} in Section~\ref{sec:thm} of the Supplement.

\section{Simulation}
\label{sec:simu}

We compare the proposed TSLA approach to several competing methods under both regression and classification settings through simulation. The competing methods include 
the elastic-net penalized linear or logistic regression with the original features (Enet), the elastic-net penalized linear or logistic regression with both the original features and their two-way interactions (Enet2), and the tree-guided rare feature selection method by \citet{yan2021rare} through the ``sum'' operation (RFS-Sum). As a benchmark, the oracle linear or logistic regression using the true aggregated features (ORE) is also included.
For any regularization method, we use 5-fold cross validation for tuning parameter selection. 

\subsection{Simulation under Regression Settings}
\label{sec:simu-reg}

Three tree structures of different complexities and dimensions are considered, as shown in Figures~\ref{fig:fig1}--\ref{fig:fig3} in Section~\ref{sec::app-simu-tree} of the Supplement. 
For Tree 1, 2, and 3, the number of original leaf nodes, $p_0$, is 15, 42, and 43, respectively, and the number of maximum child nodes, $k$, is 4, 11, and 10, respectively. In each case, the original design matrix
{\cre$\X_0\in \{0, 1\}^{n\times p_0}$} is generated as a sparse binary matrix with independent Bernoulli$(0.1)$ entries. Through the feature expansion step in Section \ref{sec:method:expansion}, the new design matrix {\cre$\X\in \{0, 1\}^{n\times p}$} is obtained, and its dimension $p$, i.e., the number of non-null leaf nodes of the expanded tree, can reach around 40, 540, and 400 for Tree~1, 2, and 3, respectively.

Under the regression settings, the continuous response variable $y$ follows the model
\begin{align*}
    y\ =& \ 2+3(x_1\ \lor\ x_2\ \lor\ x_3\ \lor\ x_4)-5x_9+
          1.5\{(x_{10}\ \lor\ x_{11}\ \lor\ x_{12})\lor\ x_{13}\}+\varepsilon\\
          =& \ 2+3(x_1\ \lor\ x_2\ \lor\ x_3\ \lor\ x_4)-5x_9+
    1.5(x_{10}\ \lor\ x_{11}\ \lor\ x_{12}\ \lor\ x_{13})+\varepsilon,
\end{align*}
where $\varepsilon\sim N(0,\sigma^2)$ with $\sigma^2$ determined to control the signal to noise ratio (SNR) defined as $\var(\mu\1+\X\bbeta)/\sigma^2$. {\cre We then generate $n$ independent samples of $y$ according to the above model using the generated design matrix $\X_0$ and independently sampled error terms.} 
We consider three scenarios,
\begin{itemize}
\item Case 1 (low-dimensional, low SNR): Tree 1 with $n=200$ and $\mbox{SNR} = 0.5$;
\item Case 2 (low-dimensional, high SNR): Tree 1 with $n=200$ and $\mbox{SNR} = 2$;
\item Case 3 (high-dimensional): Tree 2 with $n=200$ and $\mbox{SNR} = 2$.
\end{itemize}

For each method, the prediction performance is measured by out-of-sample mean squared error (MSE) evaluated on independently generated testing data of size 200. The feature aggregation accuracy is measured by false positive rate (FPR) and false negative rate (FNR) in selecting the correct feature grouping pattern. For each setting, the simulation is repeated for 100 times.

Table~\ref{tab:tab1} reports the means and standard errors of the MSE, FNR, and FPR; we have also conducted paired $t$-tests to compare these performance metrics between TSLA and the other methods, and the results are reported in Section~\ref{sec::app-simu-reg} of the Supplement. In all cases, the proposed TSLA approach has the best performance. Based on the paired $t$-tests, the improvement over other competing methods is always significant at the significance level $\alpha = 0.01$. The performance gain in prediction from TSLA is more substantial when the model dimension is high and the SNR is low, where proper feature reduction is the most critical.
The RFS-Sum method is the second best in prediction, indicating the benefit of utilizing the tree structure as side information. In terms of feature aggregation, TSLA outperforms RFS-Sum and has low FNR and FPR in general.

We remark that Case~3 is a high-dimensional scenario with $n<p$ for TSLA and Enet2, but not for other models that do not consider interactions. We have designed another simulation setup with $n<p$ for all the methods; the results reported in Section~\ref{sec::app-simu-highdim} of the Supplement indicate that TSLA still performs the best. These results show the promise of our proposed framework.

\begin{table}[!tbp]
\centering
\caption{Simulation: prediction and feature aggregation performance in regression settings. Reported are the means and standard errors (in parentheses) of the MSE and FNR/FPR over 100 repeated experiments.}
\label{tab:tab1}
\small
\begin{tabular}{l|ccccc|cc}
\toprule
 & \multicolumn{5}{c}{MSE}& \multicolumn{2}{|c}{FNR/FPR} \\
 & ORE & TSLA  & RFS-Sum & Enet & Enet2    &  TSLA  & RFS-Sum\\ 
  \midrule
Case 1  & 
\begin{tabular}[c]{@{}c@{}}9.750\\(0.137)\end{tabular} & \begin{tabular}[c]{@{}c@{}}10.360\\(0.145)\end{tabular} & \begin{tabular}[c]{@{}c@{}}10.635\\(0.143)\end{tabular} & \begin{tabular}[c]{@{}c@{}}10.760\\(0.148)\end{tabular} & \begin{tabular}[c]{@{}c@{}}11.600\\(0.165)\end{tabular} & 
\begin{tabular}[c]{@{}c@{}}0.105 / 0.100 \\(0.020 / 0.010)\end{tabular} &
\begin{tabular}[c]{@{}c@{}}0.200 / 0.163\\(0.025 / 0.009)\end{tabular}\\ 
Case 2  & 
\begin{tabular}[c]{@{}c@{}}2.437\\(0.030)\end{tabular} & \begin{tabular}[c]{@{}c@{}}2.676\\(0.037)\end{tabular} & \begin{tabular}[c]{@{}c@{}}2.965\\(0.037)\end{tabular} & \begin{tabular}[c]{@{}c@{}}3.010\\(0.037)\end{tabular} & \begin{tabular}[c]{@{}c@{}}3.227\\(0.049)\end{tabular} & 
\begin{tabular}[c]{@{}c@{}}0.115 / 0.040\\(0.021 / 0.006)\end{tabular} &
\begin{tabular}[c]{@{}c@{}}0.190 / 0.096\\(0.024 / 0.007)\end{tabular} \\  
Case 3  &
\begin{tabular}[c]{@{}c@{}}2.461\\(0.034)\end{tabular} & \begin{tabular}[c]{@{}c@{}}2.981\\(0.038)\end{tabular} & \begin{tabular}[c]{@{}c@{}}3.192\\(0.038)\end{tabular} & \begin{tabular}[c]{@{}c@{}}3.208\\(0.042)\end{tabular} & \begin{tabular}[c]{@{}c@{}}3.724\\(0.050)\end{tabular} & 
\begin{tabular}[c]{@{}c@{}} 0.000 / 0.183\\(0.000 / 0.005)\end{tabular} &
\begin{tabular}[c]{@{}c@{}} 0.000 / 0.243\\(0.000 / 0.006)\end{tabular} \\
\bottomrule
\end{tabular}
\end{table}

\subsection{Simulation under Classification Settings}
\label{sec:simu-bin}

We use the same tree structures as introduced in Section~\ref{sec:simu-reg}. The binary response $y$ follows model $y \sim \mbox{Bernoulli}(q)$, where  $q  = \exp(\eta)/\{1+\exp(\eta)\}$, and
\begin{align*}
  \eta  = & -5a+b[z+5(x_1\ \lor\ x_2\ \lor\ x_3\ \lor\ x_4)+4x_9-1.5\{(x_{10}\ \lor\ x_{11}\ \lor\ x_{12})\lor\ x_{13})\}]+ \\
          & c(x_{18}\ \lor\ x_{19})-d(x_{20}\ \lor\ x_{21}\ \lor\ x_{22})  \\
          = & -5a+b[z+5(x_1\ \lor\ x_2\ \lor\ x_3\ \lor\ x_4)+4x_9-1.5(x_{10}\ \lor\ x_{11}\ \lor\ x_{12}\ \lor\ x_{13})]+ \\
          & c(x_{18}\ \lor\ x_{19})-d(x_{20}\ \lor\ x_{21}\ \lor\ x_{22}),  \,\mbox{with}\,z\sim N(0,0.25).
\end{align*}
{\cre Based on the above model, we obtain $n$ independent samples of $y$ with the generated design matrix $\X_0$ and $n$ independently sampled values of $z$.} 
We consider several scenarios,
\begin{itemize}
\item Case~1 (low-dimensional, $q\approx 0.17$): 
  Tree 1 with $n=200$ and $(a,b,c,d) = (1, 1, 0, 0)$;
\item Case~2 (low-dimensional, $q\approx 0.17$): 
  Tree 1 with $n=200$ and $(a,b,c,d) = (0.66, 0.6, 0, 0)$;
\item Case~3 (low-dimensional, $q\approx 0.30$): 
  Tree 1 with $n=200$ and $(a,b,c,d) = (0.7, 1, 0, 0)$;
\item Case~4 (high-dimensional, $q\approx 0.17$): 
  Tree 3 with $n=100$ and $(a,b,c,d) = (0.9, 1, 1.5, -3.5)$.
  \end{itemize}
For each method, we evaluate out-of-sample classification performance with independently generated testing data of sample size 1000. Besides the area under the receiver operating characteristic curve (AUC) value, we also report the area under the precision-recall curve (AUPRC) value, and the sensitivity and positive predictive value (PPV) under either 90\% or 95\% specificity; the latter measures are more useful in situations with a rare outcome, as in the suicide risk modeling to be presented in Section \ref{sec:real:suicide}. 
The rest of the setups are the same as in the regression settings.

The results on prediction performance for Case 1 and Case 4 are reported in Table~\ref{tab:tab2}. 
The rest of the results, including feature aggregation performance, estimation of unpenalized coefficients, and paired $t$-test results for comparing TSLA to other methods, can be found in Section~\ref{sec::app-simu-log} of the Supplement. An additional example with $n<p$ for all the methods is also presented in Section~\ref{sec::app-simu-highdim} of the Supplement. In all the cases, TSLA leads to significantly better performance in all metrics than the competing methods at significance level $\alpha=0.01$. 
Both TSLA and RFS-Sum perform better than the other methods. TSLA often outperforms RFS-Sum with a large margin in the AUPRC and sensitivity measures, while the improvement in the overall AUC is not substantial. We find that the practical difference between the methods gets smaller when the response becomes more balanced. 
The other observations are similar as those in the regression settings.

\begin{table}[tbp]
\centering
\caption{Simulation: prediction performance in classification settings under Case 1 and Case 4. Reported are the means and standard errors (in parentheses).}
\label{tab:tab2}
\footnotesize
\begin{tabular}{l|cccccc}
\toprule
               &            &            & \multicolumn{2}{c}{Sensitivity} & \multicolumn{2}{c}{PPV}         \\
  Model          & AUC        & AUPRC      & 90\% specificity & 95\% specificity & 90\% specificity & 95\% specificity \\
  \midrule
               & \multicolumn{6}{c}{Case 1}\\
\hline  
ORE&0.923 (0.001) & 0.713 (0.004) & 0.708 (0.006) & 0.523 (0.007) & 0.593 (0.003) & 0.682 (0.004) \\ 
TSLA&0.904 (0.002)& 0.644 (0.006) & 0.637 (0.008) & 0.444 (0.007) & 0.566 (0.004) & 0.644 (0.005) \\
RFS-Sum & 0.887 (0.003) & 0.555 (0.006) & 0.588 (0.008) & 0.367 (0.008) & 0.546 (0.004) & 0.596 (0.007) \\
Enet& 0.870 (0.003) & 0.552 (0.005) & 0.551 (0.007) & 0.365 (0.006) & 0.530 (0.004) & 0.599 (0.004) \\
Enet2&0.842 (0.004) & 0.513 (0.005) & 0.504 (0.006) & 0.318 (0.006) & 0.509 (0.004) & 0.564 (0.006)\\
  \hline
               & \multicolumn{6}{c}{Case 4}\\
  \hline
ORE&0.926 (0.002) & 0.756 (0.005) & 0.794 (0.008) & 0.593 (0.010) & 0.643 (0.004) & 0.727 (0.004) \\ 
TSLA&0.893 (0.003) & 0.635 (0.006) & 0.655 (0.009) & 0.446 (0.009) & 0.597 (0.004) & 0.666 (0.005) \\
RFS-Sum&0.881 (0.003) & 0.581 (0.006) & 0.618 (0.009) & 0.376 (0.009) & 0.583 (0.004) & 0.626 (0.006) \\ 
Enet&0.795 (0.007) & 0.479 (0.009) & 0.461 (0.011) & 0.295 (0.011) & 0.501 (0.008) & 0.554 (0.009) \\ 
Enet2&0.698 (0.008) & 0.369 (0.008) & 0.334 (0.011) & 0.204 (0.010) & 0.417 (0.008) & 0.456 (0.011) \\ 
\bottomrule
\end{tabular}
\end{table}

\subsection{Simulation: Effects of Tree Structure and Feature Rarity}
\label{sec:simu-k}

We have further investigated the impact of the tree structure and the feature rarity on the performance of the feature aggregation methods.

Both TSLA and RFS-Sum utilize the tree structure of the features. Our analysis suggests that the maximum number of child nodes, $k$, is an important structural parameter that may affect the performance of TSLA. To examine the effects of $k$, we construct simulation models with different tree structures by varying $k \in \{2, 3, 4\}$ while keeping the number of leaf nodes $p_0$ and the underlying true model the same. Our results show that the performance of TSLA is indeed better when $k=2$ as compared to $k=3,4$, while the performance of RFS-Sum is not affected much by $k$. This may be due to the fact that TSLA uses group-level regularization while RFS-Sum uses entrywise regularization. Nonetheless, TSLA outperforms the other methods in all the cases. The detailed setups and the results are summarized in Section~\ref{sec::app-simu-k-rp} of the Supplement.

To investigate the impact of feature rarity, we generate the original design matrix $\X_0$ as a sparse binary matrix with independent Bernoulli$(\delta)$ entries under three different rarity levels $\delta \in \{0.05, 0.10, 0.15\}$. The rest of the data generation follows the same process as described in the above sections. Our results show that the performance of TSLA and RFS-Sum deteriorates with decreasing $\delta$, and their difference becomes smaller when the features get sparser and  the other settings are kept the same. Indeed, the difference between ``or'' and ``sum'' operations could be small when being applied on very rare features. However, the difference of the two operations could become more and more substantial as the features are being aggregated at higher levels and thus becoming less sparse. From our results, TSLA always performs significantly better than RFS-Sum, even with $\delta = 0.05$. The detailed results are summarized in Section~\ref{sec::app-simu-k-rp} of the Supplement.

\section{Suicide Risk Modeling with EHR Data}
\label{sec:real:suicide}

\subsection{Motivation}
Suicide is a serious public health problem. According to the Centers for Disease Control and Prevention (CDC), suicide was the tenth leading cause of death in the U.S. in 2019. Recent evidence also suggests that the Covid-19 pandemic may contribute to a widespread emotional distress among the general population and further elevate the risk for mental health and suicide. With the availability of large-scale EHR data, it is of great interest to advance data-driven suicide prevention, for which a natural first step is to use medical history of patients, as recorded by ICD codes, to understand and predict the occurrence of suicidal behaviors
\citep{Barak-Corren2017,Walsh2017,Simon2018,su2020machine}. In particular, it has been demonstrated from previous studies that various prior mental and behavioral issues may be important risk factors for suicide attempts~\citep{nordentoft2011absolute, su2020machine}. This motivates us to investigate the association between suicide attempt and prior mental health issues recorded by the ``F'' chapter of the ICD-10 codes on ``mental, behavioral and neurodevelopmental disorders'' in the EHR data. As mentioned above, the common practice of collapsing full-digit ICD codes to fewer digits may lead to substantial information loss. Here, we aim to apply our proposed data-driven feature aggregation approach to determine the right levels of specificity of the hundreds of mental health diagnosis conditions for better revealing their associations with suicide.

\subsection{Data and Setup}

We use data from the \emph{Kansas Health Information Network} (KHIN), which contains information on inpatient, outpatient, and emergency room encounters for patients in the network from January 1, 2014 to December 31, 2017. Suicide attempts are identified by both the external cause of injury ICD codes and other ICD code combinations that are indicative of suicidal behaviors \citep{Patrick2010,Chen2017a}. Since the data contains both ICD-9 and ICD-10 codes, we convert the ICD-9 codes to ICD-10 using the public toolkit AHRQ MapIT (\href{https://qualityindicators.ahrq.gov/resources/toolkits}{https://qualityindicators.ahrq.gov/resources/toolkits}) and all the non-unique mappings related to suicide are also checked and validated by our health research team.

We conduct the study with a nested case-control design \citep{Thomas1977, Prentice1978,ERNSTER1994} and use risk-set sampling to match the cases and the controls. This approach is frequently used with observational data when the outcome is rare; some examples of studying suicide with this design include~\citet{martinez2005antidepressant}, \citet{agerbo2002familial}, and \citet{Mitra2022NLP}. 
In our study, the base cohort consists of all patients who were with at least one diagnosis record or procedure claim 
and were between 18 to 64 years of age at the time of the first suicide diagnosis (for patients who had suicide attempt diagnosis) or the last visit (for patients who did not have suicide attempt diagnosis). Patients whose first recorded visit was suicide attempt are excluded due to lack of historical information. The base cohort in total contains 272,643 patients.

The outcome of interest is the occurrence of suicide attempt during the study period.
Cases include all patients in the base cohort who had suicide attempt diagnosis in the study period. For each case, the index date is set as the date of the suicide attempt, and we randomly sample 10 controls without replacement among patients who were of the same gender and age, had not experienced suicide attempts, and had a non-suicidal visit within 30 days before or after the index date; for the case and each of the controls, all the records before the index date are aggregated to produce the demographic features and binary features of prior diagnoses. 
By design, a patient can be a control for multiple cases and a case could serve as a control for another case who attempted suicide at an earlier date; this is typical in nested case-control studies. The flowchart of the design is presented in Section~\ref{sec::app-suicide-data} of the Supplement. 

The final case-control data set consists of 13,398 records, in which 1218 are cases and 12180 are matched controls. There are 356 full-digit ICD-10 codes under the ``F'' chapter, which can be collapsed into 70 three-digit codes. If we pre-screen the full-digit ICD codes to discard those with less than 0.04\% prevalence, there will be 173 full-digit ICD codes left, which can be aggregated to 48 three-digit ICD codes.
Table~\ref{tab:tab3} provides some summary statistics of the nested case-control data. The prevalence of the F-chapter ICD codes is much higher in the cases than in the controls, as expected. The detailed prevalence information for each of the three-digit ICD codes is shown in Section~\ref{sec::app-suicide-data} of the Supplement.

\begin{table}[tbp]
  \caption{Suicide risk study: summary statistics for the nested case-control dataset. Controls are matched to cases on age and gender. }
\label{tab:tab3}
\small
\centering
\begin{tabular}{lcc}
\toprule
 & \multicolumn{2}{c}{Summary Statistics} \\
 \cline{2-3}
Variable  & Cases ($n$ = 1218)  & Controls ($n$ = 12180) \\
\midrule
Male ($n$\%)  & 485 (39.81\%) & 4850 (39.81\%) \\

Age ($n$(\%))   & \multicolumn{2}{c}{} \\
\multicolumn{1}{c}{18-25 years old}        & 424 (34.81\%)     & 4240 (34.81\%)\\
\multicolumn{1}{c}{26-39 years old}        & 389 (31.94\%)     & 3890 (31.94\%)\\
\multicolumn{1}{c}{40-64 years old}        & 405 (33.25\%)     & 4050 (33.25\%)\\
F-code prevalence (\%)                                   & \multicolumn{1}{l}{}                 & \multicolumn{1}{l}{}                 \\
  \multicolumn{1}{c}{Full-digit ($p_0$ = 356)}  & 0.79\% & 0.22\%  \\
  \multicolumn{1}{c}{Three-digit ($p_0$ = 70)} & 3.15\%& 0.94\%\\
F-code prevalence after pre-screening(\%)           & \multicolumn{1}{l}{}                 & \multicolumn{1}{l}{}                 \\
  \multicolumn{1}{c}{Full-digit ($p_0$ = 173)}  & 1.57\% & 0.43\%  \\
  \multicolumn{1}{c}{Three-digit ($p_0$ = 48)} & 4.49\%& 1.33\%\\
\bottomrule
\end{tabular}
\end{table}

With the highly sparse and binary ICD features, we apply the proposed logic selection and aggregation framework to study the association between the hundreds of mental health diagnoses and the risk of suicide attempt. We test two models with the TSLA approach: one based on the 356 full-digit ICD codes (TSLA) and another based on the 173 full-digit codes after the prevalence-based screening (TSLA-screen). Several competing methods are considered, including RFS-Sum, Enet, and Enet2. 
Age and gender are included as covariates in all the models. 

\subsection{Comparison with Competing Methods in Prediction}

Though our main focus is on exploring the association between prior mental health conditions and suicide risk based on the proposed logic aggregation framework, it is important to demonstrate that the models produced by TSLA are of high-quality as compared to those by the other competing methods. We thus first evaluate the out-of-sample prediction performance of all methods with a 10-fold splitting procedure. The cases and the controls are both randomly divided to 10 folds, and each time we use 9 folds of the cases and the controls for training and one fold for testing. We use the same evaluation metrics as in Section \ref{sec:simu-bin} and average the results from the 10 splits. Specifically, with each method, we compute out-of-sample AUC and AUPRC, and in addition, we identify either 5\% or 10\% of the testing patients with the highest risk and evaluate the associated sensitivity and PPV; due to the low prevalence of the events, these would be roughly similar to considering 5\% or 10\% specificity.

\begin{table}[tb]
  \caption{Suicide risk study: prediction performance with 10-fold splitting procedure. Reported are the means and standard errors across the 10 splits. For each metric, the best result is highlighted.}
\label{tab:tab4}
\centering
\footnotesize
\begin{tabular}{l|cccccc}
\toprule
 &  &  & \multicolumn{2}{c}{Sensitivity}  & \multicolumn{2}{c}{PPV}  \\ 
Model  & AUC   & AUPRC    & 10\% positive    & 5\% positive   & 10\% positive    & 5\% positive \\
\midrule
TSLA &     
\begin{tabular}[c]{@{}c@{}}\textbf{72.9\%}\\ (0.8\%)\end{tabular} & \begin{tabular}[c]{@{}c@{}}\textbf{29.0\%}\\ (1.5\%)\end{tabular} & \begin{tabular}[c]{@{}c@{}}\textbf{36.1\%}\\ (1.4\%)\end{tabular} & \begin{tabular}[c]{@{}c@{}}\textbf{22.7\%}\\ (1.2\%)\end{tabular} & \begin{tabular}[c]{@{}c@{}}\textbf{32.8\%}\\ (1.3\%)\end{tabular} & \begin{tabular}[c]{@{}c@{}}\textbf{41.3\%}\\ (2.1\%)\end{tabular}\\
TSLA-screen   & 
\begin{tabular}[c]{@{}c@{}}72.9\%\\ (0.8\%)\end{tabular} & \begin{tabular}[c]{@{}c@{}}28.9\%\\ (1.6\%)\end{tabular} & \begin{tabular}[c]{@{}c@{}}36.0\%\\ (1.4\%)\end{tabular} & \begin{tabular}[c]{@{}c@{}}22.5\%\\ (1.2\%)\end{tabular} & \begin{tabular}[c]{@{}c@{}}32.7\%\\ (1.3\%)\end{tabular} & \begin{tabular}[c]{@{}c@{}}41.0\%\\ (2.2\%)\end{tabular} \\
RFS-Sum &
\begin{tabular}[c]{@{}c@{}}72.7\%\\ (0.8\%)\end{tabular} & \begin{tabular}[c]{@{}c@{}}28.7\%\\ (1.4\%)\end{tabular} & \begin{tabular}[c]{@{}c@{}}35.5\%\\ (1.2\%)\end{tabular} & \begin{tabular}[c]{@{}c@{}}21.8\%\\ (1.2\%)\end{tabular} & \begin{tabular}[c]{@{}c@{}}32.2\%\\ (1.1\%)\end{tabular} & \begin{tabular}[c]{@{}c@{}}39.7\%\\ (2.2\%)\end{tabular} \\
Enet   & 
\begin{tabular}[c]{@{}c@{}}72.5\%\\ (0.9\%)\end{tabular} & \begin{tabular}[c]{@{}c@{}}28.1\%\\ (1.5\%)\end{tabular} & \begin{tabular}[c]{@{}c@{}}35.2\%\\ (1.1\%)\end{tabular} & \begin{tabular}[c]{@{}c@{}}21.5\%\\ (1.1\%)\end{tabular} & \begin{tabular}[c]{@{}c@{}}32.0\%\\ (1.0\%)\end{tabular} & \begin{tabular}[c]{@{}c@{}}39.1\%\\ (2.0\%)\end{tabular} \\
Enet2  & 
\begin{tabular}[c]{@{}c@{}}71.9\%\\ (0.9\%)\end{tabular} & \begin{tabular}[c]{@{}c@{}}28.0\%\\ (1.6\%)\end{tabular} & \begin{tabular}[c]{@{}c@{}}34.8\%\\ (1.3\%)\end{tabular} & \begin{tabular}[c]{@{}c@{}}21.8\%\\ (1.1\%)\end{tabular} & \begin{tabular}[c]{@{}c@{}}31.7\%\\ (1.2\%)\end{tabular} & \begin{tabular}[c]{@{}c@{}}39.6\%\\ (2.0\%)\end{tabular} \\
\bottomrule
\end{tabular}
\end{table}

Table~\ref{tab:tab4} shows the out-of-sample prediction performance of all methods. We also provide the paired $t$-test results to compare TSLA and other methods in Section~~\ref{sec::app-suicide-result} of the Supplement. 
We remark that the baseline model of using age and gender as the only covariates leads to a mean AUC of 52.3\% (0.4\%), because the two variables are already being used in constructing the case-control cohort; we thus omit it in the discussion. The TSLA models have comparable or better performance in AUC, AUPRC, sensitivity, and PPV than the other methods, although the improvements are generally not substantial and are mostly not statistically significant based on the paired $t$-tests assuming the results across the 10 splits are independent of each other. 
Nonetheless, TSLA leads to the highest sensitivity, 22.7\%, when 5\% of patients are determined as at high risk, as compared to 21.8\% from the best competitor RFS-Sum. This has an important clinical implication: if we label 5\% of patients as the high risk group of suicide, our model is able to capture 22.7\% of patients who are going to attempt suicide. We also observe that the performance of TSLA remains similar with or without the pre-screening of ICD codes. 
One plausible explanation is that the inclusion of the extremely rare ICD codes, while potentially bringing in useful information, may at the same time introduce noise and make the selection and aggregation more challenging due to increased model dimensionality. 

We remark that this study does not intend to build the best possible predictive model of suicide with EHR data; however, the results do suggest that TSLA could be utilized as an effective tool of feature engineering and selection to improve existing suicide risk algorithms \citep{su2020machine}. 

\subsection{Aggregation Patterns of F-Chapter ICD Codes for Suicide Risk}

To examine the aggregation patterns of the F-chapter ICD codes in modeling the risk of suicide attempt, we fit the TSLA model with the full nested case-control data. The hierarchical tree structure of the full-digit F-chapter ICD codes is with $p_0=356$ leaf nodes and of depth $h=5$.

\begin{figure}[tbp]
  \centering
  \includegraphics[width=.9\linewidth]{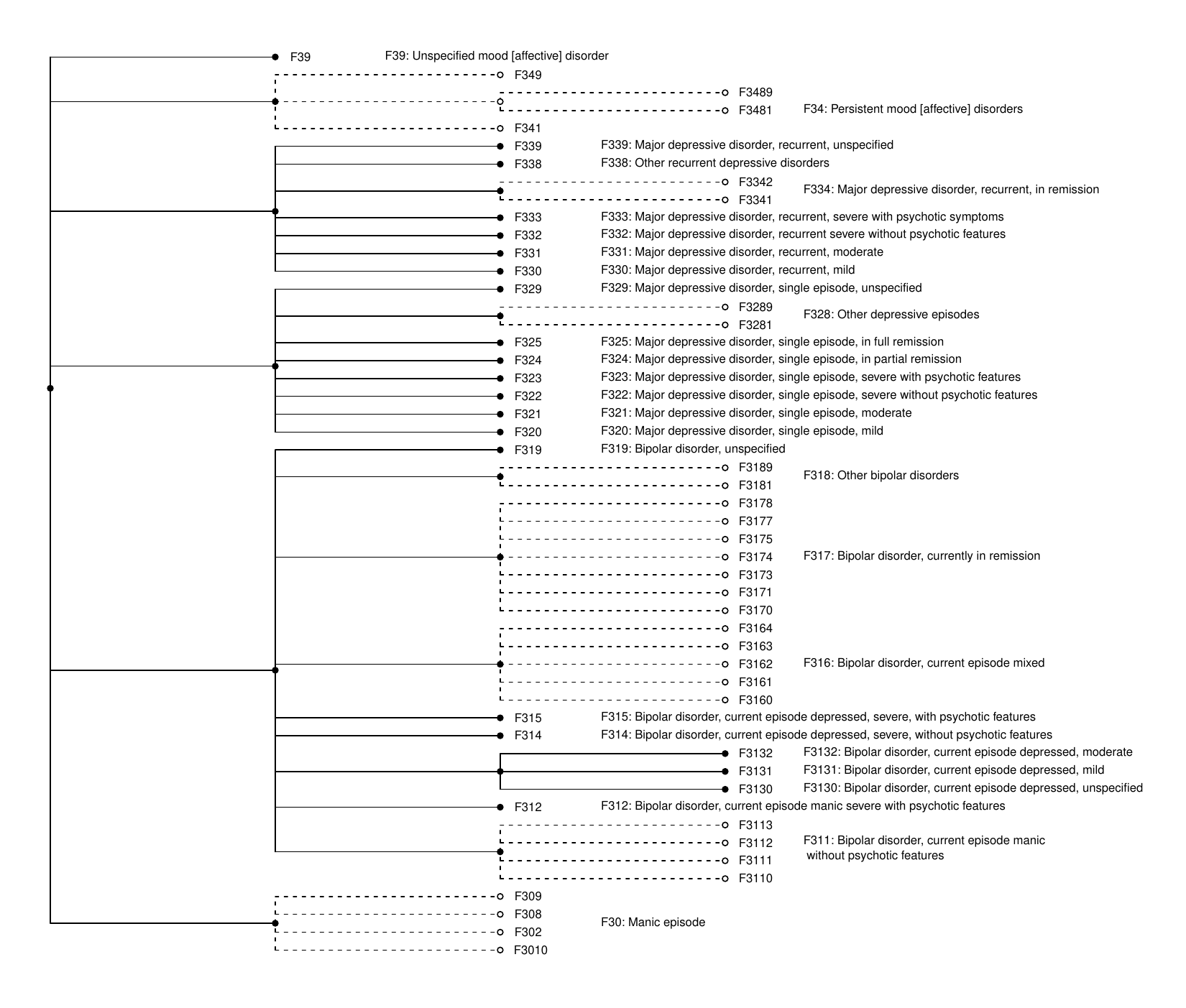}
  \caption{Suicide risk study: selection and aggregation of F3 codes (Mood (affective) disorders). The selected codes are indicated by closed circles. The codes that are being aggregated are indicated by open circles and dashed lines.}
  \label{fig:fig3}
\end{figure}

\begin{figure}[tbp]
\centering
\includegraphics[width=.9\linewidth]{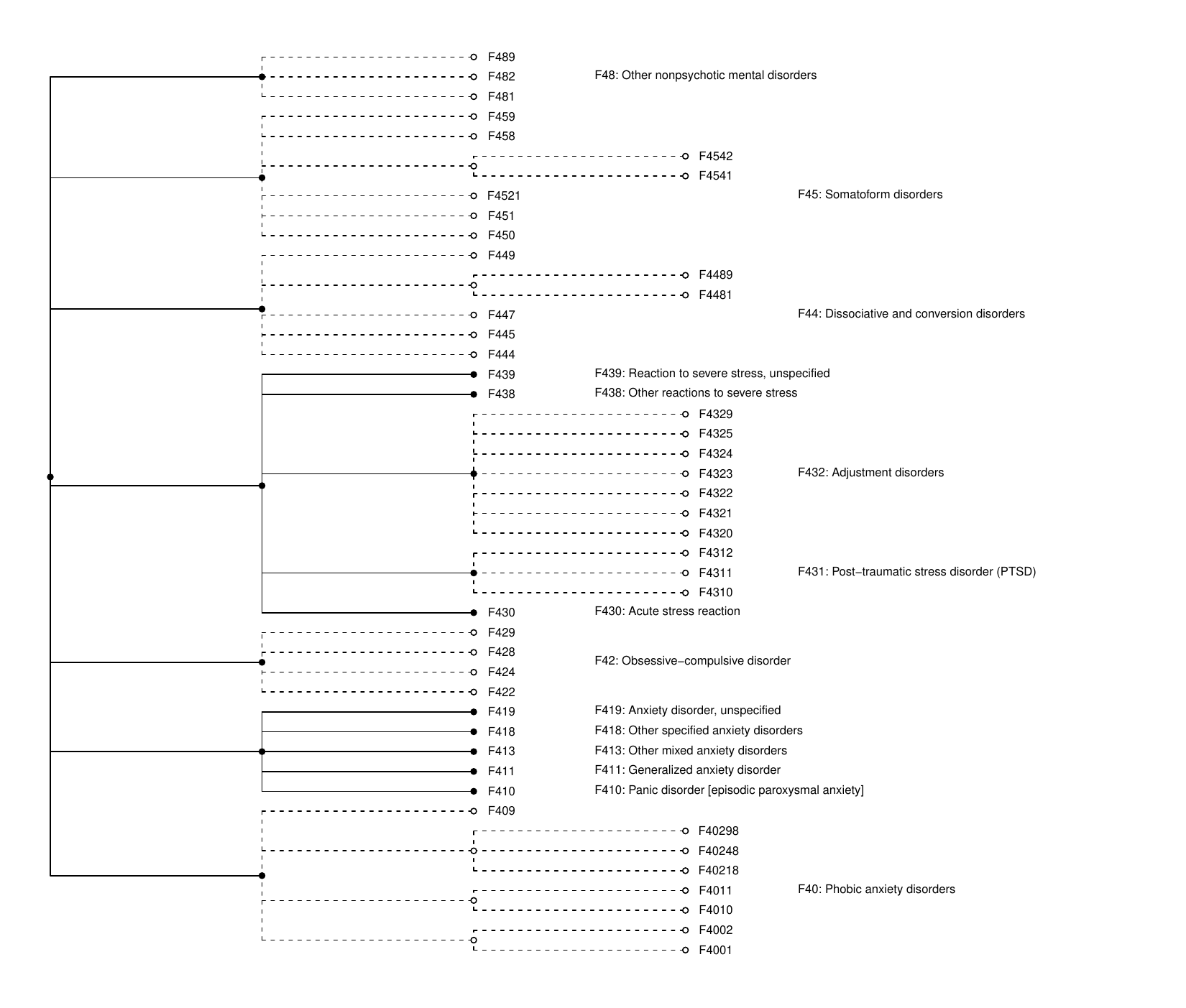}
\caption{Suicide risk study: selection and aggregation of F4 codes (Anxiety, dissociative, stress-related, somatoform and other nonpsychotic mental disorders). The settings are the same as in Figure \ref{fig:fig3}.} 
\label{fig:fig4}
\end{figure}

\begin{figure}[tbp]
  \centering
  \includegraphics[width=.9\linewidth]{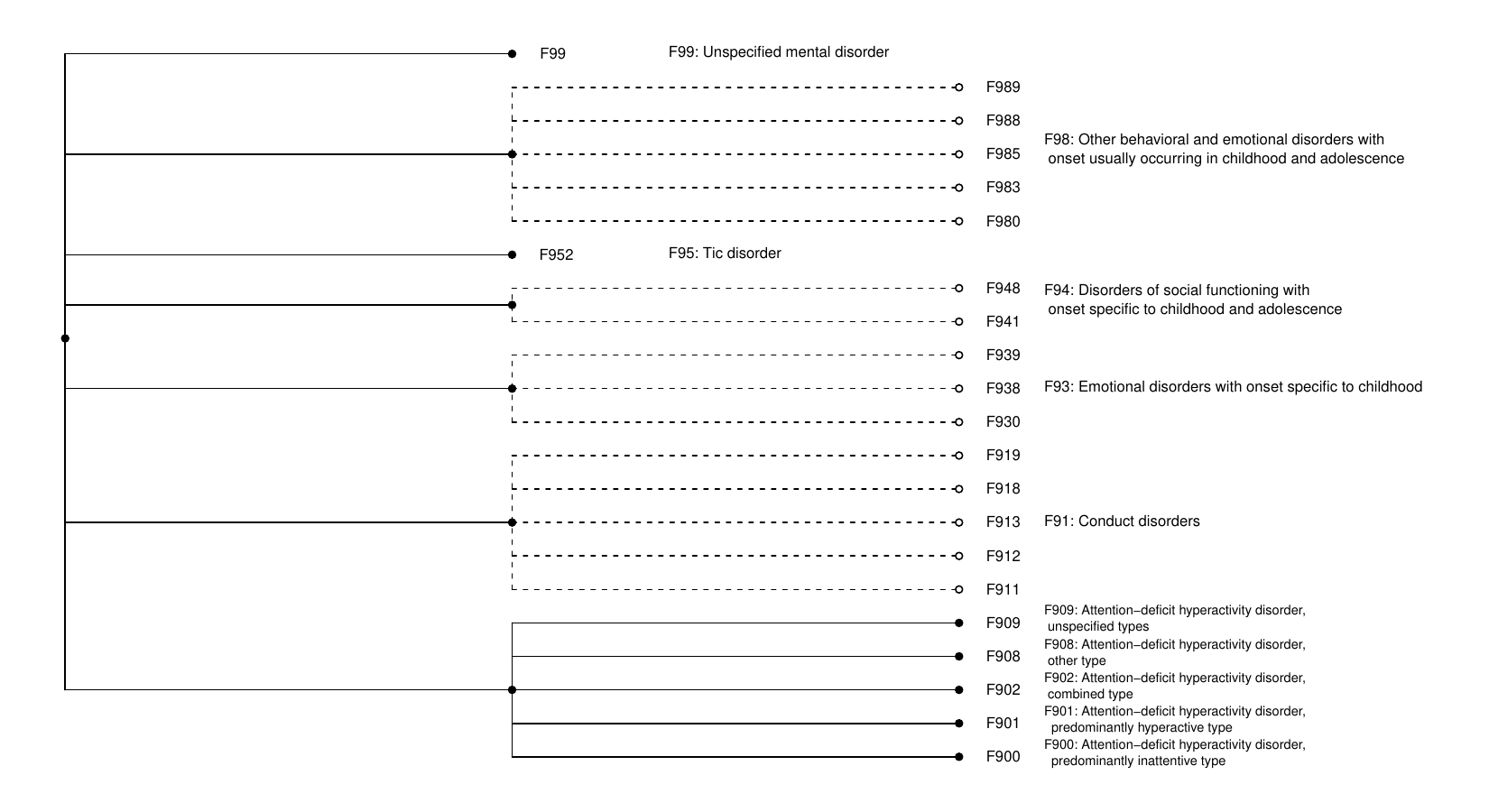}
  \caption{Suicide risk study: selection and aggregation of F9 codes (Behavioral and emotional disorders with onset usually occurring in childhood and adolescence \& unspecified mental disorder). The settings are the same as in Figure \ref{fig:fig3}.}
  \label{fig:fig5}
\end{figure}

From the fitted TSLA model with $p_0=356$ binary features of ICD codes, 297 features are aggregated into 41 denser features in conformity to the tree structure, while the rest of the 59 raw features are kept unaggregated. 
The ICD codes under the two-digit categories F0, F5, F6, F7, and F8 are all aggregated to their first two digits, while more detailed ICD codes are kept in the other categories F1, F2, F3, F4, and F9. 
To examine further, Figures~\ref{fig:fig3}, \ref{fig:fig4}, and \ref{fig:fig5} show the selection and aggregation patterns of the categories F3, F4, and F9, respectively. 
The aggregation patterns produced by TSLA with the prevalence-based pre-screening (TSLA-screen) are quite similar. Additional detailed results are reported in Section~\ref{sec::app-suicide-result} and \ref{sec::app-suicide-result-screen} of the Supplement.

In the category of F3 (Mood (affective) disorders), most full-digit codes can be aggregated to four-digit level. Only a few five-digit codes, namely, F3132, F3131, and F3130, which indicate different severity levels of bipolar disorder with ongoing episode of depression, show distinct effects on suicide risk. On the other hand, most codes cannot be further aggregated to three-digit level. This reveals the complexity of the linkage between mood disorders and suicide and the danger of naively aggregating all ICD codes to the three-digit level. In particular, the aggregation pattern suggests that different subtypes of bipolar disorder coded under F31 could have differential effects on suicide. Recent epidemiological studies have shown that patients with bipolar disorder have a higher risk of committing suicide than patients with other psychiatric or medical disorders, but it remains unclear whether any bipolar disorder subtype is associated with a higher level of suicidality than the others \citep{Dome2019,Miller2020}. Our results indicate that all 11 raw or aggregated conditions under F31 have positive effects on suicide risk; among those, F314 (Bipolar disorder, current episode depressed, severe, without psychotic features) shows the largest effect with an estimated odds ratio 2.16, and F317 (bipolar disorder, currently in remission) shows the least effect with an estimated odds ratio 1.04. These shed light on distinguishing the bipolar disorder subtypes that are the most susceptible to suicide risk. 

In the category F4 (Anxiety, dissociative, stress-related, somatoform and other nonpsychotic mental disorders), the sub-categories of F41 (Other anxiety disorders), and F43 (Reaction to severe stress, and adjustment disorders) are distinctive, while all the other codes can be aggregated to the three-digit level. This is partly because some of their sub-categories, e.g., F419 (Anxiety disorder, unspecified), F431 (Post-traumatic stress disorder (PTSD)), and F439 (Reaction to severe stress, unspecified), show relatively strong associations with suicide risk \citep{Nepon2010, gradus2010posttraumatic}, with estimated odds ratio 1.27, 1.38, and 1.41, respectively. In the category F9 (Behavioral and emotional disorders with onset usually occurring in childhood and adolescence \& unspecified mental disorder), the sub-categories of F90 (Attention-deficit hyperactivity disorders (ADHD)) have distinct effects on suicide. Indeed, existing studies have shown that ADHD is associated with an increased risk of suicidal behavior, and the impacts on suicide from the three main subtypes of ADHD, namely, inattentive (F900), hyperactive (F901), and combined (F902), are different~\citep{furczyk2014adult,james2004attention}.

Overall, our TSLA analysis suggests that while a large number of the F-chapter ICD codes can be aggregated to the three-digit level, there are several important cases where the finer sub-categories should be kept as they show distinct effects on suicide risk. These findings are well supported by epidemiological studies and meta reviews and provide valuable insights on the importance and specificity of mental, behavioral, and neurodevelopmental disorders to suicide risk.

\section{Discussion}
\label{sec:dis}

We develop a novel tree-guided feature selection and logic aggregation framework with the logic operation of ``or'', for dealing with sparse, binary features with hierarchical structure from EHR data. By tree-guided feature expansion, reparameterization, and regularization, our approach converts the notoriously difficult combinatorial logic regression problem into a convex learning problem that permits more scalable computation. Numerical studies demonstrate that the framework can make productive use of the rare features of diagnosis conditions from EHR data to enhance model prediction and interpretation. The proposed approach is widely applicable.

There are many directions for future research.
Given that EHR data are often of very large-scale, it is important to explore more scalable computational algorithms or strategies; in particular, it would be promising to develop stagewise estimation techniques \citep{tibshirani2015general} for feature aggregation. From our theoretical derivations and empirical results, it is possible to better quantify the error bound of TSLA by more explicitly taking into account the feature rarity and the aggregation pattern of the true model. The TSLA approach can be extended to other modeling frameworks such as survival analysis, for which the SPG algorithm can still be utilized to handle various combinations of penalty functions and loss functions. 
Given that EHR data are longitudinal in nature and utilizing longitudinal diagnosis information could lead to better risk identification in suicide risk study, it is pressing to extend our work to longitudinal models. While our method is designed to better harvest information from rare features,
it remains a challenging issue to deal with extremely rare features of very high dimensions. As suggested by the empirical results in Section~\ref{sec:real:suicide}, in such scenarios it could be beneficial to consider complementing the TSLA analysis with some pre-screening or pre-aggravation strategies \citep{FanLv2008}, which could alleviate the computational burden without much information loss.   
Last but not least, we will explore more general logic operations on rare binary features with a more general graphical association structure.




\clearpage
\appendix

\noindent {\bf \LARGE Supplement}

\section{Computation}
\label{sec::app-alm}

To solve the optimization problem in (10) of the main paper, we first rewrite it as an unconstrained problem of $\bgamma$ and $\mu$ by replacing $\bbeta$ with $\A\bgamma$, which gives
\begin{align}
    (\hat\mu,\ \hat\bgamma)=\, &\arg\min_{\mu, \bgamma}\{\frac{1}{2n}\|\y-\mu\1-\X\A\bgamma\|^2+
    \lambda(1-\alpha)\|\D\A\bgamma\|_1+
    \lambda\alpha P_T(\bgamma)\},
    \label{eq:optimprob2}
\end{align}
where $\D\in \mathbb{R}^{p\times p}$ is a diagonal matrix with the $j^{th}$ diagonal element as $\tilde{w}_j$. 
This is a convex problem, but the
penalty function is non-smooth. 
We thus adopt a \emph{Smoothing Proximal Gradient} (SPG) algorithm \citep{chen2012smoothing} for the optimization.

There are three key steps in the SPG algorithm. First, we decouple the penalty terms via dual norms. To be specific, since the $\ell_1$ norm is the dual norm of the $\ell_{\infty}$ norm, the second term in \eqref{eq:optimprob2}
can be expressed as
\begin{align*}
    \lambda(1-\alpha)\|\D\A\bgamma\|_1\ =\ \max_{\beeta_1\in \mathbb{Q}_1} \beeta_1^{\trans}\C_1\bgamma,
\end{align*}
where $\C_1=\lambda(1-\alpha)\D\A$, $\beeta_1\in\mathbb{R}^{p}$, and $\mathbb{Q}_1=\{\beeta_1 \mid  \|\beeta_1\|_{\infty}\leq 1\}$.
Similarly, since the dual norm of the $\ell_2$ norm is still the $\ell_2$ norm,
we can write the third term in \eqref{eq:optimprob2} as
\begin{align*}
     \lambda\alpha P_T(\bgamma)\
     & =\ \lambda\alpha\sum_{g\in \mathcal{G}}w_g\|\bgamma_g\|
     =\ \lambda\alpha\sum_{g\in\mathcal{G}}w_g\max_{
     \|\beeta_g\|\leq 1}\beeta_g^{\trans}\bgamma_g \\
    & =\ \max_{\beeta_2\in \mathbb{Q}_2}\sum_{g\in \mathcal{G}}\lambda\alpha w_g\beeta_g^{\trans}\bgamma_g
    =\max_{\beeta_2\in \mathbb{Q}_2}\beeta_2^{\trans}\C_2\bgamma,
\end{align*}
where $\beeta_g\in\mathbb{R}^{p_g}$, $\beeta_2$ is a vector of length $\sum_{g\in\mathcal{G}}p_g$, collecting all $\beeta_g$ for $g\in\mathcal{G}$, and  
$\mathbb{Q}_2=\{\beeta_2 \mid \|\beeta_g\|\leq 1,\ \forall g\in \mathcal{G}\}$.
The matrix $\C_2 \in \mathbb{R}^{\sum_{g\in\mathcal{G}}p_g\times |\T|}$ is defined as follows. The rows of $\C_2$ correspond to each pair of $(u,g) \in \{(u,g) \mid u\in g,\  g\in\mathcal{G},\ u\in\T\}$ and the columns of $\C_2$ correspond to each node $v\in\T$. Then each element of $\C_2$ is given by ${\C_2}_{((u,g),v)}=\lambda\alpha w_g$ if $u=v$ and ${\C_2}_{((u,g),v)}=0$ otherwise.
It then follows that the penalty part in \eqref{eq:optimprob2} can be expressed as
\begin{equation}
    \label{eq:dualpenalty}
    \Omega(\bgamma)=\lambda(1-\alpha)\|\D\A\bgamma\|_1+\lambda\alpha P_T(\bgamma)\ =\
    \max_{\beeta_1\in \mathbb{Q}_1} \beeta_1^{\trans}\C_1\bgamma+
    \max_{\beeta_2\in \mathbb{Q}_2}\beeta_2^{\trans}\C_2\bgamma.
\end{equation}

In the second step, by using a smoothing technique proposed in \citet{nesterov2005smooth}, we obtain a smooth approximation of $\Omega(\bgamma)$ as
\begin{align}
    f_{\tau}(\bgamma)\ =\ \max_{\beeta_1\in\mathbb{Q}_1}(\beeta_1^{\trans}\C_1\bgamma-\frac{\tau}{2} \|\beeta_1\|^2)+
    \max_{\beeta_2\in\mathbb{Q}_2}(\beeta_2^{\trans}\C_2\bgamma-\frac{\tau}{2} \|\beeta_2\|^2),
    \label{eq:smoother}
\end{align}
with $\tau$ being the positive smoothness parameter.
Then $f_{\tau}(\bgamma)$ is smooth with respect to
$\bgamma$ and the Lipschitz continuous gradient is given as $\C_1^{\trans}\beeta_1^*+\C_2^{\trans}\beeta_2^*$, where
$(\beeta_1^*,\beeta_2^*)$ is the optimal solution of \eqref{eq:smoother}. The Lipschitz constant is $L_{\tau}=\|\C_1\|^2/\tau+\|\C_2\|^2/\tau$, where $\|\C_1\|$ and $\|\C_2\|$ are the matrix spectral norms.
As we use a combination of group lasso penalty and generalized lasso penalty, $\beeta_1^*$ and $\beeta_2^*$ have closed-form expressions, and the Lipschitz constant is explicit \citep{chen2012smoothing}.
By substituting $\Omega(\bgamma)$ in \eqref{eq:optimprob2} with its smooth approximation $f_{\tau}({\bgamma})$, we obtain the approximated optimization problem
\begin{align}
    (\hat\mu,\ \hat\bgamma)\ =\ \arg\min_{\mu, \bgamma} h(\mu,\ \bgamma)\ =\  \arg\min_{\mu, \bgamma}\{\frac{1}{2n}\|\y-\mu\1-\X\A\bgamma\|^2+
    f_{\tau}(\bgamma)\}.
    \label{eq:appprob}
\end{align}
The gradient of $h$ with respect to $\mu$ and $\bgamma$ is
straightforward with an explicit Lipschitz constant $L$. 

Finally, 
we perform optimization based on the approximated problem \eqref{eq:appprob}, which is convex and smooth.
For large-scale data, the \emph{fast iterative shrinkage-thresholding algorithm} (FISTA)~\citep{beck2009fast} is applied in SPG for optimization.
The FISTA combines the gradient descent method with the Nesterov acceleration technique~\citep{nesterov1983method}, which leads to faster convergence than standard gradient-based methods.

Let $t$ be the index of the iteration, and
denote $\mu^{(0)}$ and $\bgamma^{(0)}$ as the initial values of the coefficients with $t=0$. Following the discussion above, the optimization steps for problem \eqref{eq:appprob} can be summarized in Algorithm~\ref{alg:fista}. In our numerical studies, we take $\mu^{(0)}=0$ and $\bgamma^{(0)}=\0$.

\begin{algorithm}[tbp]
	\caption{Fast Iterative Shrinkage-Thresholding Algorithm for TSLA}
	\label{alg:fista}
        \KwIn{$\X$,  $\y$, $\C_1$, $\C_2$, $\A$, $L$, $\tau$,  $\mu^{(0)}$, $\bgamma^{(0)}$}
        \textbf{Initialization:}  $\theta^{(0)}=1$, $w_1^{(0)}=\mu^{(0)}$, $\w_2^{(0)}=\bgamma^{(0)}$

        \textbf{Repeat}\linebreak
        1. Compute ${\beeta_1^*}^{(t)}$ and ${\beeta_2^*}^{(t)}$ with $\bgamma^{(t)}$, $\C_1$, $\C_2$ and $\tau$. \linebreak
        2. Get the gradient $\nabla h(w_1^{(t)})$ and $\nabla h(\w_2^{(t)})$ with \linebreak
        $\nabla h(w_1^{(t)})=-(1/n)\1^{\trans}(\y-w_1^{(t)}\1-\X\A\w_2^{(t)})$,
        $\nabla h(\w_2^{(t)})=-(1/n)(\X\A)^{\trans}(\y-w_1^{(t)}\1-\X\A\w_2^{(t)})+\C_1^{\trans}{\beeta_1^*}^{(t)}+\C_2^{\trans}{\beeta_2^*}^{(t)}$.  \linebreak
        3. Update the current estimations by \linebreak
        $\mu^{(t+1)}=w_1^{(t)}-(1/L)\nabla h({w_1}^{(t)})$,
        $\bgamma^{(t+1)}=\w_2^{(t)}-(1/L)\nabla h(\w_2^{(t)})$. \linebreak
        4. Set $\theta^{(t+1)}=(1+\sqrt{1+4(\theta^{(t)})^2})/2$. \linebreak
        5. Set \linebreak 
        $w_1^{(t+1)}=\mu^{(t+1)}+\{(\theta^{(t)}-1)/\theta^{(t+1)}\}(\mu^{(t+1)}-\mu^{(t)})$, 
        $\w_2^{(t+1)}=\bgamma^{(t+1)}+\{(\theta^{(t)}-1)/\theta^{(t+1)}\}(\bgamma^{(t+1)}-\bgamma^{(t)})$.

    \textbf{Until} convergence \linebreak $(|\mu^{(t+1)}-\mu^{(t)}|+\|\bgamma^{(t+1)}-\bgamma^{(t)}\|_1)/(|\mu^{(t)}|+\|\bgamma^{(t)}\|_1)\leq\ 10^{-5}$.

    \KwOut{$\mu^{(t)}$ and $\bgamma^{(t)}$ at convergence }
\end{algorithm}

The optimal combination of the tuning parameters $\alpha$ and $\lambda$ can be determined via cross validation. The range of $\alpha$ is set as $[0,\ 1]$. For $\lambda$, we compute an approximated upper bound $\lambda^*$ and take the range as $(0, \ \lambda^*)$. 
More specifically, for the tuning parameter $\lambda$, we let
\begin{align}
    \lambda^*=\max\{\lambda_1^*,\ \lambda_2^*\},\
    \lambda_1^*=\max_{1\leq j\leq p}\frac{|\x_j^{\trans}\y|}{n\tilde{w}_j},
    \ \lambda_2^*=\max_{g\in\mathcal{G}}\frac{\|\X_g^{\trans}\y\|}{n w_g},
\end{align}
where for the group $g$, $\X_g=\X\A{\bf P}_g$, and {\cre${\bf P}_g \in \{0, 1\}^{|\T|\times|\T|}$} is a diagonal matrix with the diagonal elements given by 
\begin{align*}
    ({\bf P}_g)_{uu}=
    \begin{cases}
      1, & \mbox{if node}\ u \in g; \\
      0, & \mbox{otherwise}.
    \end{cases}
\end{align*}
The weights $\tilde{w}_j$ and $w_g$ are specified in Section~2.3 of the main paper. Note that, $\lambda_1^*$ is the smallest value of $\lambda$ for which all the $\beta$ coefficients are estimated to be zero when $\alpha=0$ and $\lambda_2^*$ is the smallest value of $\lambda$ for which all the $\gamma$ coefficients are zero when $\alpha=1$. This bound works well in practice. The smoothness parameter $\tau$ controls the gap between $\Omega(\bgamma)$ and $f_{\tau}(\bgamma)$. In practice, for large-scale problems, setting $\tau$ as a small positive value can provide good approximation accuracy \citep{chen2012smoothing}; we set $\tau$ as $10^{-3}$ in all the numerical studies.

The algorithm can be readily extended to the settings of logistic regression when $y$ is a binary response. 
The objective function for logistic regression can be derived by substituting the first term in \eqref{eq:optimprob2} with the negative log-likelihood function:
\begin{align*}
    \sum_{i=1}^n[-w_1y_i\log\mu_i-w_0(1-y_i)\log(1-\mu_i)], \qquad \mu_i=\frac{e^{\mu+\x_i^{\trans}\A\bgamma}}{1+e^{\mu+\x_i^{\trans}\A\bgamma}},
\end{align*}
where $w_1$ and $w_0$ are the weights
of positive and negative cases, accordingly. Since the negative log-likelihood function is convex and smooth, the application of SPG is straightforward.

\clearpage
\section{Theory}
\label{sec:thm}

\subsection{Theoretical Results}

In this section, we study the theoretical properties of the estimator from the proposed TSLA approach under a general high-dimensional learning framework. We consider the regression model without the intercept term
\begin{align}
	\y = \X\bbeta + \bvare,
	\label{eq:lr2}
\end{align}
where $\X$ is the expanded binary matrix. As all the binary features are rare, we assume that there is a baseline group of observations for which the feature vector is a zero vector. The intercept then represents the group mean for these baseline observations. Without loss of generality, we assume that the response $\y$ in \eqref{eq:lr2} has been centered by the baseline mean. Then the intercept term can be excluded from the model.

Under the TSLA framework, we investigate the finite sample properties of the regularized estimator:
\begin{align}
    (\hat\bbeta,\ \hat\bgamma)=\, &\arg\min_{\bbeta, \bgamma}\{\frac{1}{2n}\|\y-\X\bbeta\|^2+
    \lambda \Omega(\bbeta, \bgamma; \alpha)\},
    \ \mbox{s.t}\ \bbeta\ =\ \A \bgamma,
    \label{eq:optimprob3}
\end{align}
where 
\begin{align}
\Omega(\bbeta, \bgamma; \alpha) = (1-\alpha)\sum_{j=1}^p\tilde{w}_j|\beta_j|+
    \alpha P_T(\bgamma). \label{eq:penalty-supp}
\end{align}

We begin with a discussion on the dimension of the feature aggregation problem under certain tree structure. 

\begin{definition}
A \emph{k-ary} tree is a rooted tree where every internal node has at most $k$ child nodes.
\end{definition}
\begin{definition}
A perfect full \emph{k-ary} tree is a rooted tree where every internal node has exactly $k$ child nodes and all the leaf nodes are at the same depth.
\end{definition}

In TSLA, the dimension of regression coefficient vector $\bbeta$, $p$, is the same as the number of leaf nodes in the expanded tree $\T$.
Based on the properties of the \emph{k-ary} tree, we can compare the model dimension before and after the tree-guided expansion, that is, compare $p_0$ and $p$ in terms of $k$ and $h$. 

\begin{lemma}
\label{lemma:k}
Suppose the tree structure of the original $p_0$ features is a perfect full $k$-ary tree of depth $h$. When all the interaction features generated by the tree-guided expansion are nonzero vectors, it holds that $p_0=k^h$, and $p = \mathcal{O}(k^h2^k/(k-1))$. Consequently, when the original tree is a general $k$-ary tree of depth $h$, we always have that $p_0\leq k^h$ and $p < k^h2^k/(k-1)$.
\end{lemma}

We remark that under the settings of rare features, some high-order interaction terms will become zero vectors, so the exact dimension of the TSLA problem, $p$, could be much smaller than $k^h2^k/(k-1)$.

Suppose that the original tree is a \emph{k-ary} tree of depth $h$, our main result is stated in Theorem~\ref{thm:1}. 

\begin{theorem}
\label{thm:1}
Assume that $\bvare\sim N_n(\0, \sigma^2\I_n)$.
Suppose $(\hat\bbeta,\ \hat\bgamma)$ is a solution to the optimization problem \eqref{eq:optimprob3}.
If we take {\cre $\lambda \geq 4\sigma\sqrt{2^k-1}\sqrt{\log (p \vee n)/n}$}, $\alpha\in[0,\ 1]$,
$\tilde{w}_j=\|\x_j\|/\sqrt{n}$ for $1\leq j\leq p$ and
$w_g=\sqrt{p_g/(2^k-1)}$ for $g\in\mathcal{G}$, then
with probability at least $1-1/n-2/(pn)$, it holds that
\begin{align}
    \frac{1}{n}\|\X\hat\bbeta-\X\bbeta^*\|^2\preceq
    \sigma\sqrt{2^k-1}\sqrt{\frac{\log {\cre (p \vee n)}}{n}}  \Omega(\bbeta^*,\bgamma^*; \alpha),
\end{align}
where $p \vee n = \max(p,n)$, $\Omega(\bbeta^*,\bgamma^*; \alpha)$ is the penalty function in \eqref{eq:penalty-supp} evaluated at the true coefficient vector $\bbeta^*$ and $\bgamma^*=\mathop{\arg\min}\limits_{\gamma: \A\gamma=\beta^*}\{P_T(\bgamma)\}$. 
The symbol $\preceq$ means that the inequality holds up to a multiplicative constant that is independent of $\sigma$, $p$, $n$, and $k$. 
\end{theorem}

Based on the proof in Section \ref{sec::app-thm-thm1} below, when $\lambda = 4\sigma\sqrt{2^k-1}\sqrt{\log (p \vee n)/n}$, the multiplicative constant can be set as $16$. Not surprisingly, the result reveals that the difficulty of the TSLA problem depends on the feature hierarchy mainly through the number of the original features ($p_0$), the depth of the tree ($h$), and the maximum number of child nodes ($k$).  
The term $\Omega(\bbeta^*,\bgamma^*; \alpha)$ measures the minimal penalty function evaluated at the truth, which is a natural measure of the complexity of the true model under both zero-sparsity and equi-sparsity.

To further understand the complexity $\Omega(\bbeta^*,\bgamma^*; \alpha)$ of the TSLA method, we introduce the following definition and results regarding the coarsest aggregation tree from \citet{yan2021rare}.

\begin{definition}
Denote $V(\T)$ as the set of nodes in tree $\T$. The set of nodes $\mathcal{B}\subseteq V(\T)$ is an aggregation set
with respect to tree $\T$ if
$\{L(u): u\in \mathcal{B}\}$ forms a partition of $L(\T)$.
\end{definition}

\begin{lemma}{\citep{yan2021rare}}
\label{lemma:lmjacob}
For any $\bbeta^*\in \mathbb{R}^p$, there exists a unique coarsest aggregation set $\mathcal{B}^*$
with respect to tree $\T$ such that
(a) $\beta_j^*=\beta_k^*$ for $j,k\in L(u),\ \forall u\in \mathcal{B}^*$,
(b) $|\beta_j^*-\beta_k^*|>0$, $j\in L(u)$ and $k\in L(v)$ for
siblings $u,v\in \mathcal{B}^*$.
\end{lemma}

Lemma~\ref{lemma:lmjacob} indicates that there is a unique coarsest ``true aggregation set'' $\mathcal{B}^*$ corresponding to the true model coefficient $\bbeta^*$. Based on this coarsest aggregation set, we can further quantify $\Omega(\bbeta^*,\bgamma^*; \alpha)$ in terms of $\|\bbeta^*\|_1$, which is stated in the next corollary.

\begin{corollary}
\label{coro:1}
With the same settings as in Theorem~\ref{thm:1} and taking {\cre $\lambda \geq 4\sigma\sqrt{2^k-1}\sqrt{\log (p \vee n)/n}$}, it holds with probability at least $1-1/n-2/(pn)$ that 
\begin{align*}
    \frac{1}{n}\|\X\hat\bbeta-\X\bbeta^*\| ^2\preceq
    \sigma\sqrt{2^k-1}\sqrt{\frac{\log {\cre(p \vee n)}}{n}}\|\bbeta^*\|_1.
\end{align*}

\end{corollary}

Recall that for lasso method, under the least square problem, the theoretical prediction error bound is $\sigma\sqrt{\log p/n}\|\bbeta^*\|_1$~\citep{buhlmann2011statistics}. This shows that the prediction error bound of TSLA is at least comparable to that of the lasso when $k$ is small or considered as a constant {\cre under the high-dimensional scenario}.

The following two corollaries show that it is possible to improve the error bound through taking into account the rarity of the features and the equi-sparsity of the true model. 

\begin{corollary}
\label{coro:2}
Assume the original tree is a perfect full $k$-ary tree of depth $h$ with $p_0$ leaf nodes, and the original features can be aggregated to nodes at depth $h_0$ along the tree. With the same settings as in Theorem~\ref{thm:1}, and taking{\cre $\lambda \geq 4\sigma\sqrt{2^k-1}\sqrt{\log (p \vee n)/n}$}, it holds with probability at least $1-1/n-2/(pn)$ that
\begin{align*}
    \frac{1}{n}\|\X\hat\bbeta-\X\bbeta^*\|^2\preceq \sigma\sqrt{2^k-1}\sqrt{\frac{\log {\cre(p \vee n)}}{n}} \left( \|\bbeta^*\|_1 - \alpha (1-f(h_0; p_0, h)) \|\bbeta_2^*\|_1 \right), 
\end{align*}
where $\bbeta_2^*$ is a sub-vector of $\bbeta^*$ that collects coefficients of the aggregated features, and $0<f(h_0;p_0,h)\leq 1$ is an increasing function of $h_0$, with $f(0;p_0, h)=1/p$, $f(h;p_0, h)=1$, and $f(h_0;p_0, h) = (2p_0^{1-h_0/h}-1)^{-1}$ when $0<h_0<h$. 
\end{corollary}

Corollary~\ref{coro:2} reveals that the potential gain of TSLA over the lasso in prediction depends on how much the features can be aggregated along the tree. Not surprisingly, the largest improvement is reached when $h_0=0$, that is, when the original features on the leaf nodes can be all aggregated into a single feature. 

\begin{corollary}
\label{coro:3}
Assume that the original features, $x_j,\ j=1, \ldots, p_0$, are independent binary random variables. We further assume:  $P(x_1 \lor x_2 \lor \cdots \lor x_{p_0} = 1) \leq a/k$, where $0<a<k/3$ is a positive constant. With $w_g=1$ for $g\in\mathcal{G}$, and the other settings keeping the same as in Theorem~\ref{thm:1}, it holds with probability at least $1- 1/n-2/(pn)-e^{3a}/n$ that $\sqrt{2^k-1}$ can be replaced with $\sqrt{e^a}$ in the prediction error bound.

\end{corollary}

Corollary \ref{coro:3} reveals that TSLA can still perform well when $k$ is large as long as the original features are sufficiently sparse; the key is to bound the magnitude of the high-order interaction terms that are needed for pursuing the potential ``or'' operations.

\subsection{Proofs of Theoretical Results}
\label{sec::app-thm}

\subsubsection{Proof of Lemma~\ref{lemma:k}}

Consider the case when the original tree $\T_0$ is a perfect full \emph{k-ary} tree of depth $h$,  in which each internal node has exactly $k$ child nodes and all the leaf nodes are at the same depth. Recall that $p_0$ is the number of leaf nodes in the original tree and $p$ is the number of leaf nodes in the expanded tree.

It follows that 
\begin{align*}
	p_0=k^h, \qquad |I(\T_0)| = \frac{p_0-1}{k-1} = \frac{k^h-1}{k-1},
\end{align*}
where $|I(\T_0)|$ is the total number of internal nodes in the original tree. 

After the tree-guided expansion, the number of internal nodes keeps unchanged, and each internal node can have up to $2^k-1$ child nodes. Assume the generated features by expansion are nonzero vectors. Then the number of total leaf nodes after expansion is  
\begin{align*}
	p &= p_0 + \frac{p_0 - 1}{k - 1} \times (2^k -k -1) \\
	    &= k^h \times (1 + \frac{2^k}{k - 1}) - \frac{2^k}{k - 1} -\frac{k +1}{k -1} \times (k^h - 1) \\
	    & = \frac{k^h2^k}{k - 1}  - k^h(\frac{k +1}{k -1} -1 ) - \frac{2^k-(k+1)}{k - 1}.
\end{align*}
When $k$ increases with fixed value of $h$, the original dimension $p_0$ grows at rate $k^h$, while the expanded dimension $p$ grows at rate $\mathcal{O}(k^h2^k/(k-1))$. 

By the definition of the general \emph{k-ary} tree with depth $h$, the maximum number of leaf/internal nodes is attained when $\T_0$ is also a perfect full \emph{k-ary} tree. It immediately follows that $p_0\leq k^h$ and $p< k^h2^k/(k-1)$.

This completes the proof.

\subsubsection{Proof of Theorem~\ref{thm:1}}
\label{sec::app-thm-thm1}

Denote $\Omega(\bbeta,\bgamma;\alpha)=(1-\alpha)P_T(\bbeta)+
    \alpha P_T(\bgamma)$
as the penalty function in problem
\eqref{eq:optimprob3},  
where
$P_T(\bbeta)=\sum_{j=1}^p \tilde{w}_j|\beta_j|$ and
$P_T(\bgamma)=\sum_{g\in \mathcal{G}}w_g\|\bgamma_g\|$.
Here, we take weights $\tilde{w}_j=\|\x_j\|/\sqrt{n}$ and $w_g=\sqrt{p_g/(2^k-1)}$,
where $p_g$ is the number of elements in group $g$, $\x_j$ is the $j^{th}$ column of $\X$, and
$k$ is the maximum number of child nodes in the original tree.
Let $(\hat{\bbeta}, \hat{\bgamma})$ be a solution to the problem~\eqref{eq:optimprob3}, we have
\begin{align*}
    \frac{1}{2n}\|\y-\X\hat\bbeta\|^2+\lambda\Omega(\hat\bbeta,\hat\bgamma; \alpha)
   \ \leq\
    \frac{1}{2n}\|\y-\X\bbeta\|^2+\lambda\Omega(\bbeta, \bgamma; \alpha),
\end{align*}
for any $(\bbeta,\ \bgamma)$ such that $\bbeta=\A\bgamma$.
Assume ($\bbeta^*, \bgamma^*$) are the true coefficient vectors satisfying $\bbeta^*=\A\bgamma^*$.
Let $\hat\Delta^{\gamma}=\hat\bgamma-\bgamma^*$ and $\hat\Delta^{\beta}=\hat\bbeta-\bbeta^*$. By plugging in $\y=\X\bbeta^*+\bvare$ and $(\bbeta,\ \bgamma)=(\bbeta^*,\ \bgamma^*)$, with some algebra, we have
\begin{align}
    \label{eq:proof1}
    \frac{1}{2n}\|\X\hat\bbeta-\X\bbeta^*\|^2+
    \lambda\Omega(\hat\bbeta, \hat\bgamma; \alpha)
    \ \leq\ \lambda\Omega(\bbeta^*, \bgamma^*; \alpha)+
    \frac{1}{n}\bvare^{\trans}\X\hat\Delta^{\beta}.
\end{align}

For each group $g\in \mathcal{G}$, define {\cre${\bf P}_g \in \{0, 1\}^{|\T|\times|\T|}$} as a diagonal matrix.
The diagonal elements of ${\bf P}_g$ are given by
\begin{align*}
    ({\bf P}_g)_{uu}=
    \begin{cases}
      1, & \mbox{if node}\ u \in g; \\
      0, & \mbox{otherwise}.
    \end{cases}
\end{align*}
Then the \emph{Child-$\ell_2$} penalty $P_T(\bgamma)$ can be re-expressed as
\begin{align*}
    P_T(\bgamma)=\sum_{g\in \mathcal{G}}w_g\|\bgamma_g\|=
    \sum_{g\in \mathcal{G}}w_g\|{\bf P}_g\bgamma\|.
\end{align*}
Based on the structure of matrix ${\bf P}_g$, we also
have the following identity:
\begin{align*}
     \sum_{g\in \mathcal{G}}{\bf P}_g{\bf P}_g=\I_{|\T|}.
\end{align*}

Recall that from the tree-guided reparameterization, we have $\bbeta^*=\A\bgamma^*$ and
$\hat{\bbeta}=\A\hat{\bgamma}$. Consider the second term on the RHS of \eqref{eq:proof1}, we have
\begin{align}
        |\frac{1}{n}\bvare^{\trans}\X\hat\Delta^{\beta}|
        &=|(1-\alpha)\frac{1}{n}\bvare^{\trans}\X\hat\Delta^{\beta}+
        \alpha\frac{1}{n}\bvare^{\trans}\X\A\hat\Delta^{\gamma}| \notag \\
        &\leq|(1-\alpha)\frac{1}{n}\bvare^{\trans}\X\hat\Delta^{\beta}|+
        |\alpha\frac{1}{n}\bvare^{\trans}\X\A\hat\Delta^{\gamma}|.
        \label{eq:error}
\end{align}

For the first part on the RHS of \eqref{eq:error}, let $v_j=n^{-1/2}\x_j^{\trans}\bvare$
and $\tilde{w}_j=\|\x_j\|/\sqrt{n}$, for $j=1, \ldots,p$, we have
\begin{align*}
        |\frac{1}{n}\bvare^{\trans}\X\hat\Delta^{\beta}|
        & = |\sum_{j=1}^p\frac{\x_j^{\trans}\bvare}{\sqrt{n}\|\x_j\|}\times
        \frac{\|\x_j\|}{\sqrt{n}}\hat\Delta_j^{\beta}|\\
        & = |\sum_{j=1}^p\frac{v_j}{\|\x_j\|}\tilde{w}_j\hat\Delta_j^{\beta}| \\
        & \leq \max_{1\leq j\leq p}\frac{|v_j|}{\|\x_j\|}(\sum_{j=1}^p\tilde{w}_j|\hat\Delta_j^{\beta}|)\\
        & \leq \max_{1\leq j\leq p}\frac{|v_j|}{\|\x_j\|}(\sum_{j=1}^p\tilde{w}_j|\hat\beta_j|
        + \tilde{w}_j|\beta_j^*|) \\
        & = \max_{1\leq j\leq p}\frac{|v_j|}{\|\x_j\|}(P_T(\bbeta^*)+P_T(\hat\bbeta)).
\end{align*}

Since $\bvare\sim N_n(\0,\sigma^2\I_n)$, $v_j\sim N(0, \sigma^2\|\x_j\|^2/n)$ for $j=1,\ldots,p$. By Lemma 6.2 of \citet{buhlmann2011statistics}, for $t>0$,
it holds that
\begin{align*}
    P \left(\max_{1\leq j\leq p}\frac{|v_j|}{\|\x_j\|}>
    \sigma\sqrt{\frac{2(t+\log p)}{n}}\ \right)\ \leq\ 2e^{-t}.
\end{align*}
{\cre Take $t=\log p + \log n>0$, we get
\begin{align}
     P \left(\ \max_{1 \leq j\leq p}\frac{|v_j|}{\|\x_j\|}>
     \sigma\sqrt{\frac{2(2\log p + \log n)}{n}}\ \right)\ \leq\ \frac{2}{pn}\ .
     \label{eq:bound1}
\end{align}}

For the second part on the RHS of \eqref{eq:error}, using the property of the matrix ${\bf P}_g$, we get
\begin{align*}
        |\frac{1}{n}\bvare^{\trans}\X\A\hat\Delta^{\gamma}|
    & = |\frac{1}{n}\bvare^{\trans}\X\A
    (\sum_{g\in \mathcal{G}}{\bf P}_g{\bf P}_g)\hat\Delta^{\gamma}|\\
    & = |\frac{1}{n}\sum_{g\in \mathcal{G}}(\bvare^{\trans}\X\A{\bf P}_g)
    ({\bf P}_g\hat\Delta^{\gamma})| \\
    & \leq \frac{1}{n}\sum_{g\in \mathcal{G}}\sqrt{\frac{2^k-1}{p_g}} \|\bvare^{\trans}\X\A {\bf P}_g \|
    \times\sqrt{\frac{p_g}{2^k-1}}\|{\bf P}_g\hat\Delta^{\gamma}\| \\
    & \leq \max_{g\in \mathcal{G}} \{\frac{1}{n}\sqrt{\frac{2^k-1}{p_g}}\|\bvare^{\trans}\X\A{\bf P}_g\|\}
    \sum_{g\in \mathcal{G}}\sqrt{\frac{p_g}{2^k-1}}\|{\bf P}_g\hat\Delta^{\gamma}\| \\
    & \leq \max_{g\in \mathcal{G}} \{\frac{1}{n}\sqrt{\frac{2^k-1}{p_g}}\|\bvare^{\trans}\X\A{\bf P}_g\|\}
    (\sum_{g\in \mathcal{G}}\sqrt{\frac{p_g}{2^k-1}}\|{\bf P}_g\gamma^*\|+\sum_{g\in \mathcal{G}}\sqrt{\frac{p_g}{2^k-1}}\|{\bf P}_g\hat\gamma\|) \\
    & = \max_{g\in \mathcal{G}} \{\frac{1}{n}\sqrt{\frac{2^k-1}{p_g}}\|\bvare^{\trans}\X\A{\bf P}_g\|\}
    (P_T(\bgamma^*)+P_T(\hat\bgamma)).
\end{align*}

To bound this second part, we use the following Lemma from \citet{hsu2012tail}.
\begin{lemma}
\label{lemma:lm1}
Let $\Z$ be an $m\times n$ matrix and let $\bSigma_z=\Z^{\trans}\Z$.
Suppose $\bvare\sim\ N_n(\0,\sigma^2\I_n)$.
For all $t>0$,
\begin{align*}
    P\left(\ \|\Z\bvare\|^2>\sigma^2(\,\tr(\bSigma_z)+2\sqrt{\tr(\bSigma_z^2)t}+2\|\bSigma_z\|t\,)\ \right)\ <\ e^{-t},
\end{align*}
where $\|\bSigma\|$ denotes the matrix spectral norm.
\end{lemma}

For each $g\in \mathcal{G}$, denote {\cre$\X_g=\X\A{\bf P}_g\in \{0, 1\}^{n\times |\T|}$},
and take $\bSigma_g=\X_g\X_g^{\trans}$.
It is not hard to see that each column of $\X_g$ is given by aggregating columns in $\X$ via the ``or'' operation.
Then the element in $\X_g$ is either 0 or 1.
For example, in Figure~2 of the main paper, when $u=u_1^0$ and $g=\{u_1^1,u_2^1,u_{(12)}^1\}$, the $\X_g$ matrix may have three nonzero columns, which are constructed with
$x_1\ \lor\ x_2\ \lor\ x_3$, $x_4\ \lor\ x_5$, and $(x_1\ \lor\ x_2\ \lor\ x_3)\ \wedge\ (x_4\ \lor\ x_5)$,  respectively. In fact, the $\X_g$ matrix always has at most $p_g$ nonzero columns.
Denote $\X_g$ as $\X_g=\{x_{ij}^g\}_{n\times |\T|}$, it follows that
\begin{align}
        & \tr(\bSigma_g) = \tr(\X_g\X_g^{\trans}) = \sum_{i=1}^n\sum_{j=1}^{|\T|}(x_{ij}^g)^2
         \leq np_g1^2=np_g, \label{eq:xgbound} \\
        & \tr(\bSigma_g^2)\leq \tr(\bSigma_g)^2\leq (np_g)^2, \notag \\
        & \|\bSigma_g\| \leq \tr(\bSigma_g) \leq np_g. \notag 
\end{align}
By Lemma~\ref{lemma:lm1}, for any $g\in \mathcal{G}$ and any $t>0$,
\begin{align*}
    & P\left(\|\X_g^{\trans}\bvare\|^2> np_g\sigma^2(1+2\sqrt{t}+2t)\ \right)\ <\ e^{-t}  \\
         \Longrightarrow
    & P\left(\frac{1}{\sqrt{n}}\|\X_g^{\trans}\bvare\|>
         \sigma\sqrt{p_g}\sqrt{1+2\sqrt{t}+2t}\ \right)\ <\ e^{-t} \\
         \Longrightarrow
    & P\left(\frac{\sqrt{2^k-1}}{n\sqrt{p_g}}\|\X_g^{\trans}\bvare\|>
         2\sqrt{2(2^k-1)}\sigma\sqrt{\frac{t}{n}}\ \right)\ <\ e^{-t},
         \ \mbox{for}\ t>\frac{1}{2}.
\end{align*}
After taking a union bound over all $g\in \mathcal{G}$, it follows that
\begin{align*}
    & P\left( \max_{g\in \mathcal{G}}\frac{1}{n}\sqrt{\frac{2^k-1}{p_g}}\|\X_g^{\trans}\bvare\|> 2\sqrt{2(2^k-1)}\sigma\sqrt{\frac{t}{n}}\ \right) < |\mathcal{G}|e^{-t}, \ \mbox{for}\ t>\frac{1}{2}.
\end{align*}
Since the expanded tree $\T$ has $p$ leaf nodes, there are at most $p-1$ internal nodes. Then we must have $|\mathcal{G}|\leq p$.
{\cre Take $t=\log p + \log n>1/2$, we get
\begin{align}
        &P\left(\max_{g\in \mathcal{G}}\frac{1}{n}\sqrt{\frac{2^k-1}{p_g}}\|\X_g^{\trans}\bvare\|> 2\sqrt{2(2^k-1)}\sigma\sqrt{\frac{\log p + \log n}{n}}\ \right) < \frac{|\mathcal{G}|}{pn}
        \notag \\
        \Longrightarrow
        &P\left(\max_{g\in \mathcal{G}}\frac{1}{n}\sqrt{\frac{2^k-1}{p_g}}\|\X_g^{\trans}\bvare\|> 2\sqrt{2(2^k-1)}\sigma\sqrt{\frac{\log p + \log n}{n}}\ \right) < \frac{1}{n}.
        \label{eq:bound2}
\end{align}

Combine the two inequalities \eqref{eq:bound1} and \eqref{eq:bound2} by taking a union bound, we get
{\footnotesize
\begin{align*}
    & P\left(\ \max_{g\in \mathcal{G}}\frac{1}{n}\sqrt{\frac{2^k-1}{p_g}}\|\X_g^{\trans}\bvare\|>
    2\sqrt{2(2^k-1)}\sigma \sqrt{\frac{\log p + \log n}{n}}\ \mbox{or}\
    \max_{1\leq j\leq p}\frac{|v_j|}{\|\x_j\|}>
      \sigma\sqrt{\frac{4\log p + 2\log n}{n}}\ \right)
  <\  \frac{1}{n}+\frac{2}{pn}.
\end{align*}
}
By taking $\lambda \geq 4\sigma\sqrt{2^k-1}\sqrt{\log (p \vee n)/n}$,
with probability at least $1-1/n-2/(pn)$,
we have
\begin{align*}
        \frac{1}{2n}\|\X\hat\bbeta-\X\bbeta^*\|^2
    \  \leq\ & \lambda\Omega(\bbeta^*,\bgamma^*; \alpha) - \lambda\Omega(\hat\bbeta,\hat\bgamma; \alpha) +
    \frac{1}{n}\bvare^{\trans}\X\hat\Delta^{\beta}\\
     \leq\ & \lambda\Omega(\bbeta^*,\bgamma^*; \alpha)-\lambda\Omega(\hat\bbeta,\hat\bgamma; \alpha)
     +
      \alpha\lambda(P_T(\bgamma^*)+P_T(\hat\bgamma))\\
  &+(1-\alpha)\lambda(P_T(\bbeta^*)+P_T(\hat\bbeta))\\
   =\ & 2\lambda\Omega(\bbeta^*,\bgamma^*; \alpha), 
\end{align*}
where $p \vee n=\max(p,n)$. The above holds for any $\bgamma^*$ such that $\bbeta^*=\A\bgamma^*$. 
By setting $\lambda = 4\sigma\sqrt{2^k-1}\sqrt{\log (p \vee n)/n}$, we can more explicitly formulate the error bound as 
\begin{align*}
        \frac{1}{n}\|\X\hat\bbeta-\X\bbeta^*\|^2
    \  \leq \ & 16 \sigma\sqrt{2^k-1}\sqrt{\frac{\log (p \vee n)}{n}}\Omega(\bbeta^*,\bgamma^*; \alpha).
\end{align*}
}
This completes the proof. 

\subsubsection{Proof of Corollary~\ref{coro:1}}\label{sec::app-thm-coro1}

Following the proof of Theorem~\ref{thm:1}, for the penalty function part, we have
\begin{align*}
    \Omega(\bbeta,\bgamma; \alpha)=(1-\alpha)\sum_{j=1}^p\tilde{w}_j|\beta_j|+
    \alpha\sum_{g\in \mathcal{G}}w_g\|\bgamma_g\|,\ \mathcal{G}=
    \{u_1^0\}\cup\{C(u),\ u\in I(\T)\}.
\end{align*}
With $\x_j$ being the $j^{th}$ column of the expanded sparse binary matrix $\X$, we have $\tilde{w}_j=\|\x_j\|/\sqrt{n}\leq\sqrt{n/n}\leq 1$.
Under the condition that the expanded tree $\T$ is at most a \emph{$(2^k-1)$-ary} tree,
it always holds that $w_g=\sqrt{p_g/(2^k-1)} \leq \sqrt{(2^k-1)/(2^k-1)}=1$.
Then we have
\begin{align*}
    \Omega(\bbeta,\bgamma; \alpha) \leq (1-\alpha)\|\bbeta\|_1+
    \alpha\sum_{g\in \mathcal{G}}\|\bgamma_g\|.
\end{align*}

Since there is no overlap of $\bgamma^*$ between different groups in $\mathcal{G}$, we immediately have 
\begin{align*}
    \sum_{g\in \mathcal{G}}\|\bgamma_g^*\|
    \leq  \sum_{g\in \mathcal{G}}\|\bgamma_g^*\|_1 = \|\bgamma^*\|_1.
\end{align*}
Let $\mathcal{B}^*$ be the unique coarsest aggregation set with respect to the true coefficient $\bbeta^*$ under the given tree $\T$. Based on the construction of $\mathcal{B}^*$ described in Lemma~\ref{lemma:lmjacob} , there exists $\bgamma^*$, such that $\bbeta^*=\A\bgamma^*$, and 
\begin{align*}
	\gamma_u^*&=\beta_j^*, \forall u \in \mathcal{B}^* \cap L(\T),  \beta_j^* \mbox{ is the regression coefficient at node } u, \\
	\gamma_u^*&=\beta_j^*, \forall j \in L(u) \mbox{ and } u \in \mathcal{B}^* \cap I(\T),  \\
	\gamma_u^*&=0, \mbox{ otherwise.}
\end{align*}
As such, immediately we have 
\begin{align*}
	\|\bgamma^*\|_1\ \leq\ \|\bbeta^*\|_1,
\end{align*}
where the equality holds when all the nodes in $\mathcal{B}^*$ are the leaf nodes, i.e., there is no aggregation pattern among the binary features. 

Then we have
\begin{align*}
    \Omega(\bbeta^*,\bgamma^*; \alpha)
    \leq (1-\alpha)\|\bbeta^*\|_1+\alpha\|\bbeta^*\|_1
    = \|\bbeta^*\|_1.
\end{align*}
This completes the proof.

\subsubsection{Proof of Corollary~\ref{coro:2}}
\label{sec::app-coro2}

Assume the original tree is a perfect full \emph{k-ary} tree of depth $h$ and has $p_0$ leaf nodes. 
Let $I(h_0)$ denotes the set of internal nodes at depth $h_0$ of the expanded tree. By the property of the perfect full \emph{k-ary} tree, there are exactly $k^{h_0}$ nodes in $I(h_0)$. For each node $u^{h_0}\in I(h_0)$, the number of leaf nodes in the sub-tree led by $u^{h_0}$ is the same. This value can be expressed as
\begin{align*}
    |L(u^{h_0})| = k^{h-h_0} + (2^k-k-1)\frac{k^{h-h_0}-1}{k-1}.
\end{align*}

Based on the assumption that all the features can be aggregated to the nodes at depth $h_0$, we decompose the elements in the true regression coefficient vector $\bbeta^*\in\mathbb R^{p}$ into two parts by the depth of the corresponding leaf nodes.
Let $\bbeta_1^*\in\mathbb{R}^{p_1}$ and $\bbeta_2^*\in\mathbb{R}^{p_2}$ be two sub-vectors of $\bbeta^*$, where
\begin{align*}
    \bbeta_1^* = (\beta_{11}^*, \cdots, \beta_{1j}^*, \cdots, \beta_{1p_1}^*)^{\trans}, \mbox{ depth}(u_{1j}) \leq h_0,
\end{align*}
and
\begin{align*}
    \bbeta_2^* = (\beta_{21}^*, \cdots, \beta_{2j}^*, \cdots, \beta_{2p_2}^*)^{\trans}, \mbox{ depth}(u_{2j}) > h_0,
\end{align*}
with $u_{ij},i=1,2,$ being the corresponding leaf nodes for the elements in $\bbeta^*$. 
Then we have $\bbeta^*=(\bbeta_1^*^{\trans},\ \bbeta_2^*^{\trans})^{\trans}$. 

Building on the perfect full \emph{k-ary} tree, we have 
\begin{align*}
    p_1 = (2^k-k-1)\frac{k^{h_0}-1}{k-1},\ p_2 = k^{h_0}|L(u^{h_0})| = k^h + (2^k-k-1)\frac{k^h-k^{h_0}}{k-1}.
\end{align*}
It is then clear that more elements will be collected in $\bbeta_2^*$ when $h_0$ decreases, that is, when the original features can be aggregated into fewer but denser features. 

Based on the discussion of the coarsest aggregation tree in Section~\ref{sec::app-thm-coro1}, when all features are aggregated to nodes at depth $h_0$, for the $\bgamma^*$ vector we have 
\begin{align*}
	\gamma_u^*&=\beta_j^* \in \bbeta_1^*,\ \forall u \in L(\T),\  \mbox{ depth}(u) \leq h_0,\  
	\beta_j^* \mbox{ is the regression coefficient at node } u, \\
	\gamma_u^*&=\beta_j^* \in \bbeta_2^*,\ \forall u \in I(h_0), \ \forall j \in L(u), \\
	\gamma_u^*&=0, \mbox{ otherwise,}
\end{align*}
where $L(\T)$ is the set of leaf nodes in the expanded tree $\T$.
Note that, there are $k^{h_0}$ distinct $\gamma_u^*$ values in the second case, and for each distinct $\gamma_u^*$, there will be $|L(u^{h_0})|$ elements in $\bbeta_2^*$ that have the same value.
With the same settings as in Section~\ref{sec::app-thm-coro1}, we have
\begin{align}
    \sum_{g\in \mathcal{G}}w_g\|\bgamma_g^*\| & \leq  \|\bgamma^*\|_1 = \|\bbeta_1^*\|_1 + \frac{\|\bbeta_2^*\|_1}{|L(u^{h_0})|},
    \label{eq:aggr1}
\end{align}
where $\|\bbeta_1^*\|_1+\|\bbeta_2^*\|_1=\|\bbeta^*\|_1$.
It is not hard to see that $|L(u^{h_0})|$ is a decreasing function of $h_0$ when $p_0$ and $h$ are held as fixed, with $|L(u^{0})|=p$ and $|L(u^{h})|=1$. Since $2^k-k-1 \geq k-1$ when $k \geq 2$, it always holds that $|L(u^{h_0})| > 2k^{h-h_0}-1$. Further, we have $p_0=k^h$ in the perfect full \emph{k-ary} tree. Thus, we can replace $|L(u^{h_0})|$ with a much simpler lower bound $2p_0^{1-h_0/h}-1$ when $0<h_0<h$.  

Combine \eqref{eq:aggr1} in $\Omega(\bbeta^*, \bgamma^*;\alpha)$, and follow the discussions in Section~\ref{sec::app-thm-coro1}, 
we get 
\begin{align*}
    \Omega(\bbeta^*, \bgamma^*;\alpha) &\leq (1-\alpha)\|\bbeta^*\|_1 + \alpha (\|\bbeta_1^*\|_1 + f(h_0;p_0, h)\|\bbeta_2^*\|_1)  \\
    & \leq \|\bbeta^*\|_1 - \alpha (1- f(h_0;p_0, h))\|\bbeta_2^*\|_1,
\end{align*}
where $f(0;p_0, h)=1/p$, $f(h;p_0, h)=1$, and $f(h_0;p_0, h) = (2p_0^{1-h_0/h}-1)^{-1}$ when $0<h_0<h$.  The results then follow immediately by combining with Theorem~\ref{thm:1}.

\subsubsection{Proof of Corollary~\ref{coro:3}}
\label{sec::app-lemma-factor}

 We start the proof by finding a tighter upper bound for $\max_{g\in \mathcal G}\tr(\X_g\X_g^{\trans})$ in \eqref{eq:xgbound} in the proof of Theorem~\ref{thm:1}.
Note that, $\tr(\X_g\X_g^{\trans})$ equals to the number of nonzero elements in the matrix $\X_g$. 

For the node group $g$ corresponding to matrix $\bX_g$, it has at most $k$ native nodes. Denote the feature aggregated at each native node as $\tilde x_j$ for $j=1,\ldots, k$.  These features can be constructed by conducting the ``or'' operation on non-overlapped groups of original leaf node features. Based on the assumptions we have made, $\tilde x_j$ can be considered as independent binary random variables with $r=P(\tilde x_j=1) \leq a/k$ for $j=1,\ldots, k$.  

Let $z_i$ be the count of nonzero elements of the $i^{th}$ row in matrix $\X_g$. Then $\tr(\X_g\X_g^{\trans})=\sum_{i=1}^n z_i$. For each row, it has at most $2^k-1$ nonzero elements, in which $k$ elements are associated with the main effects, and the rest of the elements are generated from interactions among $\tilde x_j$. Consider the values of $z_i$ when we have exactly $0, 1, 2, \ldots, k$ out of $k$ random variables taking value 1, it is not hard to see that $z_i$ is a nonnegative discrete random variable which can only takes value in $\{0, 2^1-1, 2^2-1, 2^3-1, \ldots, 2^k-1\}$ with probability
\begin{align*}
	P(z_i = 2^j-1) = {k \choose j}r^j(1-r)^{k-j},\ j = 0, 1, \cdots, k.
\end{align*}
It then follows that 
\begin{align*}
	E(z_i) & = {k\choose 1}r^1(1-r)^{k-1}2 + {k\choose 2}r^2(1-r)^{k-2}2^2 + 
	\cdots {k\choose k}r^k 2^k - \sum_{j=1}^k{k\choose j}r^j(1-r)^{k-j} \\  
	& = \sum_{j=1}^k{k\choose j}(2r)^j(1-r)^{k-j} -  \sum_{j=1}^k{k\choose j}r^j(1-r)^{k-j} \\
	& = (1-r+2r)^k -  {k\choose 0}(1-r)^k - (1 - {k\choose 0}(1-r)^k) \\ 
	& =  (1+r)^k -  1.
\end{align*}
Similarly, we can get
\begin{align*}
	E(z_i^2) & \leq  {k\choose 0}r^0(1-r)^{k} + {k\choose 1}r^1(1-r)^{k-1}4 + {k\choose 2}r^2(1-r)^{k-2}4^2 + 
	\cdots {k\choose k}r^k(1-r)^{k-k} 4^k\\ 
	&  = (1 - r + 4r)^k \\ 
	& =  (1 + 3r)^k. 
\end{align*}
It then follows that $\var(z_i)\leq (1 + 3r)^k$. 

Since $z_i$ are independent, when the sample size $n$ is finite, by the Chebyshev inequality, we have
\begin{align*}
	P(|\sum_{i=1}^n z_i - E(\sum_{i=1}^n z_i)| \geq t) \leq \frac{\var(\sum_{i=1}^n z_i)}{t^2} \leq \frac{n(1 + 3r)^k}{t^2}, \ \forall t>0.
\end{align*}
By taking $t=n$, it follows that
\begin{align*}
	P(|\sum_{i=1}^n z_i - E(\sum_{i=1}^n z_i)| \geq n) \leq \frac{n(1 + 3r)^k}{n^2} = \frac{(1 + 3r)^k}{n}.
\end{align*}
Since we assume $r\leq a/k$, it always holds that $(1+a/k)^k\leq e^a$ for $k>0$ and $0<a<k$ as a constant. Finally, we get
\begin{align*}
	P(|\sum_{i=1}^n z_i - E(\sum_{i=1}^n z_i)| \geq n) \leq \frac{e^{3a}}{n}.
\end{align*}

Further, by the triangle inequality, we have
\begin{align}
	\sum_{i=1}^n z_i \leq |\sum_{i=1}^n z_i - \sum_{i=1}^n E(z_i)| + |\sum_{i=1}^n E(z_i)|  =|\sum_{i=1}^n z_i - \sum_{i=1}^n E(z_i)| + n(1+r)^k -  n.
\end{align}
Then with probability at least $1- e^{3a}/n$, 
\begin{align}
	\max_{g\in \mathcal G}\tr(\X_g\X_g^{\trans}) = \sum_{i=1}^n z_i \leq n + n(1+r)^k -  n \leq ne^a.
\end{align}
The rest of the proof follows the discussions in Section~\ref{sec::app-thm-thm1}. By changing the weights $w_g=1$ for all $g$, and taking {\cre $\lambda \geq 4\sigma\sqrt{2^k-1}\sqrt{\log (p \vee n)/n}$}, we have 
\begin{align*}
	 \frac{1}{n}\|\X\hat\bbeta-\X\bbeta^*\|^2\preceq
    \lambda  \Omega(\bbeta^*,\bgamma^*; \alpha),
\end{align*}
with probability at least {\cre $1-1/n-2/(pn)-e^{3a}/n$}.

This completes the proof.

\clearpage

\section{Supplemental Materials for Simulation Studies}
\label{sec::app-simu}

\subsection{Tree Structures Used in Section~4.1 and 4.2 of the Main Paper}
\label{sec::app-simu-tree}

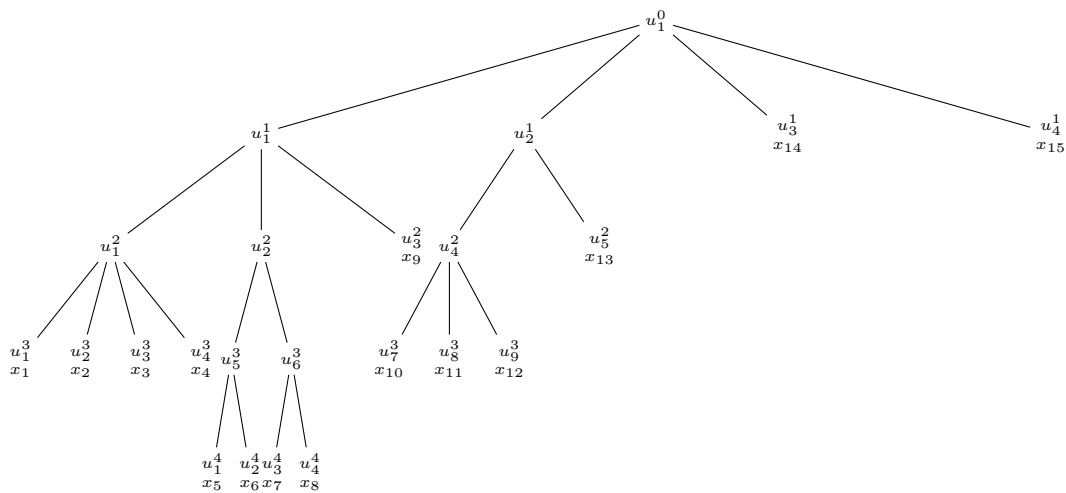
\begin{figure}[htp]
        \centering
        \begin{tikzpicture}
        \tiny
            \tikzstyle{level 1}=[sibling distance=35mm]
            \tikzstyle{level 2}=[sibling distance=20mm]
            \tikzstyle{level 3}=[sibling distance=8mm]
            \tikzstyle{level 4}=[sibling distance=5mm]
            \node {$u_1^0$} [align=center]
                child {node {$u_1^1$}
                    child {node {$u_1^2$}
                        child {node {$u_1^3$ \\ $x_1$}}
                        child {node {$u_2^3$ \\ $x_2$}}
                        child {node {$u_3^3$ \\ $x_3$}}
                        child {node {$u_4^3$ \\ $x_4$}}}
                    child {node {$u_2^2$}
                        child {node {$u_5^3$}
                            child {node {$u_1^4$ \\ $x_5$}}
                            child {node {$u_2^4$ \\ $x_6$}}}
                        child {node {$u_6^3$}
                            child {node {$u_3^4$ \\ $x_7$}}
                            child {node {$u_4^4$ \\ $x_8$}}}}
                    child {node {$u_3^2$ \\ $x_9$}}}
                child {node {$u_2^1$}
                    child {node {$u_4^2$}
                        child {node {$u_7^3$ \\ $x_{10}$}}
                        child {node {$u_8^3$ \\ $x_{11}$}}
                        child {node {$u_9^3$ \\ $x_{12}$}}}
                    child {node {$u_5^2$ \\ $x_{13}$}}}
                child {node {$u_3^1$ \\ $x_{14}$}}
                child {node {$u_4^1$ \\ $x_{15}$}};
    \end{tikzpicture}
    \caption{Tree structure 1.}
    \label{fig:fig6}
\end{figure}

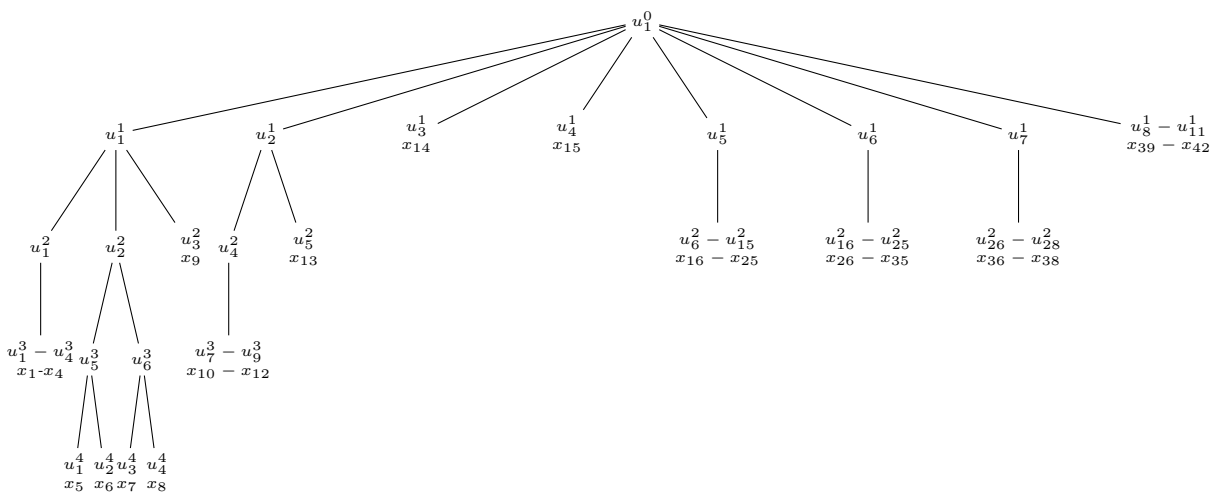
\begin{figure}[htp]
        \centering
        \begin{tikzpicture}
        \tiny
            \tikzstyle{level 1}=[sibling distance=20mm]
            \tikzstyle{level 2}=[sibling distance=10mm]
            \tikzstyle{level 3}=[sibling distance=7mm]
            \tikzstyle{level 4}=[sibling distance=4mm]
            \node {$u_1^0$} [align=center]
                child {node {$u_1^1$}
                    child {node {$u_1^2$}
                        child {node {$u_1^3-u_4^3$ \\ $x_1$-$x_4$}}}
                    child {node {$u_2^2$}
                        child {node {$u_5^3$}
                            child {node {$u_1^4$ \\ $x_5$}}
                            child {node {$u_2^4$ \\ $x_6$}}}
                        child {node {$u_6^3$}
                            child {node {$u_3^4$ \\ $x_7$}}
                            child {node {$u_4^4$ \\ $x_8$}}}}
                    child {node {$u_3^2$ \\ $x_9$}}}
                child {node {$u_2^1$}
                    child {node {$u_4^2$}
                        child {node {$u_7^3-u_9^3$ \\ $x_{10}-x_{12}$}}}
                    child {node {$u_5^2$ \\ $x_{13}$}}}
                child {node {$u_3^1$ \\ $x_{14}$}}
                child {node {$u_4^1$ \\ $x_{15}$}}
                child {node {$u_5^1$}
                    child {node {$u_6^2-u_{15}^2$ \\ $x_{16}-x_{25}$}}}
                child {node {$u_6^1$}
                    child {node {$u_{16}^2-u_{25}^2$ \\ $x_{26}-x_{35}$}}}
                child {node {$u_7^1$}
                    child {node {$u_{26}^2-u_{28}^2$ \\ $x_{36}-x_{38}$}}}
                child {node {$u_{8}^1-u_{11}^1$ \\ $x_{39}-x_{42}$}};
    \end{tikzpicture}
    \caption{Tree structure 2.}
    \label{fig:fig7}
\end{figure}

\begin{figure}[htp]
        \centering
        \begin{tikzpicture}
        \tiny
            \tikzstyle{level 1}=[sibling distance=17mm]
            \tikzstyle{level 2}=[sibling distance=9.5mm]
            \tikzstyle{level 3}=[sibling distance=7mm]
            \tikzstyle{level 4}=[sibling distance=4mm]
            \node {$u_1^0$} [align=center]
                child {node {$u_1^1$}
                    child {node {$u_1^2$}
                        child {node {$u_1^3-u_4^3$ \\ $x_1$-$x_4$}}}
                    child {node {$u_2^2$}
                        child {node {$u_5^3$}
                            child {node {$u_1^4$ \\ $x_5$}}
                            child {node {$u_2^4$ \\ $x_6$}}}
                        child {node {$u_6^3$}
                            child {node {$u_3^4$ \\ $x_7$}}
                            child {node {$u_4^4$ \\ $x_8$}}}}
                    child {node {$u_3^2$ \\ $x_9$}}}
                child {node {$u_2^1$}
                    child {node {$u_4^2$}
                        child {node {$u_7^3-u_9^3$ \\ $x_{10}-x_{12}$}}}
                    child {node {$u_5^2$ \\ $x_{13}$}}}
                child {node {$u_3^1$ \\ $x_{14}$}}
                child {node {$u_4^1$ \\ $x_{15}$}}
                child {node {$u_5^1$}
                    child {node {$u_6^2-u_7^2$ \\ $x_{16}-x_{17}$}}}
                child {node {$u_6^1$}
                    child {node {$u_8^2-u_9^2$ \\ $x_{18}-x_{19}$}}}
                child {node {$u_7^1$}
                        child {node {$u_{10}^2$}
                            child {node {$u_{10}^3-u_{11}^3$ \\ $x_{20}-x_{21}$}}}
                        child {node {$u_{11}^2$ \\ $x_{22}$}}}
                child {node {$u_8^1$}
                    child {node {$u_{12}^2-u_{21}^2$ \\ $x_{23}-x_{32}$}}}
                child {node {$u_9^1$}
                    child {node {$u_{22}^2-u_{31}^2$ \\ $x_{33}-x_{42}$}}}
                child {node {$u_{10}^1$ \\ $x_{43}$}};
    \end{tikzpicture}
    \caption{Tree structure 3.}
    \label{fig:fig8}
\end{figure}
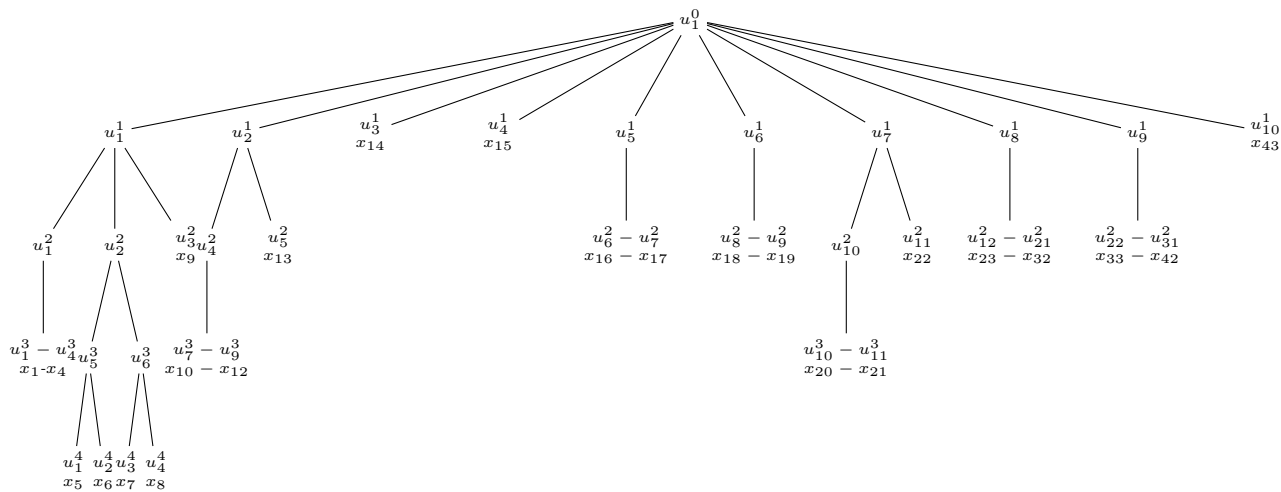

Figure \ref{fig:fig6}, Figure \ref{fig:fig7}, and Figure \ref{fig:fig8} show the original tree structures used in the simulation studies in Section~4.1 and 4.2 of the main paper.

\subsection{Additional Simulation Results in Regression Settings}
\label{sec::app-simu-reg}

This section complements the results reported in Section~4.1 of the main paper. Table~\ref{tab:tab5} reports the results from the paired $t$-test on model performance metrics between TSLA and other methods in regression settings.

\begin{table}[htp]
\caption{Simulation: paired $t$-test on model performance metrics between TSLA and other methods in regression settings. Reported are the means, standard errors (in parentheses) of the paired differences, and the $p$-values (in parentheses) of the $t$-test over 100 repeated experiments.}
\label{tab:tab5}
\centering
\footnotesize
\begin{tabular}{l|ccc|cc}
\toprule
       & \multicolumn{3}{c|}{MSE}     & FNR & FPR\\
       & RFS-Sum   & Enet   & Enet2       & \multicolumn{2}{c}{RFS-Sum} \\
\midrule
Case 1 &  
\begin{tabular}[c]{@{}c@{}}-0.276 (0.037)\\ (\textless 0.0001)\end{tabular} & \begin{tabular}[c]{@{}c@{}}-0.400 (0.040)\\ (\textless 0.0001)\end{tabular} & \begin{tabular}[c]{@{}c@{}}-1.240 (0.068)\\ (\textless 0.0001)\end{tabular} &
\begin{tabular}[c]{@{}c@{}}-0.095 (0.026) \\ (0.0005)\end{tabular} &
\begin{tabular}[c]{@{}c@{}}-0.063 (0.010) \\ (\textless 0.0001)\end{tabular}  \\
Case 2 &  
\begin{tabular}[c]{@{}c@{}}-0.290 (0.019)\\ (\textless 0.0001)\end{tabular} & \begin{tabular}[c]{@{}c@{}}-0.334 (0.020)\\ (\textless 0.0001)\end{tabular} & \begin{tabular}[c]{@{}c@{}}-0.551 (0.029)\\ (\textless 0.0001)\end{tabular}  &
\begin{tabular}[c]{@{}c@{}}-0.075 (0.030) \\ (0.0129)\end{tabular}  &
\begin{tabular}[c]{@{}c@{}}-0.056 (0.008) \\ (\textless 0.0001)\end{tabular} \\
Case 3 &  
\begin{tabular}[c]{@{}c@{}}-0.212 (0.021)\\ (\textless 0.0001)\end{tabular} & \begin{tabular}[c]{@{}c@{}}-0.227 (0.026)\\ (\textless 0.0001)\end{tabular} & \begin{tabular}[c]{@{}c@{}}-0.743 (0.039)\\ (\textless 0.0001)\end{tabular}  &
\begin{tabular}[c]{@{}c@{}} 0.000 (0.000) \\ (NA)\end{tabular}  &
\begin{tabular}[c]{@{}c@{}}-0.060 (0.007) \\ (\textless 0.0001)\end{tabular}  \\
\bottomrule
\end{tabular}
\end{table}

\clearpage
\subsection{Additional Simulation Results in Classification Settings}
\label{sec::app-simu-log}

\begin{figure}[H]
  \centering
  \includegraphics[width=\linewidth]{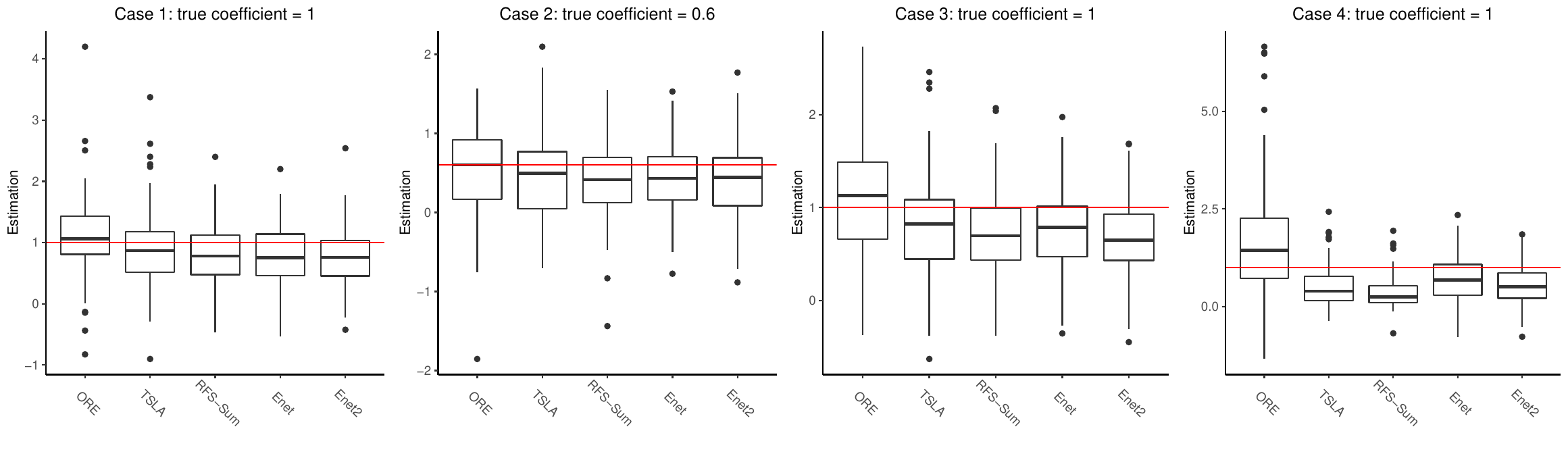}
  \caption{Simulation: boxplots for estimations of the unpenalized coefficient in classification settings over 100 repeated experiments. The horizontal line in each plot marks the true coefficient. }
  \label{fig:fig9}
\end{figure}

This section complements the results reported in Section~4.2 of the main paper. Table~\ref{tab:tab6} reports the results on feature aggregation. Table~\ref{tab:tab7} records the prediction results for Case~2 and Case~3 in the classification settings. 
Table~\ref{tab:tab8} reports the paired $t$-test results for each case, and 
Figure~\ref{fig:fig9}
reports the estimation of the regression coefficients of $z$, the unpenalized covariate. 

\begin{table}[H]
\centering
\footnotesize
\caption{Simulation: feature aggregation performance in classification settings.}
\label{tab:tab6}
\begin{tabular}{l|cc|cc}
\toprule
      & \multicolumn{2}{c|}{FNR/FPR} & \multicolumn{2}{c}{Paired $t$-test}           \\
\midrule
    & TSLA                 & RFS-Sum                     & mean difference & $p$-value                   \\
Case 1 & \begin{tabular}[c]{@{}c@{}}0.225 / 0.091\\ (0.025) / (0.010)\end{tabular} 
            & \begin{tabular}[c]{@{}c@{}}0.235 / 0.166 \\ (0.025) / (0.010)\end{tabular}
            & \begin{tabular}[c]{@{}c@{}}-0.010 / -0.074\\ (0.028) / (0.013)\end{tabular} 
            & 0.7170 / \textless 0.0001 \\
Case 2 & \begin{tabular}[c]{@{}c@{}}0.275 / 0.126\\ (0.025) / (0.011)\end{tabular} 
            & \begin{tabular}[c]{@{}c@{}}0.245 / 0.167 \\ (0.025) / (0.012)\end{tabular}
            & \begin{tabular}[c]{@{}c@{}}0.030 / -0.041\\ (0.032) / (0.014)\end{tabular} 
            & 0.3453 / 0.0028 \\
Case 3 & \begin{tabular}[c]{@{}c@{}}0.190 / 0.096\\ (0.024) / (0.010)\end{tabular} 
            & \begin{tabular}[c]{@{}c@{}}0.260 / 0.136 \\ (0.025) / (0.009)\end{tabular}
            & \begin{tabular}[c]{@{}c@{}}-0.070 / -0.040\\ (0.031) / (0.010)\end{tabular} 
            & 0.0261 / 0.0002 \\
Case 4 & \begin{tabular}[c]{@{}c@{}}0.065 / 0.122\\ (0.017) / (0.006)\end{tabular} 
            & \begin{tabular}[c]{@{}c@{}}0.000 / 0.191 \\ (0.000) / (0.007)\end{tabular}
            & \begin{tabular}[c]{@{}c@{}}0.065 / -0.068\\ (0.017) / (0.008)\end{tabular} 
            & 0.0002 / \textless 0.0001 \\
\bottomrule
\end{tabular}
\end{table}

\begin{table}[H]
\footnotesize
\centering
\caption{Simulation: prediction performance in classification settings under Case 2 and Case 3. Reported are the means and standard errors (in parentheses). }
\label{tab:tab7}
\begin{tabular}{l|ccccccc}
\toprule
               &            &            & \multicolumn{2}{c}{Sensitivity} & \multicolumn{2}{c}{PPV}         \\
Model          & AUC        & AUPRC      & 90\% specificity & 95\% specificity & 90\% specificity & 95\% specificity \\
\midrule
               & \multicolumn{6}{c}{Case 2}\\
\hline  
ORE &0.851 (0.002) & 0.557 (0.005) & 0.504 (0.008) & 0.337 (0.006) & 0.510 (0.004) & 0.582 (0.005)\\ 
TSLA &0.823 (0.004) & 0.498 (0.005) & 0.453 (0.007) & 0.296 (0.006) & 0.484 (0.004) & 0.550 (0.005)\\
RFS-Sum & 0.815 (0.003) & 0.450 (0.005) & 0.417 (0.007) & 0.255 (0.006) & 0.463 (0.004) & 0.510 (0.007)\\ 
Enet & 0.795 (0.003) & 0.443 (0.004) & 0.405 (0.006) & 0.251 (0.005) & 0.456 (0.004) & 0.509 (0.005)\\
Enet2 & 0.751 (0.005) & 0.399 (0.006) & 0.366 (0.007) & 0.215 (0.005) & 0.430 (0.005) & 0.468 (0.006)\\
 \hline
               & \multicolumn{6}{c}{Case 3}\\
  \hline
ORE & 0.936 (0.001) & 0.853 (0.003) & 0.816 (0.004) & 0.610 (0.007) & 0.778 (0.002) & 0.839 (0.002)\\
TSLA & 0.922 (0.001) & 0.813 (0.004) & 0.770 (0.006) & 0.536 (0.009) & 0.767 (0.002) & 0.819 (0.003)\\
RFS-Sum & 0.916 (0.002) & 0.779 (0.004) & 0.752 (0.006) & 0.491 (0.008) & 0.763 (0.002) & 0.806 (0.003)\\ 
Enet & 0.909 (0.002) & 0.773 (0.003) & 0.720 (0.007) & 0.464 (0.008) & 0.754 (0.002) & 0.797 (0.003)\\ 
Enet2 & 0.900 (0.002) & 0.752 (0.003) & 0.681 (0.007) & 0.418 (0.007) & 0.744 (0.002) & 0.780 (0.003)\\   
\bottomrule
\end{tabular}
\end{table}

\begin{table}[H]
\footnotesize
\centering
\caption{Simulation: paired $t$-test on model performance metrics between TSLA and other methods for Case 1--Case 4 in classification settings.}
\label{tab:tab8}
\begin{tabular}{l|cccccc}
\toprule
        &         &    & \multicolumn{2}{c}{Sensitivity} & \multicolumn{2}{c}{PPV}   \\
        & AUC    & AUPRC  & 90\% specificity & 95\% specificity  & 90\% specificity & 95\% specificity   \\
\midrule      
               & \multicolumn{6}{c}{Case 1}\\
\hline  
RFS-Sum &
\begin{tabular}[c]{@{}c@{}}0.017 (0.002)\\ (\textless 0.0001)\end{tabular} & \begin{tabular}[c]{@{}c@{}}0.089 (0.005)\\ (\textless 0.0001)\end{tabular} & \begin{tabular}[c]{@{}c@{}}0.050 (0.007)\\ (\textless 0.0001)\end{tabular} & \begin{tabular}[c]{@{}c@{}}0.077 (0.007)\\ (\textless 0.0001)\end{tabular} & \begin{tabular}[c]{@{}c@{}}0.020 (0.003)\\ (\textless 0.0001)\end{tabular} & \begin{tabular}[c]{@{}c@{}}0.048 (0.005)\\ (\textless 0.0001)\end{tabular} \\
Enet    & 
\begin{tabular}[c]{@{}c@{}}0.034 (0.002)\\ (\textless 0.0001)\end{tabular} & \begin{tabular}[c]{@{}c@{}}0.093 (0.005)\\ (\textless 0.0001)\end{tabular} & \begin{tabular}[c]{@{}c@{}}0.086 (0.007)\\ (\textless 0.0001)\end{tabular} & \begin{tabular}[c]{@{}c@{}}0.079 (0.006)\\ (\textless 0.0001)\end{tabular} & \begin{tabular}[c]{@{}c@{}}0.036 (0.003)\\ (\textless 0.0001)\end{tabular} & \begin{tabular}[c]{@{}c@{}}0.045 (0.004)\\ (\textless 0.0001)\end{tabular} \\
Enet2   & 
\begin{tabular}[c]{@{}c@{}}0.062 (0.004)\\ (\textless 0.0001)\end{tabular} & \begin{tabular}[c]{@{}c@{}}0.132 (0.006)\\ (\textless 0.0001)\end{tabular} & \begin{tabular}[c]{@{}c@{}}0.133 (0.008)\\ (\textless 0.0001)\end{tabular} & \begin{tabular}[c]{@{}c@{}}0.126 (0.007)\\ (\textless 0.0001)\end{tabular} & \begin{tabular}[c]{@{}c@{}}0.058 (0.003)\\ (\textless 0.0001)\end{tabular} & \begin{tabular}[c]{@{}c@{}}0.081 (0.005)\\ (\textless 0.0001)\end{tabular} \\
\hline
               & \multicolumn{6}{c}{Case 2}\\
\hline  
RFS-Sum &
\begin{tabular}[c]{@{}c@{}}0.008 (0.003)\\ (0.0021)\end{tabular} & \begin{tabular}[c]{@{}c@{}}0.047 (0.004)\\ (\textless 0.0001)\end{tabular} & \begin{tabular}[c]{@{}c@{}}0.036 (0.006)\\ (\textless 0.0001)\end{tabular} & \begin{tabular}[c]{@{}c@{}}0.042 (0.006)\\ (\textless 0.0001)\end{tabular} & \begin{tabular}[c]{@{}c@{}}0.021 (0.004)\\ (\textless 0.0001)\end{tabular} & \begin{tabular}[c]{@{}c@{}}0.040 (0.006)\\ (\textless 0.0001)\end{tabular} \\
Enet    & 
\begin{tabular}[c]{@{}c@{}}0.029 (0.004)\\ (\textless 0.0001)\end{tabular} & \begin{tabular}[c]{@{}c@{}}0.055 (0.005)\\ (\textless 0.0001)\end{tabular} & \begin{tabular}[c]{@{}c@{}}0.048 (0.006)\\ (\textless 0.0001)\end{tabular} & \begin{tabular}[c]{@{}c@{}}0.045 (0.005)\\ (\textless 0.0001)\end{tabular} & \begin{tabular}[c]{@{}c@{}}0.027 (0.004)\\ (\textless 0.0001)\end{tabular} & \begin{tabular}[c]{@{}c@{}}0.041 (0.005)\\ (\textless 0.0001)\end{tabular} \\
Enet2   & 
\begin{tabular}[c]{@{}c@{}}0.073 (0.005)\\ (\textless 0.0001)\end{tabular} & \begin{tabular}[c]{@{}c@{}}0.099 (0.005)\\ (\textless 0.0001)\end{tabular} & \begin{tabular}[c]{@{}c@{}}0.087 (0.007)\\ (\textless 0.0001)\end{tabular} & \begin{tabular}[c]{@{}c@{}}0.082 (0.006)\\ (\textless 0.0001)\end{tabular} & \begin{tabular}[c]{@{}c@{}}0.054 (0.005)\\ (\textless 0.0001)\end{tabular} & \begin{tabular}[c]{@{}c@{}}0.082 (0.006)\\ (\textless 0.0001)\end{tabular} \\
\hline
              & \multicolumn{6}{c}{Case 3}\\
\hline  
RFS-Sum &
\begin{tabular}[c]{@{}c@{}}0.006 (0.001)\\ (\textless 0.0001)\end{tabular} & \begin{tabular}[c]{@{}c@{}}0.035 (0.003)\\ (\textless 0.0001)\end{tabular} & \begin{tabular}[c]{@{}c@{}}0.017 (0.005)\\ (0.0003)\end{tabular} & \begin{tabular}[c]{@{}c@{}}0.044 (0.006)\\ (\textless 0.0001)\end{tabular} & \begin{tabular}[c]{@{}c@{}}0.004 (0.001)\\ (0.0007)\end{tabular} & \begin{tabular}[c]{@{}c@{}}0.013 (0.002)\\ (\textless 0.0001)\end{tabular} \\
Enet    & 
\begin{tabular}[c]{@{}c@{}}0.014 (0.001)\\ (\textless 0.0001)\end{tabular} & \begin{tabular}[c]{@{}c@{}}0.041 (0.003)\\ (\textless 0.0001)\end{tabular} & \begin{tabular}[c]{@{}c@{}}0.050 (0.006)\\ (\textless 0.0001)\end{tabular} & \begin{tabular}[c]{@{}c@{}}0.072 (0.007)\\ (\textless 0.0001)\end{tabular} & \begin{tabular}[c]{@{}c@{}}0.013 (0.002)\\ (\textless 0.0001)\end{tabular} & \begin{tabular}[c]{@{}c@{}}0.022 (0.002)\\ (\textless 0.0001)\end{tabular} \\
Enet2   & 
\begin{tabular}[c]{@{}c@{}}0.023 (0.002)\\ (\textless 0.0001)\end{tabular} & \begin{tabular}[c]{@{}c@{}}0.062 (0.003)\\ (\textless 0.0001)\end{tabular} & \begin{tabular}[c]{@{}c@{}}0.089 (0.006)\\ (\textless 0.0001)\end{tabular} & \begin{tabular}[c]{@{}c@{}}0.118 (0.008)\\ (\textless 0.0001)\end{tabular} & \begin{tabular}[c]{@{}c@{}}0.023 (0.002)\\ (\textless 0.0001)\end{tabular} & \begin{tabular}[c]{@{}c@{}}0.039 (0.003)\\ (\textless 0.0001)\end{tabular} \\
\hline
             & \multicolumn{6}{c}{Case 4}\\
\hline  
RFS-Sum &
\begin{tabular}[c]{@{}c@{}}0.012 (0.002)\\ (\textless 0.0001)\end{tabular} & \begin{tabular}[c]{@{}c@{}}0.054 (0.004)\\ (\textless 0.0001)\end{tabular} & \begin{tabular}[c]{@{}c@{}}0.037 (0.006)\\ (\textless 0.0001)\end{tabular} & \begin{tabular}[c]{@{}c@{}}0.070 (0.006)\\ (\textless 0.0001)\end{tabular} & \begin{tabular}[c]{@{}c@{}}0.014 (0.002)\\ (\textless 0.0001)\end{tabular} & \begin{tabular}[c]{@{}c@{}}0.040 (0.004)\\ (\textless 0.0001)\end{tabular}  \\
Enet    & 
\begin{tabular}[c]{@{}c@{}}0.098 (0.007)\\ (\textless 0.0001)\end{tabular} & \begin{tabular}[c]{@{}c@{}}0.156 (0.008)\\ (\textless 0.0001)\end{tabular} & \begin{tabular}[c]{@{}c@{}}0.194 (0.011)\\ (\textless 0.0001)\end{tabular} & \begin{tabular}[c]{@{}c@{}}0.151 (0.011)\\ (\textless 0.0001)\end{tabular} & \begin{tabular}[c]{@{}c@{}}0.096 (0.007)\\ (\textless 0.0001)\end{tabular} & \begin{tabular}[c]{@{}c@{}}0.112 (0.008)\\ (\textless 0.0001)\end{tabular} \\
Enet2   & 
\begin{tabular}[c]{@{}c@{}}0.195 (0.007)\\ (\textless 0.0001)\end{tabular} & \begin{tabular}[c]{@{}c@{}}0.265 (0.009)\\ (\textless 0.0001)\end{tabular} & \begin{tabular}[c]{@{}c@{}}0.321 (0.013)\\ (\textless 0.0001)\end{tabular} & \begin{tabular}[c]{@{}c@{}}0.242 (0.013)\\ (\textless 0.0001)\end{tabular} & \begin{tabular}[c]{@{}c@{}}0.180 (0.009)\\ (\textless 0.0001)\end{tabular} & \begin{tabular}[c]{@{}c@{}}0.210 (0.011)\\ (\textless 0.0001)\end{tabular} \\
\bottomrule
\end{tabular}
\end{table}

\subsection{Simulation Results: Effects of Tree Structure and Feature Rarity
}\label{sec::app-simu-k-rp}

This section complements the results reported in Section~4.3 of the main paper.

\subsubsection{Effects of Tree Structure}\label{sec::app-simu-k}

To investigate the effects of the maximum number of child nodes, $k$, we construct three tree structures with $k \in \{2, 3, 4\}$. At the same time, the number of leaf nodes $p_0$ keeps the same. The original tree structures are shown in Figures~\ref{fig:fig10}--\ref{fig:fig12}, with decreasing node group numbers 12, 10, and 7.  All the tree structures have 19 leaf nodes. 

We assume the underlying model for different $k$ values to be the same. Specifically, we take 
\begin{align*}
	\eta\ =\ -2.5\ +\ 3(x_1\ \lor\ x_2\ \lor\ x_3\ \lor\ x_4)\ -\ 1.5(x_5\ \lor\ x_6\ \lor\ x_7 )\ +\ 2x_8\ -\ 2x_{9},
\end{align*}
where the original design matrix $\X_0\in \{0, 1\}^{n\times p_0}$ is generated with independent $\mbox{Bernoulli}(0.1)$ entries. {\cre Based on the generated design matrix $\X_0$, under the regression settings, we generate $n$ independent samples of $y$ from $y=\eta+\varepsilon$, in which each $\varepsilon$ is independently generated from $N(0, \sigma^2)$ with $\mbox{SNR} =0.5$. 
Under the classification settings, we obtain $n$ independent samples of $y$ from $\mbox{Bernoulli}(q)$, where $q=\exp(z+\eta)/\{1+\exp(z+\eta)\}$, and  
each $z$ is independently sampled from $N(0, 0.25)$. The average value of $q$ is around 0.22. }

Both TSLA and RFS-Sum rely on a specific tree structure, and we fit each model with all three $k$ values. Other competing methods are not affected by $k$. We use 200 observations for training, and the results are measured on 200/1000 independent samples for regression/classification settings. 

Figure~\ref{fig:fig13}(a) reports the MSE across 100 repetitions under the regression settings, and the paired $t$-test results are recorded in Table~\ref{tab:tab9}. Table~\ref{tab:tab10} records the results under the classification settings, and Figure~\ref{fig:fig14}(a) reports the estimation of the regression coefficients of $z$. The results show that the performance of TSLA is better when $k=2$ as compared to $k=3,\ 4$, and the performance of RFS-Sum is not affected much by $k$. For all values of $k$, TSLA is the best model. 

\begin{figure}[H]
	\centering
        \begin{tikzpicture}
        \tiny
            \tikzstyle{level 1}=[sibling distance=80mm]
            \tikzstyle{level 2}=[sibling distance=40mm]
            \tikzstyle{level 3}=[sibling distance=20mm]
            \tikzstyle{level 4}=[sibling distance=15mm]
            \tikzstyle{level 5}=[sibling distance=10mm]
            \node {$u_1^0$} [align=center]
            	child {node {$u_1^1$}
			child {node {$u_1^2$}
				child {node {$u_1^3$}
					child {node {$u_1^4$}
						child {node {$u_1^5$ \\ $x_1$}}
						child {node {$u_2^5$ \\ $x_2$}}}
					child {node {$u_2^4$}
						child {node {$u_3^5$ \\ $x_3$}}
						child {node {$u_4^5$ \\ $x_4$}}}}
				child {node {$u_2^3$}
					child {node {$u_3^4$ \\ $x_5$}}
					child {node {$u_4^4$}
						child {node {$u_5^5$ \\ $x_6$}}
						child {node {$u_6^5$ \\ $x_7$}}}}}
			child {node {$u_2^2$}
				child {node {$u_3^3$}
					child {node {$u_5^4$ \\ $x_8$}}
					child {node {$u_6^4$ \\ $x_9$}}}
				child {node {$u_4^3$}
					child {node {$u_7^4$ \\ $x_{10}$}}
					child {node {$u_8^4$ \\ $x_{11}$}}}}}
		child {node {$u_2^1$}
			child {node {$u_3^2$}
				child {node {$u_5^3$}
					child {node {$u_9^4$ \\ $x_{12}$}}
					child {node {$u_{10}^4$ \\ $x_{13}$}}}
				child {node {$u_6^3$}
					child {node {$u_{11}^4$ \\ $x_{14}$}}
					child {node {$u_{12}^4$ \\ $x_{15}$}}}}
			child {node {$u_4^2$}
				child {node {$u_7^3$}
					child {node {$u_{13}^4$ \\ $x_{16}$}}
					child {node {$u_{14}^4$ \\ $x_{17}$}}}
				child {node {$u_8^3$}
					child {node {$u_{15}^4$ \\ $x_{18}$}}
					child {node {$u_{16}^4$ \\ $x_{19}$}}}}};
	\end{tikzpicture}
	\caption{Tree structure for $k$ = 2. }
	\label{fig:fig10}
	
\end{figure}
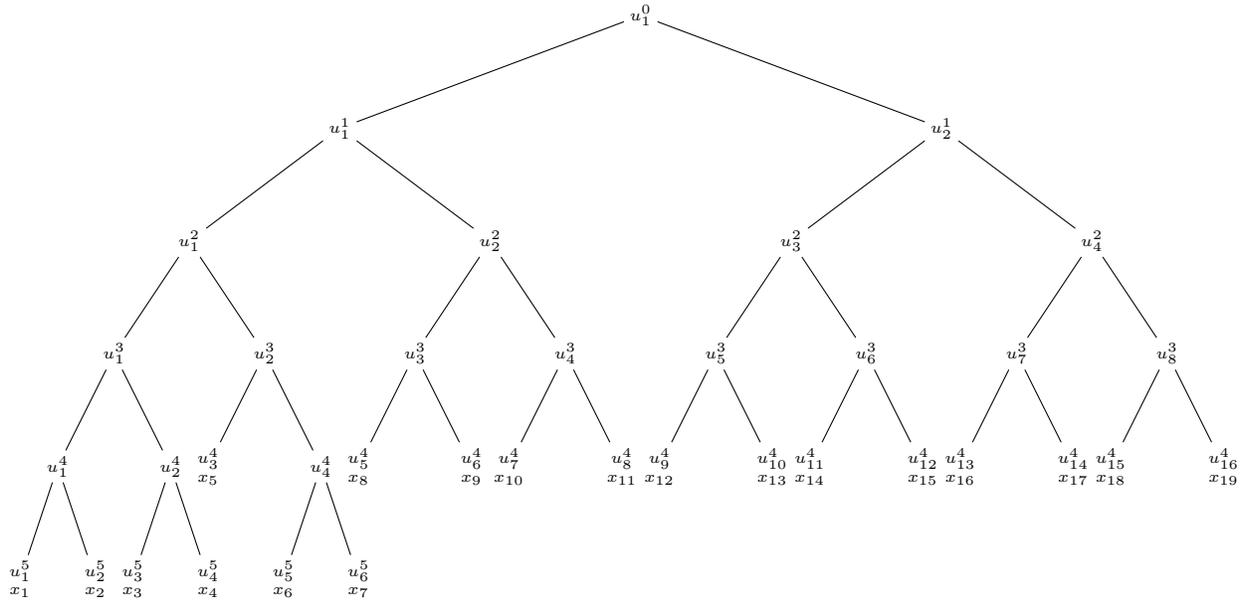

\begin{figure}[H]
        \centering
        \begin{tikzpicture}
        \tiny
            \tikzstyle{level 1}=[sibling distance=65mm]
            \tikzstyle{level 2}=[sibling distance=20mm]
            \tikzstyle{level 3}=[sibling distance=8mm]
            \tikzstyle{level 4}=[sibling distance=5mm]
            \node {$u_1^0$} [align=center]
                child {node {$u_1^1$}
                		child {node {$u_1^2$}
				child {node {$u_1^3$ \\ $x_1$}}
				child {node {$u_2^3$ \\ $x_2$}}
				child {node {$u_3^3$ \\ $x_3$}}}
			child {node {$u_2^2$ \\ $x_4$}}
			child {node {$u_3^2$}
				child {node {$u_4^3$ \\ $x_5$}}
				child {node {$u_5^3$ \\ $x_6$}}
				child {node {$u_6^3$ \\ $x_7$}}}}
                child {node {$u_2^1$}
                		child {node {$u_4^2$}
				child {node {$u_7^3$ \\ $x_8$}}
				child {node {$u_8^3$ \\ $x_9$}}
				child {node {$u_9^3$ \\ $x_{10}$}}}
			child {node {$u_5^2$}
				child {node {$u_{10}^3$ \\ $x_{11}$}}
				child {node {$u_{11}^3$ \\ $x_{12}$}}
				child {node {$u_{12}^3$ \\ $x_{13}$}}}
			child {node {$u_6^2$}
				child {node {$u_{13}^3$ \\ $x_{14}$}}
				child {node {$u_{14}^3$ \\ $x_{15}$}}
				child {node {$u_{15}^3$ \\ $x_{16}$}}}}
                child {node {$u_3^1$}
                		child {node {$u_7^2$ \\ $x_{17}$}}
			child {node {$u_8^2$ \\ $x_{18}$}}
			child {node {$u_9^2$ \\ $x_{19}$}}};
    \end{tikzpicture}
    \caption{Tree structure for $k$ = 3. }
    \label{fig:fig11}
\end{figure}
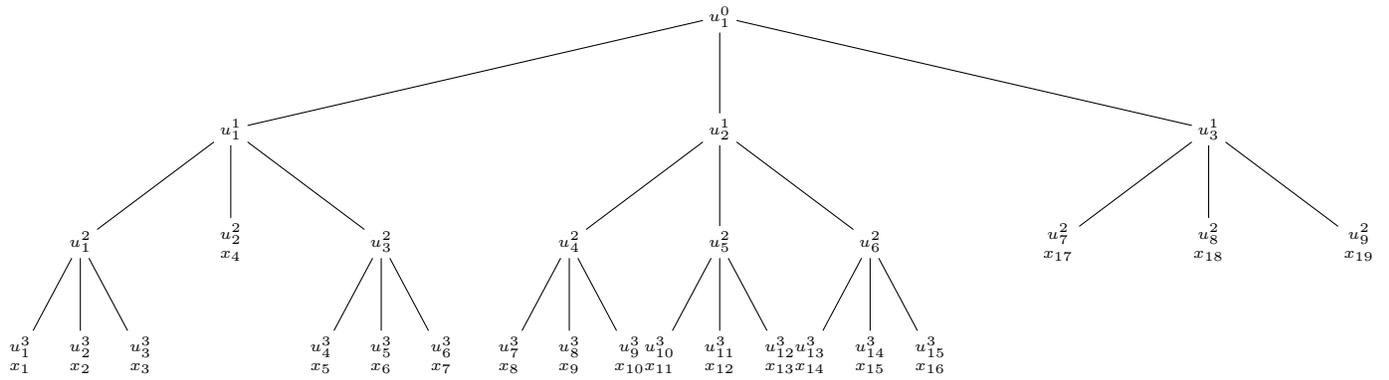

\begin{figure}[H]
        \centering
        \begin{tikzpicture}
        \tiny
            \tikzstyle{level 1}=[sibling distance=45mm]
            \tikzstyle{level 2}=[sibling distance=10mm]
            \tikzstyle{level 3}=[sibling distance=5mm]
            \tikzstyle{level 4}=[sibling distance=5mm]
            \node {$u_1^0$} [align=center]
                child {node {$u_1^1$}
                		child {node {$u_1^2$}
				child {node {$u_1^3$ \\ $x_1$}}
				child {node {$u_2^3$ \\ $x_2$}}
				child {node {$u_3^3$ \\ $x_3$}}
				child {node {$u_4^2$ \\ $x_4$}}}
			child {node {$u_2^2$ \\ $x_5$}}
			child {node {$u_3^2$ \\ $x_6$}}
			child {node {$u_4^2$ \\ $x_7$}}}
                child {node {$u_2^1$}
                		child {node {$u_5^2$ \\ $x_8$}}
			child {node {$u_6^2$ \\ $x_9$}}
			child {node {$u_7^2$ \\ $x_{10}$}}
			child {node {$u_8^2$ \\ $x_{11}$}}}
                child {node {$u_3^1$}
                		child {node {$u_9^2$ \\ $x_{12}$}}
			child {node {$u_{10}^2$ \\ $x_{13}$}}
			child {node {$u_{11}^2$ \\ $x_{14}$}}
			child {node {$u_{12}^2$ \\ $x_{15}$}}}
		child {node {$u_4^1$}
                		child {node {$u_{13}^2$ \\ $x_{16}$}}
			child {node {$u_{14}^2$ \\ $x_{17}$}}
			child {node {$u_{15}^2$ \\ $x_{18}$}}
			child {node {$u_{16}^2$ \\ $x_{19}$}}};

    \end{tikzpicture}
    \caption{Tree structure for $k$ = 4.}
    \label{fig:fig12}
\end{figure}
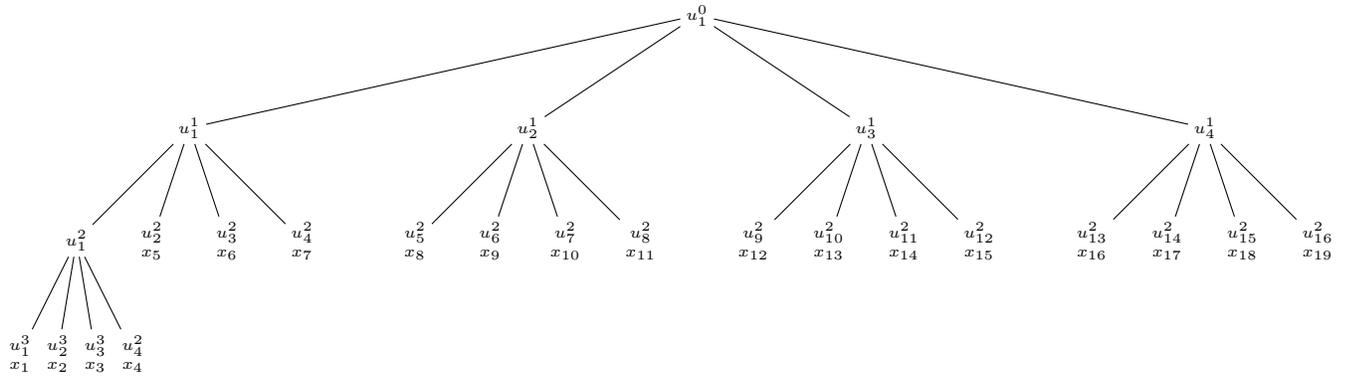

\begin{figure}[htp]
     \centering
     \begin{subfigure}[b]{0.49\textwidth}
         \centering
         \includegraphics[width=\textwidth, height=0.65\textwidth]{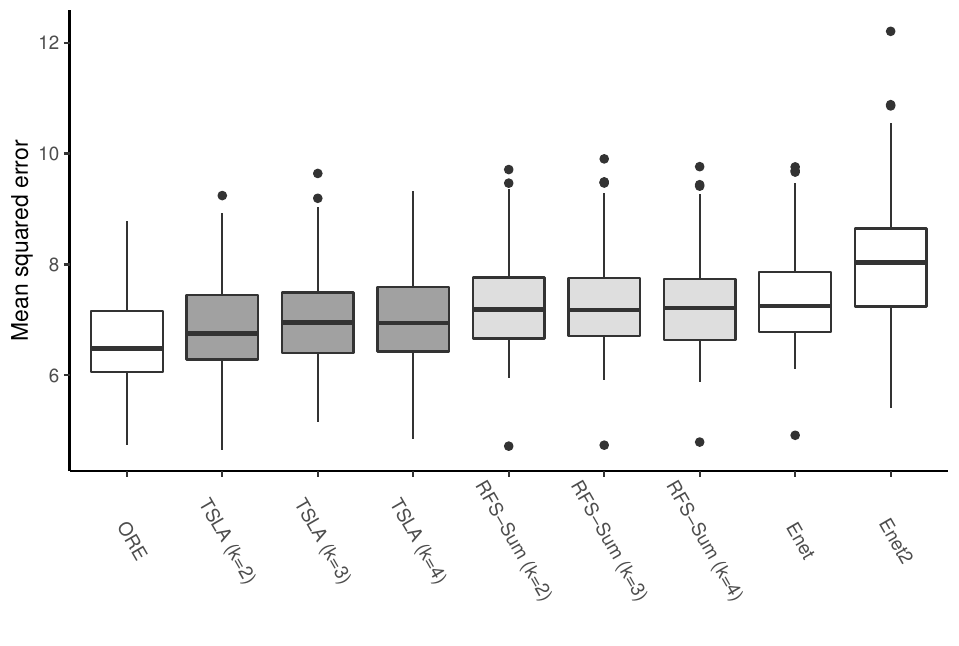}
         \caption{Effects of $k$}
         \label{fig:simu-k}
     \end{subfigure}
     \hfill
     \begin{subfigure}[b]{0.49\textwidth}
         \centering
         \includegraphics[width=\textwidth, height=0.65\textwidth]{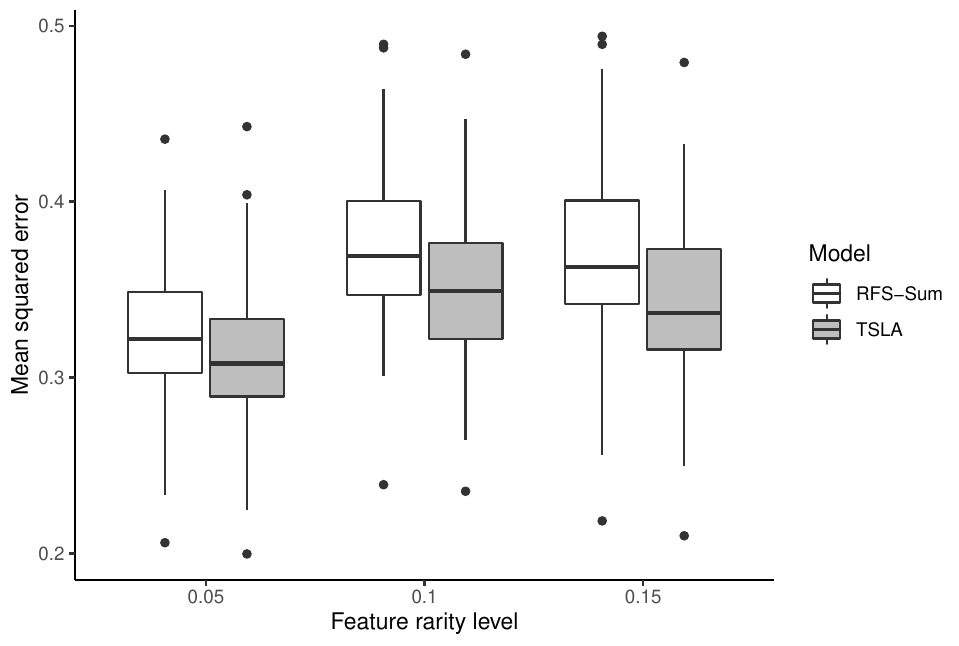}
         \caption{Effects of feature rarity}
         \label{fig:simu-rp}
       \end{subfigure}
       \caption{Simulation: (a) boxplot of out-of-sample mean squared error with different $k$ values. (b) boxplot of out-of-sample mean squared error corresponding to different feature rarity levels. The values are scaled by the Frobenius norm of the original design matrix $\X_0$.}
  \label{fig:fig13}
\end{figure}

\begin{table}[H]
\footnotesize
\centering
\caption{Simulation: paired $t$-test on MSE between TSLA and other methods for different values of $k$. The comparison between TSLA and RFS-Sum is made with matched value of $k$.
}
\label{tab:tab9}
\begin{tabular}{l|ccc}
\toprule
        &   RFS-Sum &  Enet & Enet2 \\
\midrule   
$k = 2$ & 
\begin{tabular}[c]{@{}c@{}}-0.451 (0.030)\\ (\textless 0.0001)\end{tabular} & \begin{tabular}[c]{@{}c@{}}-0.561 (0.037)\\ (\textless 0.0001)\end{tabular} & \begin{tabular}[c]{@{}c@{}}-1.259 (0.068)\\ (\textless 0.0001)\end{tabular} \\
$k = 3$ & 
\begin{tabular}[c]{@{}c@{}}-0.275 (0.027)\\ (\textless 0.0001)\end{tabular} & \begin{tabular}[c]{@{}c@{}}-0.377 (0.033)\\ (\textless 0.0001)\end{tabular} & \begin{tabular}[c]{@{}c@{}}-1.075 (0.068)\\ (\textless 0.0001)\end{tabular}\\
$k = 4$ &  
\begin{tabular}[c]{@{}c@{}}-0.254 (0.029)\\ (\textless 0.0001)\end{tabular} & \begin{tabular}[c]{@{}c@{}}-0.382 (0.034)\\ (\textless 0.0001)\end{tabular} & \begin{tabular}[c]{@{}c@{}}-1.080 (0.068)\\ (\textless 0.0001)\end{tabular} \\
\bottomrule
\end{tabular}
\end{table}

\begin{table}[H]
\centering
\caption{Simulation: prediction performance for different values of $k$ in classification settings. All the paired $t$-tests between TSLA and the competing methods are significant at the significance level $\alpha=0.01$.
}
\label{tab:tab10}
\footnotesize
\begin{tabular}{l|ccccccc}
\toprule
               &            &            & \multicolumn{2}{c}{Sensitivity} & \multicolumn{2}{c}{PPV}         \\
Model          & AUC        & AUPRC      & 90\% specificity & 95\% specificity & 90\% specificity & 95\% specificity \\
\midrule
ORE & 0.867 (0.001) & 0.690 (0.003) & 0.633 (0.004) & 0.451 (0.005) & 0.662 (0.002) & 0.737 (0.003) \\
TSLA ($k=2$)& 0.849 (0.002) & 0.642 (0.004) & 0.582 (0.006) & 0.393 (0.007) & 0.642 (0.003) & 0.707 (0.004) \\
TSLA ($k=3$)& 0.838 (0.002) & 0.616 (0.005) & 0.557 (0.007) & 0.361 (0.006) & 0.632 (0.004) & 0.688 (0.004)  \\
TSLA ($k=4$)& 0.839 (0.002) & 0.621 (0.005) & 0.553 (0.007) & 0.368 (0.007) & 0.630 (0.003) & 0.692 (0.004) \\
RFS-Sum ($k=2$)& 0.833 (0.002) & 0.585 (0.004) & 0.516 (0.007) & 0.315 (0.006) & 0.614 (0.004) & 0.657 (0.005)  \\
RFS-Sum ($k=3$)& 0.829 (0.003) & 0.585 (0.005) & 0.518 (0.006) & 0.318 (0.006) & 0.615 (0.004) & 0.660 (0.005)  \\
RFS-Sum ($k=4$)& 0.829 (0.002) & 0.578 (0.004) & 0.505 (0.006) & 0.302 (0.006) & 0.609 (0.004) & 0.648 (0.005) \\
Enet & 0.818 (0.003) & 0.573 (0.005) & 0.489 (0.007) & 0.305 (0.006) & 0.600 (0.004) & 0.651 (0.005) \\
Enet2 & 0.774 (0.005) & 0.510 (0.006) & 0.410 (0.007) & 0.245 (0.006) & 0.556 (0.005) & 0.598 (0.006)  \\
\bottomrule
\end{tabular}
\end{table}

\begin{figure}[htp]
     \centering
     \begin{subfigure}[b]{0.49\textwidth}
         \centering
         \includegraphics[width=\textwidth, height=0.65\textwidth]{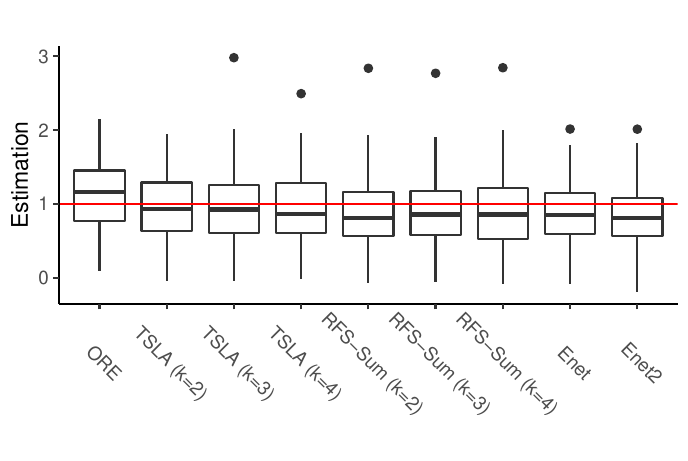}
         \caption{Effects of $k$}
     \end{subfigure}
     \hfill
     \begin{subfigure}[b]{0.49\textwidth}
         \centering
         \includegraphics[width=\textwidth, height=0.65\textwidth]{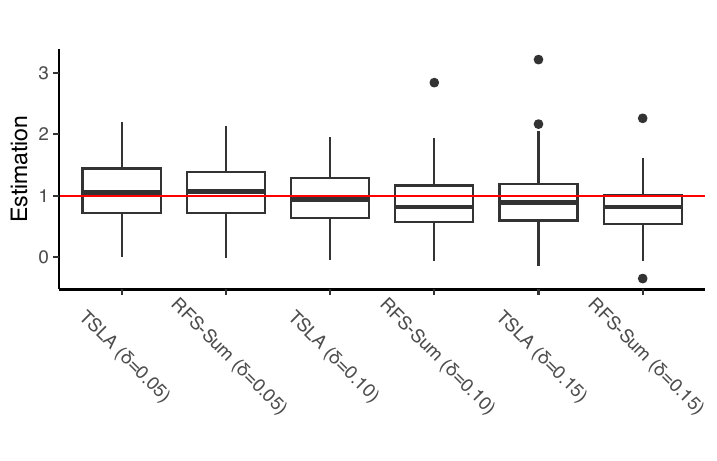}
         \caption{Effects of feature rarity}
      \end{subfigure}
      \hfill
     \caption{Simulation: boxplots for estimations of the unpenalized coefficient in classification settings: (a) with different $k$ values; (b) with different feature rarity levels. The horizontal line in each plot marks the true coefficient.}
  \label{fig:fig14}
\end{figure}

\subsubsection{Effects of Feature Rarity}
\label{sec::app-simu-rp}

In this section, we generate the original design matrix $\X_0$ as a sparse binary matrix with independent Bernoulli$(\delta)$ entries with three rarity levels $\delta \in \{0.05, 0.10, 0.15\}$. We adopt the tree structure in Figure~\ref{fig:fig10}, and the rest of the data generation follows the same process described in Section~\ref{sec::app-simu-k} of the Supplement. 

Figure~\ref{fig:fig13}(b) shows the out-of-sample MSE over 100 experiments under the regression settings, and the paired $t$-test results between TSLA and RFS-Sum are reported in Table~\ref{tab:tab11}. The results are scaled by the Frobenius norm of the original design matrix $\X_0$. The detailed prediction results are summarized in Tables~\ref{tab:tab12}--\ref{tab:tab13} for classification. TSLA always performs significantly better than RFS-Sum even with $\delta = 0.05$. Figure~\ref{fig:fig14}(b) reports the estimation of the regression coefficients of $z$ under the classification settings. 

\begin{table}[H]
\centering
\footnotesize
\caption{Simulation: paired $t$-test on MSE between TSLA and RFS-Sum with different feature rarity levels in regression settings. 
}
\label{tab:tab11}
\begin{tabular}{l|cc}
\toprule
         &  \multicolumn{2}{c}{Paired $t$-test} \\
         &  mean difference     & $p$-value     \\
\midrule
$\delta$ = 0.05 &  -0.011 (0.001)& \textless 0.0001 \\
$\delta$ = 0.10 &  -0.023 (0.002)& \textless 0.0001 \\
$\delta$ = 0.15 &  -0.028 (0.002)& \textless 0.0001 \\
\bottomrule
\end{tabular}
\end{table}

\begin{table}[H]
\centering
\caption{Simulation: prediction performance for different feature rarity levels in classification settings.
}
\label{tab:tab12}
\footnotesize
\begin{tabular}{l|ccccccc}
\toprule
               &            &            & \multicolumn{2}{c}{Sensitivity} & \multicolumn{2}{c}{PPV}         \\
Model          & AUC        & AUPRC      & 90\% specificity & 95\% specificity & 90\% specificity & 95\% specificity \\
\midrule
TSLA ($\delta$ = 0.05) & 0.809 (0.004) & 0.539 (0.006) & 0.579 (0.007) & 0.423 (0.008) & 0.551 (0.004) & 0.640 (0.005) \\  
RFS-Sum ($\delta$ = 0.05) & 0.804 (0.003) & 0.516 (0.006) & 0.562 (0.008) & 0.392 (0.008) & 0.543 (0.004) & 0.621 (0.005) \\
TSLA ($\delta$ = 0.10) & 0.849 (0.002) & 0.642 (0.004) & 0.582 (0.006) & 0.393 (0.007) & 0.642 (0.003) & 0.707 (0.004)  \\
RFS-Sum ($\delta$ = 0.10) & 0.833 (0.002) & 0.585 (0.004) & 0.516 (0.007) & 0.315 (0.006) & 0.614 (0.004) & 0.657 (0.005)  \\
TSLA ($\delta$ = 0.15) & 0.850 (0.002) & 0.680 (0.005) & 0.549 (0.006) & 0.376 (0.006) & 0.676 (0.003) & 0.740 (0.003) \\
RFS-Sum ($\delta$ = 0.15) & 0.818 (0.002) & 0.599 (0.004) & 0.442 (0.006) & 0.265 (0.005) & 0.626 (0.003) & 0.666 (0.005) \\
\bottomrule
\end{tabular}
\end{table}

\begin{table}[H]
\centering
\caption{Simulation: paired $t$-test on model performance metrics between TSLA and RFS-Sum for different feature rarity levels in classification settings.}
\label{tab:tab13}
\footnotesize
\begin{tabular}{l|ccccccc}
\toprule
               &            &            & \multicolumn{2}{c}{Sensitivity} & \multicolumn{2}{c}{PPV}         \\
         & AUC        & AUPRC      & 90\% specificity & 95\% specificity & 90\% specificity & 95\% specificity \\
\midrule
$\delta$ = 0.05 &  
\begin{tabular}[c]{@{}c@{}}0.005 (0.002)\\ (0.0257)\end{tabular} & \begin{tabular}[c]{@{}c@{}}0.023 (0.003)\\ (\textless 0.0001)\end{tabular} & \begin{tabular}[c]{@{}c@{}}0.017 (0.004)\\ (0.0001)\end{tabular} & \begin{tabular}[c]{@{}c@{}}0.030 (0.005)\\ (\textless 0.0001)\end{tabular} & \begin{tabular}[c]{@{}c@{}}0.008 (0.002)\\ (0.0002)\end{tabular} & \begin{tabular}[c]{@{}c@{}}0.019 (0.003)\\ (\textless 0.0001)\end{tabular}   \\
$\delta$ = 0.10 &   
\begin{tabular}[c]{@{}c@{}}0.016 (0.002)\\ (\textless 0.0001)\end{tabular} & \begin{tabular}[c]{@{}c@{}}0.057 (0.003)\\ (\textless 0.0001)\end{tabular} & \begin{tabular}[c]{@{}c@{}}0.065 (0.005)\\ (\textless 0.0001)\end{tabular} & \begin{tabular}[c]{@{}c@{}}0.078 (0.006)\\ (\textless 0.0001)\end{tabular} & \begin{tabular}[c]{@{}c@{}}0.029 (0.002)\\ (\textless 0.0001)\end{tabular} & \begin{tabular}[c]{@{}c@{}}0.049 (0.004)\\ (\textless 0.0001)\end{tabular}  \\
$\delta$ = 0.15 & 
\begin{tabular}[c]{@{}c@{}}0.032 (0.002)\\ (\textless 0.0001)\end{tabular} & \begin{tabular}[c]{@{}c@{}}0.081 (0.003)\\ (\textless 0.0001)\end{tabular} & \begin{tabular}[c]{@{}c@{}}0.107 (0.005)\\ (\textless 0.0001)\end{tabular} & \begin{tabular}[c]{@{}c@{}}0.111 (0.006)\\ (\textless 0.0001)\end{tabular} & \begin{tabular}[c]{@{}c@{}}0.050 (0.002)\\ (\textless 0.0001)\end{tabular} & \begin{tabular}[c]{@{}c@{}}0.074 (0.004)\\ (\textless 0.0001)\end{tabular}  \\
\bottomrule
\end{tabular}
\end{table}

\subsection{An Additional High-Dimensional Example}
\label{sec::app-simu-highdim}

We conduct an additional simulation study under the high-dimensional scenario, where $p_0$ is 199. The tree structure is shown in Figure~\ref{fig:fig15}. We take 
\begin{align*}
	\eta\ &=\ -1\ +2\ x_1\ +\ 3\ (x_2\ \lor\ x_3\ \lor\ x_4)\ -\ 1.5\ (x_{11}\ \lor\ \ x_{12}\ \lor\ \cdots\ \lor\ x_{19})\\
	\ &-\ (x_{38}\ \lor\ x_{39}\ \lor\ \cdots\ \lor\ x_{64}), 
\end{align*}
with the original design matrix $\X_0\in \{0, 1\}^{n\times p_0}$ generated with independent $\mbox{Bernoulli}(0.1)$ entries. 
{\cre Based on the generated design matrix $\X_0$, under the regression settings, we generate $n$ independent samples of $y$ from $y=\eta+\varepsilon$, in which each $\varepsilon$ is independently sampled from $N(0, \sigma^2)$ with $\mbox{SNR} =0.5$.  
Under the classification settings, we obtain $n$ independent samples of $y$ from $\mbox{Bernoulli}(q)$, where $q=\exp(z+\eta)/\{1+\exp(z+\eta)\}$, and each $z$ is independently sampled from $N(0, 0.25)$.} The average value of $q$ is around 0.22. We use $n=100$ observations for training and the results are measured on 300/1100 independent samples for regression/classification cases. 

\begin{figure}[tpb]
	\centering
	\begin{tikzpicture}[grow = right, level distance = 60pt]
	\tiny
	\tikzstyle{level 1}=[sibling distance=72mm]
         \tikzstyle{level 2}=[sibling distance=24mm]
         \tikzstyle{level 3}=[sibling distance=8.7mm]
         \tikzstyle{level 4}=[sibling distance=2.9mm]
         \node {$u_1^0$} [align=center]
         	child {node {$u_1^1$}
			child {node {$u_1^2$ \\ $x_1$}}
			child {node {$u_2^2$}
				child {node {$u_1^3$}
					child {node {$u_1^4$\ \ $x_2$}}
					child {node {$u_2^4$\ \ $x_3$}}
					child {node {$u_3^4$\ \ $x_4$}}}
				child {node {$u_2^3$}
					child {node {$u_4^4\ \ x_5$}}
					child {node {$u_5^4\ \  x_6$}}
					child {node {$u_6^4\ \ x_7$}}}
				child {node {$u_3^3$}
					child {node {$u_7^4\ \ x_8$}}
					child {node {$u_8^4\ \ x_9$}}
					child {node {$u_9^4\ \ x_{10}$}}}}
			child {node {$u_3^2$}
				child {node {$u_4^3$}
					child {node {$u_{10}^4-u_{12}^4$}
						child {node {$u_1^5-u_9^5$ \\ $x_{11}-x_{19}$}}}}
				child {node {$u_5^3$}
					child {node {$u_{13}^4-u_{15}^4$}
						child {node {$u_{10}^5-u_{18}^5$ \\ $x_{20}-x_{28}$}}}}
				child {node {$u_6^3$}
					child {node {$u_{16}^4-u_{18}^4$}
						child {node {$u_{19}^5-u_{27}^5$ \\ $x_{29}-x_{37}$}}}}}}
		child {node {$u_2^1$}
			child {node {$u_4^2$}
				child {node {$u_7^3$}
					child {node {$u_{19}^4-u_{21}^4$}
						child {node {$u_{28}^5-u_{36}^5$ \\ $x_{38}-x_{46}$}}}}
				child {node {$u_8^3$}
					child {node {$u_{22}^4-u_{24}^4$}
						child {node {$u_{37}^5-u_{45}^5$ \\ $x_{47}-x_{55}$}}}}
				child {node {$u_9^3$}
					child {node {$u_{25}^4-u_{27}^4$}
						child {node {$u_{46}^5-u_{54}^5$ \\ $x_{56}-x_{64}$}}}}}
			child {node {$u_5^2$}
				child {node {$u_{10}^3$}
					child {node {$u_{28}^4-u_{30}^4$}
						child {node {$u_{55}^5-u_{63}^5$ \\ $x_{65}-x_{73}$}}}}
				child {node {$u_{11}^3$}
					child {node {$u_{31}^4-u_{33}^4$}
						child {node {$u_{64}^5-u_{72}^5$ \\ $x_{74}-x_{82}$}}}}
				child {node {$u_{12}^3$}
					child {node {$u_{34}^4-u_{36}^4$}
						child {node {$u_{73}^5-u_{81}^5$ \\ $x_{83}-x_{91}$}}}}}
			child {node {$u_6^2$}
				child {node {$u_{13}^3$}
					child {node {$u_{37}^4-u_{39}^4$}
						child {node {$u_{82}^5-u_{90}^5$ \\ $x_{92}-x_{100}$}}}}
				child {node {$u_{14}^3$}
					child {node {$u_{40}^4-u_{42}^4$}
						child {node {$u_{91}^5-u_{99}^5$ \\ $x_{101}-x_{109}$}}}}
				child {node {$u_{15}^3$}
					child {node {$u_{43}^4-u_{45}^4$}
						child {node {$u_{100}^5-u_{108}^5$ \\ $x_{110}-x_{118}$}}}}}}
		child {node {$u_3^1$}
			child {node {$u_7^2$}
				child {node {$u_{16}^3$}
					child {node {$u_{46}^4-u_{48}^4$}
						child {node {$u_{109}^5-u_{117}^5$ \\ $x_{119}-x_{127}$}}}}
				child {node {$u_{17}^3$}
					child {node {$u_{49}^4-u_{51}^4$}
						child {node {$u_{118}^5-u_{126}^5$ \\ $x_{128}-x_{136}$}}}}
				child {node {$u_{18}^3$}
					child {node {$u_{52}^4-u_{54}^4$}
						child {node {$u_{127}^5-u_{135}^5$ \\ $x_{137}-x_{145}$}}}}}
			child {node {$u_8^2$}
				child {node {$u_{19}^3$}
					child {node {$u_{55}^4-u_{57}^4$}
						child {node {$u_{136}^5-u_{144}^5$ \\ $x_{146}-x_{154}$}}}}
				child {node {$u_{20}^3$}
					child {node {$u_{58}^4-u_{60}^4$}
						child {node {$u_{145}^5-u_{153}^5$ \\ $x_{155}-x_{163}$}}}}
				child {node {$u_{21}^3$}
					child {node {$u_{61}^4-u_{63}^4$}
						child {node {$u_{154}^5-u_{162}^5$ \\ $x_{164}-x_{172}$}}}}}
			child {node {$u_9^2$}
				child {node {$u_{22}^3$}
					child {node {$u_{64}^4-u_{66}^4$}
						child {node {$u_{163}^5-u_{171}^5$ \\ $x_{173}-x_{181}$}}}}
				child {node {$u_{23}^3$}
					child {node {$u_{67}^4-u_{69}^4$}
						child {node {$u_{172}^5-u_{180}^5$ \\ $x_{182}-x_{190}$}}}}
				child {node {$u_{24}^3$}
					child {node {$u_{70}^4-u_{72}^4$}
						child {node {$u_{181}^5-u_{190}^5$ \\ $x_{191}-x_{199}$}}}}}};

	\end{tikzpicture}
	\caption{Tree structure for the high-dimensional example: a tree of depth 5 with 199 leaf nodes. At depth 4, three nodes are marked with one branch. At depth 5, nine nodes are marked with one branch, and the nine nodes are equally divided into three child node groups of their parent nodes at depth 4.}
	\label{fig:fig15}
\end{figure}

\begin{figure}[H]
  \centering
  \includegraphics[width=0.45\linewidth]{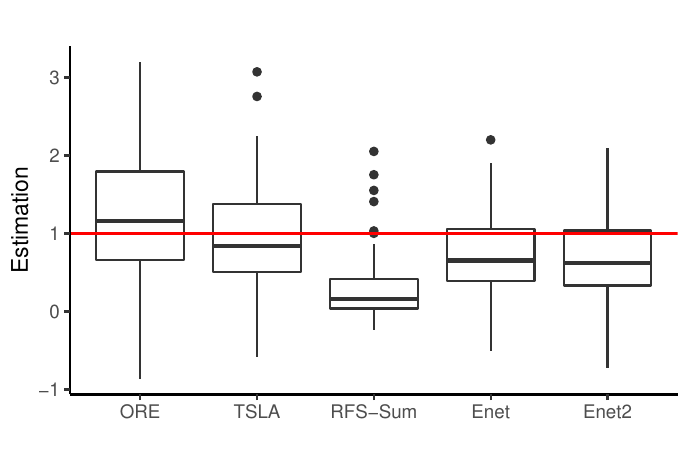}
  \caption{Simulation: boxplot for estimations of the unpenalized coefficient under high-dimensional scenario in classification settings. The horizontal line marks the true coefficient.}
  \label{fig:fig16}
\end{figure}

The detailed results are shown in Tables~\ref{tab:tab14}--\ref{tab:tab15} for regression and in Tables~\ref{tab:tab16}--\ref{tab:tab18} for classification. Figure~\ref{fig:fig16} reports the estimation of the regression coefficient of $z$ in classification. TSLA still provides substantially better results than the competing models. It is worth noting that the dimension of the TSLA model, $p$, is around 550, which is much larger than $p_0$. The results further suggest the stable performance of the TSLA model under the high-dimensional scenario. 

\begin{table}[H]
\footnotesize
\centering
\caption{Simulation: prediction and feature aggregation performance under high-dimensional scenario in regression setting.
}
\label{tab:tab14}
\begin{tabular}{ccccc|cc}
\toprule
      \multicolumn{5}{c|}{MSE}  &  \multicolumn{2}{c}{FNR/FPR}  \\
       ORE    &   TSLA & RFS-Sum &  Enet & Enet2  & TSLA & RFS-Sum\\
\midrule   
\begin{tabular}[c]{@{}c@{}}5.748 \\ (0.057) \end{tabular}  &  \begin{tabular}[c]{@{}c@{}}6.928 \\ (0.092)\end{tabular}& \begin{tabular}[c]{@{}c@{}}7.544 \\ (0.101) \end{tabular}  &  \begin{tabular}[c]{@{}c@{}}7.819 \\ (0.101) \end{tabular}&  \begin{tabular}[c]{@{}c@{}}7.958 \\ (0.092) \end{tabular}  & \begin{tabular}[c]{@{}c@{}}0.250 / 0.178 \\ (0.017) / (0.009) \end{tabular}  & \begin{tabular}[c]{@{}c@{}}0.288 / 0.358 \\ (0.015) / (0.007) \end{tabular} \\
\bottomrule 
\end{tabular}
\end{table}

\begin{table}[H]
\footnotesize
\centering
\caption{Simulation: paired $t$-test on model performance metrics under high-dimensional scenario in regression setting.}
\label{tab:tab15}
\begin{tabular}{ccc|cc}
\toprule
      \multicolumn{3}{c|}{MSE}  &  FNR & FPR  \\
        RFS-Sum &  Enet & Enet2  &  \multicolumn{2}{c}{RFS-Sum}\\
\midrule   
\begin{tabular}[c]{@{}c@{}}-0.616 (0.055)\\ (\textless 0.0001) \end{tabular}  &  \begin{tabular}[c]{@{}c@{}}-0.891 (0.068)\\ (\textless 0.0001) \end{tabular}& \begin{tabular}[c]{@{}c@{}}-1.030 (0.054)\\ (\textless 0.0001) \end{tabular}  &  \begin{tabular}[c]{@{}c@{}}-0.037 (0.023)\\ (0.1040)  \end{tabular}  & \begin{tabular}[c]{@{}c@{}}-0.180 (0.009) \\ (\textless 0.0001) \end{tabular} \\
\bottomrule 
\end{tabular}
\end{table}

\begin{table}[H]
\centering
\caption{Simulation: prediction performance under high-dimensional scenario in classification setting.} 
\label{tab:tab16}
\footnotesize
\begin{tabular}{l|ccccccc}
\toprule
               &            &            & \multicolumn{2}{c}{Sensitivity} & \multicolumn{2}{c}{PPV}         \\
Model          & AUC        & AUPRC      & 90\% specificity & 95\% specificity & 90\% specificity & 95\% specificity \\
\midrule
ORE& 0.848 (0.003) & 0.655 (0.005) & 0.590 (0.005) & 0.435 (0.009) & 0.634 (0.003) & 0.711 (0.008)  \\
TSLA& 0.756 (0.006) & 0.499 (0.008) & 0.414 (0.010) & 0.270 (0.008) & 0.540 (0.007) & 0.602 (0.008) \\
RFS-Sum& 0.744 (0.006) & 0.472 (0.007) & 0.399 (0.010) & 0.240 (0.007) & 0.530 (0.008) & 0.573 (0.009) \\
Enet& 0.676 (0.007) & 0.398 (0.009) & 0.311 (0.012) & 0.199 (0.012) & 0.453 (0.009) & 0.499 (0.011) \\
Enet2& 0.623 (0.006) & 0.336 (0.008) & 0.275 (0.018) & 0.193 (0.021) & 0.381 (0.009) & 0.415 (0.012)\\
\bottomrule
\end{tabular}
\end{table}

\begin{table}[H]
\centering
\caption{Simulation: paired $t$-test on model performance metrics under the high-dimensional scenario in classification setting.} 
\label{tab:tab17}
\footnotesize
\begin{tabular}{l|ccccccc}
\toprule
               &            &            & \multicolumn{2}{c}{Sensitivity} & \multicolumn{2}{c}{PPV}         \\
Model          & AUC        & AUPRC      & 90\% specificity & 95\% specificity & 90\% specificity & 95\% specificity \\
\midrule
RFS-Sum &   
\begin{tabular}[c]{@{}c@{}}0.012 (0.005)\\ (0.0101)\end{tabular} & \begin{tabular}[c]{@{}c@{}}0.027 (0.006)\\ (\textless 0.0001)\end{tabular} & \begin{tabular}[c]{@{}c@{}}0.015 (0.008)\\ (0.0588)\end{tabular} & \begin{tabular}[c]{@{}c@{}}0.030 (0.006)\\ (\textless 0.0001)\end{tabular} & \begin{tabular}[c]{@{}c@{}}0.010 (0.005)\\ (0.0782)\end{tabular} & \begin{tabular}[c]{@{}c@{}}0.029 (0.007)\\ (\textless 0.0001)\end{tabular} \\
Enet & 
\begin{tabular}[c]{@{}c@{}}0.080 (0.006)\\ (\textless 0.0001)\end{tabular} & \begin{tabular}[c]{@{}c@{}}0.101 (0.007)\\ (\textless 0.0001)\end{tabular} & \begin{tabular}[c]{@{}c@{}}0.102 (0.012)\\ (\textless 0.0001)\end{tabular} & \begin{tabular}[c]{@{}c@{}}0.070 (0.012)\\ (\textless 0.0001)\end{tabular} & \begin{tabular}[c]{@{}c@{}}0.087 (0.008)\\ (\textless 0.0001)\end{tabular} & \begin{tabular}[c]{@{}c@{}}0.103 (0.010)\\ (\textless 0.0001)\end{tabular}  \\
Enet2 & 
\begin{tabular}[c]{@{}c@{}}0.133 (0.007)\\ (\textless 0.0001)\end{tabular} & \begin{tabular}[c]{@{}c@{}}0.164 (0.009)\\ (\textless 0.0001)\end{tabular} & \begin{tabular}[c]{@{}c@{}}0.139 (0.020)\\ (\textless 0.0001)\end{tabular} & \begin{tabular}[c]{@{}c@{}}0.077 (0.023)\\ (0.0010)\end{tabular} & \begin{tabular}[c]{@{}c@{}}0.159 (0.010)\\ (\textless 0.0001)\end{tabular} & \begin{tabular}[c]{@{}c@{}}0.187 (0.012)\\ (\textless 0.0001)\end{tabular} \\
\bottomrule
\end{tabular}
\end{table}

\begin{table}[H]
\centering
\footnotesize
\caption{Simulation: feature aggregation performance under high-dimensional scenario in classification setting.} \label{tab:tab18}
\begin{tabular}{cc|cc}
\toprule
 \multicolumn{2}{c|}{FNR/FPR} & \multicolumn{2}{c}{Paired $t$-test}           \\
 TSLA                  & RFS-Sum                    & mean difference & $p$-value                   \\
 \midrule
 \begin{tabular}[c]{@{}c@{}}0.398 / 0.116\\ (0.022) / (0.009)\end{tabular}  &
 \begin{tabular}[c]{@{}c@{}}0.493 / 0.257\\ (0.019) / (0.011)\end{tabular}  &
 \begin{tabular}[c]{@{}c@{}}-0.095 / -0.141\\ (0.024) / (0.013)\end{tabular} & 
 0.0002 / \textless 0.0001 \\
\bottomrule
\end{tabular}
\end{table}

\clearpage
\section{Supplemental Materials for Suicide Risk Modeling}
\label{sec::app-suicide}

This section complements the results reported in Section~5 of the main paper. 

\subsection{Data}\label{sec::app-suicide-data}

\begin{figure}[H]
  \centering
  \includegraphics[width=\linewidth]{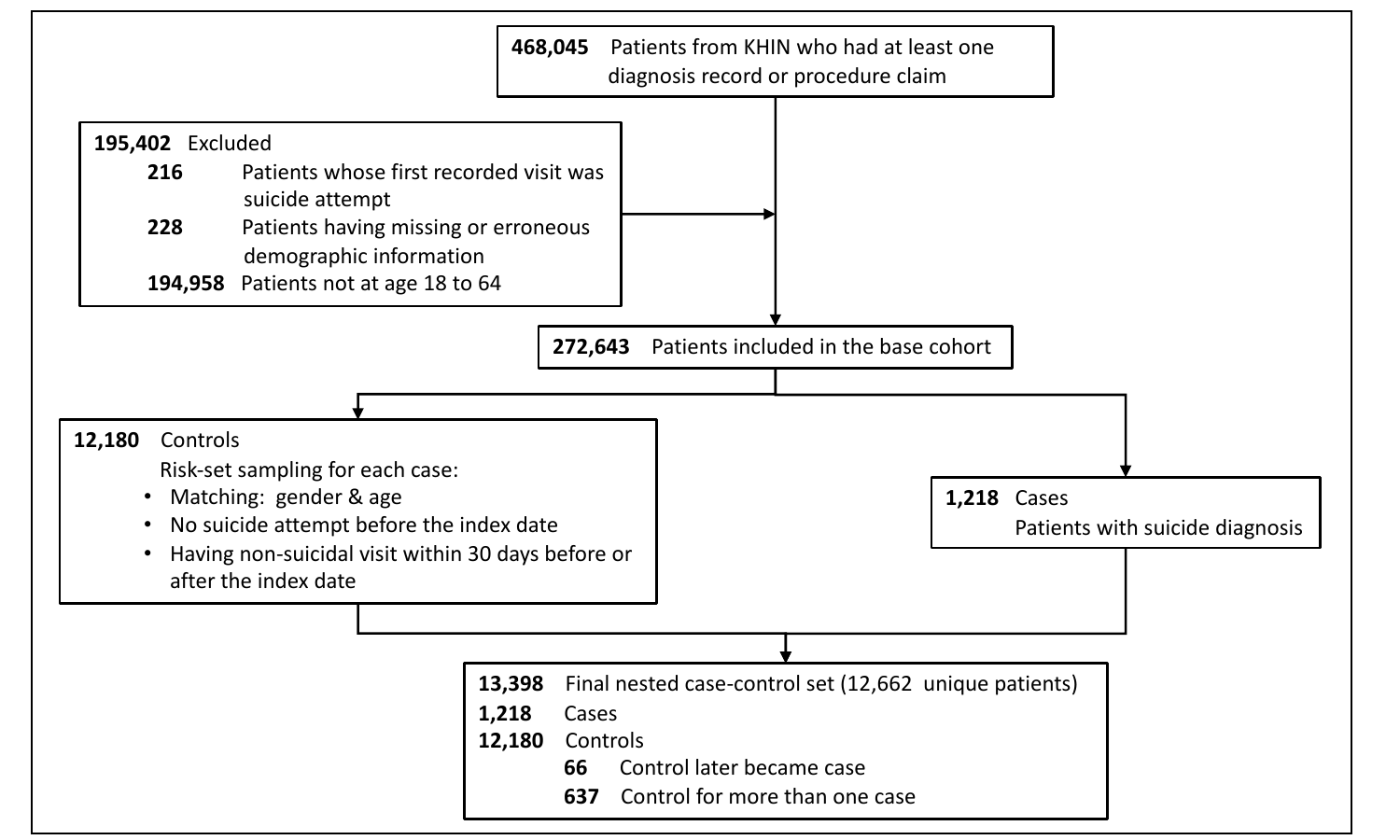}
  \caption{Suicide risk study: flowchart of the nested case-control design.}
  \label{fig:fig17}
\end{figure}

The flowchart of the study design used in the suicide risk modeling is shown in Figure~\ref{fig:fig17}. 

\begin{figure}[htp]
  \centering
  \includegraphics[width=\linewidth, height=0.4\linewidth]{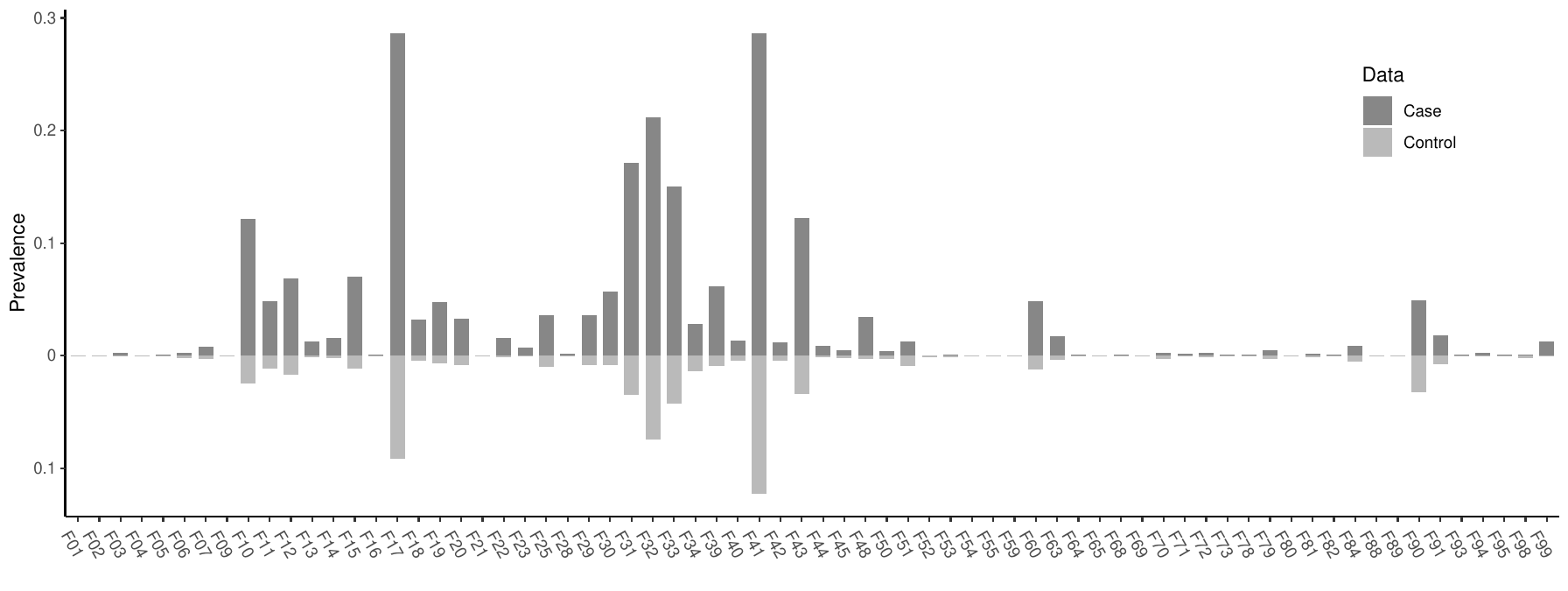}
  \caption{Suicide risk study: prevalence of three-digit ICD codes in the F-chapter. There are 70 codes in total.}
  \label{fig:fig18}
\end{figure}

Figures~\ref{fig:fig18} shows the prevalence of the three-digit F-chapter codes in cases and controls. 

\subsection{Additional Modeling Results}\label{sec::app-suicide-result}

We report the paired $t$-test results in Table~\ref{tab:tab19} to compare the prediction performance between TSLA and the other methods. The $t$-test should be interpreted with some caution, as the results from the 10 splits are not independent. 
\begin{table}[H]
\footnotesize
\centering
\caption{Suicide risk study: paired $t$-test on model performance metrics between TSLA and other methods over the 10 splits. Reported are the means and standard errors (in parentheses) of the paired differences, and the $p$-values (in parentheses).}
\label{tab:tab19}
\begin{tabular}{l|cccccc}
\toprule
        &         &    & \multicolumn{2}{c}{Sensitivity} & \multicolumn{2}{c}{PPV}  \\
        & AUC    & AUPRC  & 10\% positive & 5\% positive  & 10\% positive & 5\% positive  \\
\midrule      
TSLA-screen &
\begin{tabular}[c]{@{}c@{}}0.000 (0.001)\\ (0.8120)\end{tabular} & \begin{tabular}[c]{@{}c@{}}0.001 (0.003)\\ (0.7794)\end{tabular} & \begin{tabular}[c]{@{}c@{}}0.002 (0.005)\\ (0.7491)\end{tabular} & \begin{tabular}[c]{@{}c@{}}0.002 (0.005)\\ (0.6434)\end{tabular} & \begin{tabular}[c]{@{}c@{}}0.001 (0.005)\\ (0.7509)\end{tabular} & \begin{tabular}[c]{@{}c@{}}0.004 (0.010)\\ (0.7061)\end{tabular} \\
RFS-Sum &
\begin{tabular}[c]{@{}c@{}}0.002 (0.001)\\ (0.1919)\end{tabular} & \begin{tabular}[c]{@{}c@{}}0.004 (0.003)\\ (0.2814)\end{tabular} & \begin{tabular}[c]{@{}c@{}}0.007 (0.006)\\ (0.2694)\end{tabular} & \begin{tabular}[c]{@{}c@{}}0.009 (0.006)\\ (0.1529)\end{tabular} & \begin{tabular}[c]{@{}c@{}}0.006 (0.005)\\ (0.2695)\end{tabular} & \begin{tabular}[c]{@{}c@{}}0.016 (0.011)\\ (0.1538)\end{tabular} \\
Enet    & 
 \begin{tabular}[c]{@{}c@{}}0.004 (0.002)\\ (0.1034)\end{tabular} & \begin{tabular}[c]{@{}c@{}}0.009 (0.004)\\ (0.0330)\end{tabular} & \begin{tabular}[c]{@{}c@{}}0.009 (0.010)\\ (0.3802)\end{tabular} & \begin{tabular}[c]{@{}c@{}}0.012 (0.005)\\ (0.0383)\end{tabular} & \begin{tabular}[c]{@{}c@{}}0.008 (0.009)\\ (0.3820)\end{tabular} & \begin{tabular}[c]{@{}c@{}}0.022 (0.009)\\ (0.0384)\end{tabular}  \\
Enet2   & 
\begin{tabular}[c]{@{}c@{}}0.010 (0.003)\\ (0.0036)\end{tabular} & \begin{tabular}[c]{@{}c@{}}0.010 (0.003)\\ (0.0075)\end{tabular} & \begin{tabular}[c]{@{}c@{}}0.013 (0.007)\\ (0.0825)\end{tabular} & \begin{tabular}[c]{@{}c@{}}0.010 (0.006)\\ (0.1116)\end{tabular} & \begin{tabular}[c]{@{}c@{}}0.012 (0.006)\\ (0.0876)\end{tabular} & \begin{tabular}[c]{@{}c@{}}0.018 (0.010)\\ (0.1114)\end{tabular} \\
\bottomrule
\end{tabular}
\end{table}

\begin{figure}[htp]
  \centering
  \includegraphics[width=\linewidth, height = 1.25\textwidth]{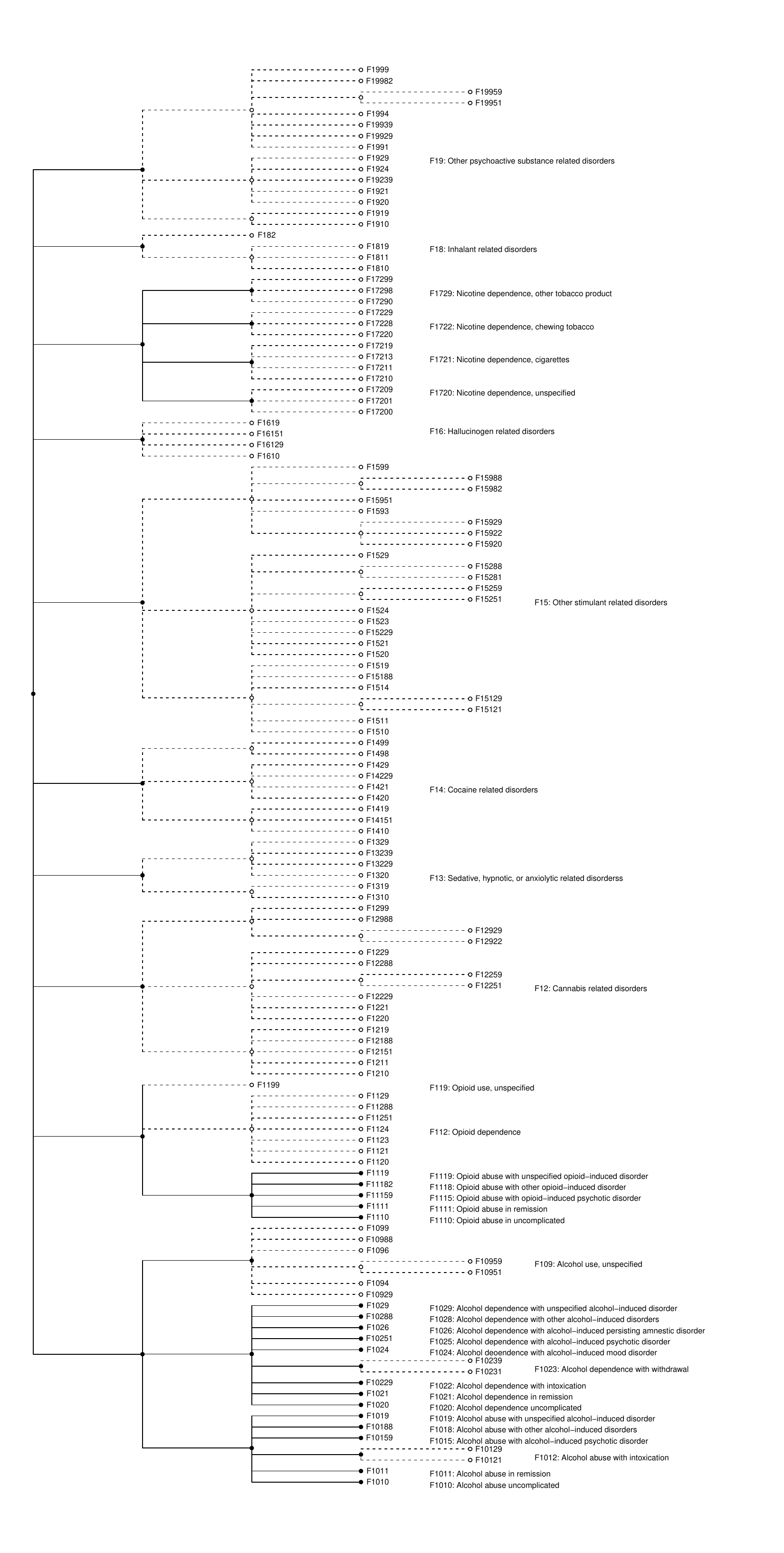}
  \caption{Suicide risk study: selection and aggregation of F1 codes (Mental and behavioral disorders due to psychoactive substance use) based on TSLA.}
  \label{fig:fig19}
\end{figure}

\begin{figure}[htp]
  \centering
  \includegraphics[width=\linewidth, height=0.4\textwidth]{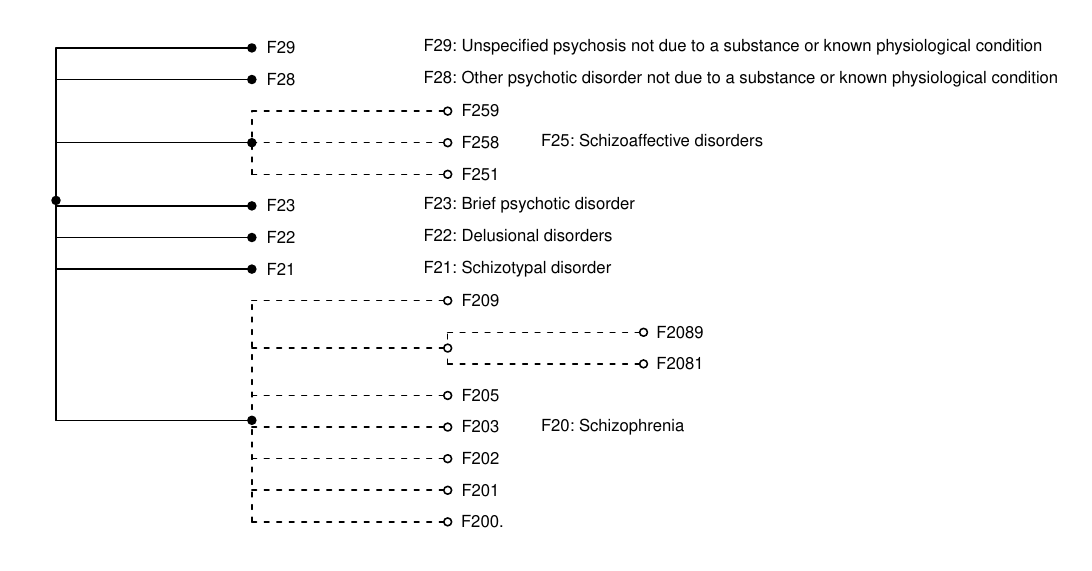}
  \caption{Suicide risk study: selection and aggregation of F2 codes (Schizophrenia, schizotypal, delusional, and other non-mood psychotic disorders) based on TSLA.}
  \label{fig:fig20}
\end{figure}

Figures \ref{fig:fig19} and \ref{fig:fig20} show the selection and aggregation patterns of the categories F1 and F2, respectively.

\clearpage
\subsection{Feature Aggregation by TSLA with the Prevalence-Based Pre-Screening}\label{sec::app-suicide-result-screen}

\begin{figure}[H]
  \centering
  \includegraphics[width=\linewidth, height=0.85\textwidth]{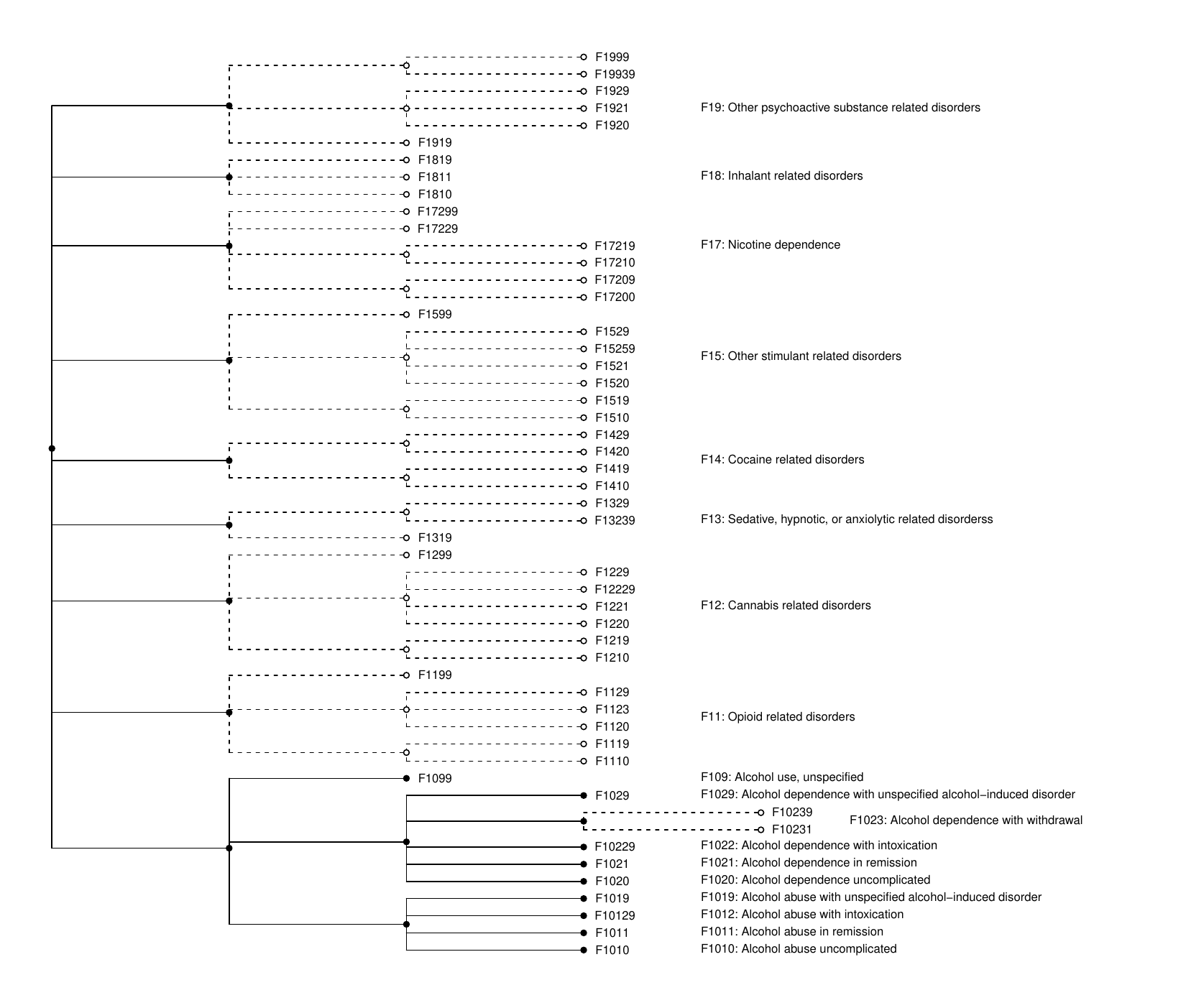}
  \caption{Suicide risk study: selection and aggregation of F1 codes (Mental and behavioral disorders due to psychoactive substance use) based on TSLA-screen.}
  \label{fig:fig21}
\end{figure}

\begin{figure}[H]
  \centering
  \includegraphics[width=\linewidth, height=0.3\textwidth]{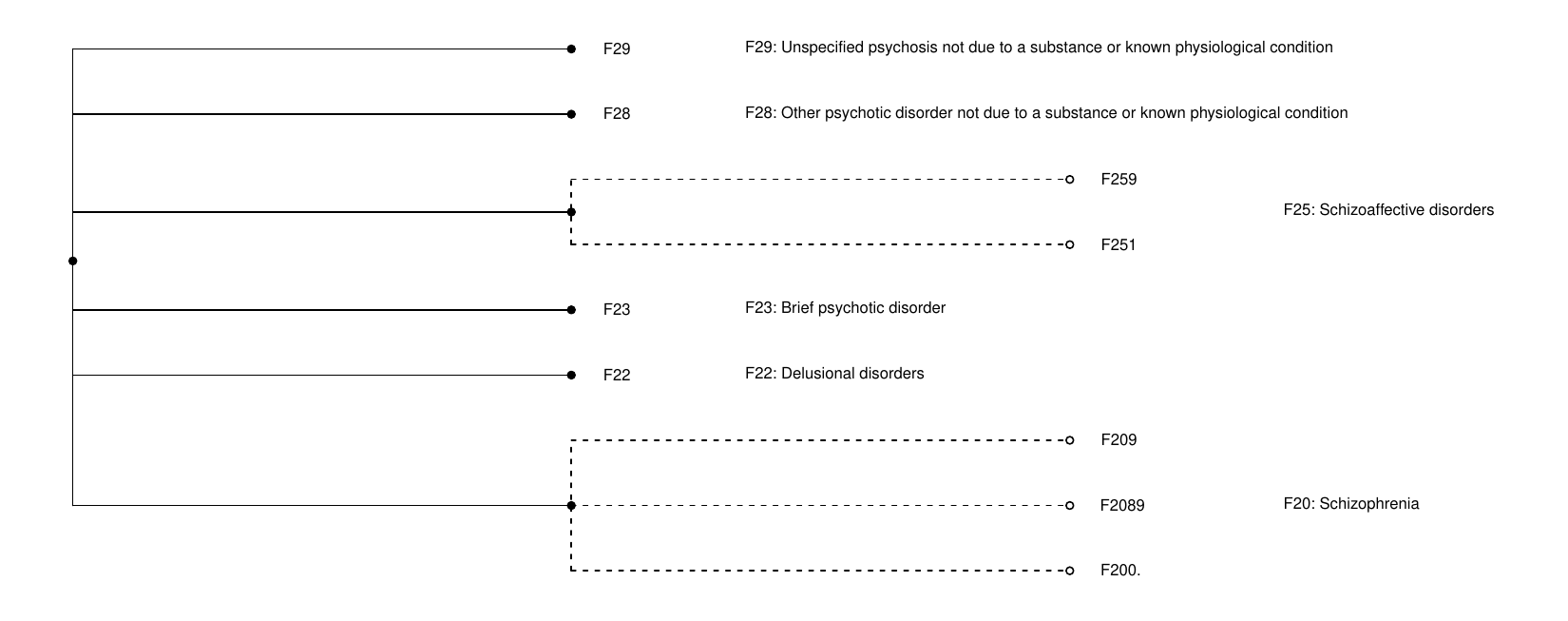}
  \caption{Suicide risk study: selection and aggregation of F2 codes (Schizophrenia, schizotypal, delusional, and other non-mood psychotic disorders) based on TSLA-screen.}
  \label{fig:fig22}
\end{figure}

\begin{figure}[H]
  \centering
  \includegraphics[width=\linewidth, height=0.8\textwidth]{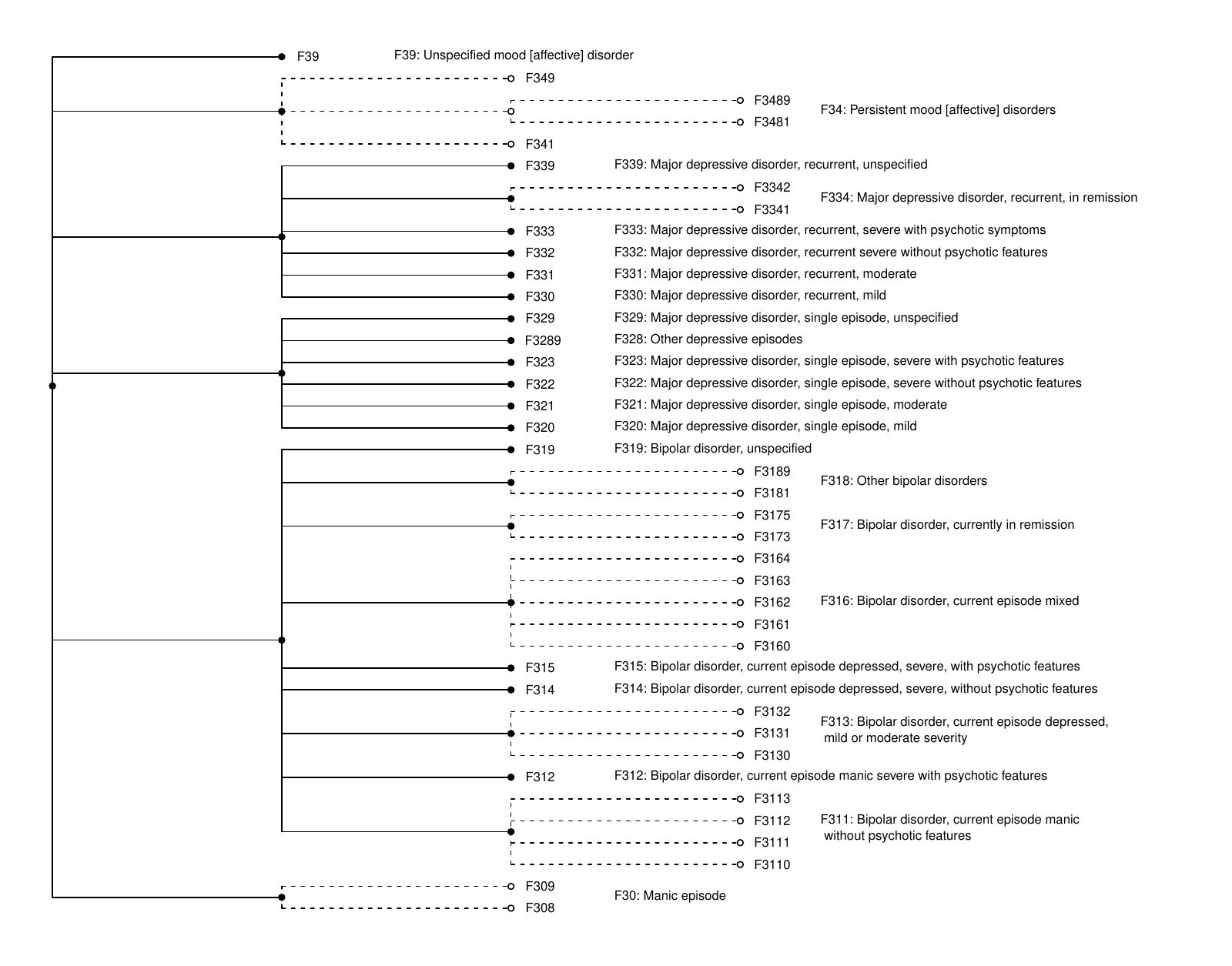}
  \caption{Suicide risk study: selection and aggregation of F3 codes (Mood (affective) disorders) based on TSLA-screen.}
  \label{fig:fig23}
\end{figure}

\begin{figure}[H]
  \centering
  \includegraphics[width=\linewidth, height=0.45\textwidth]{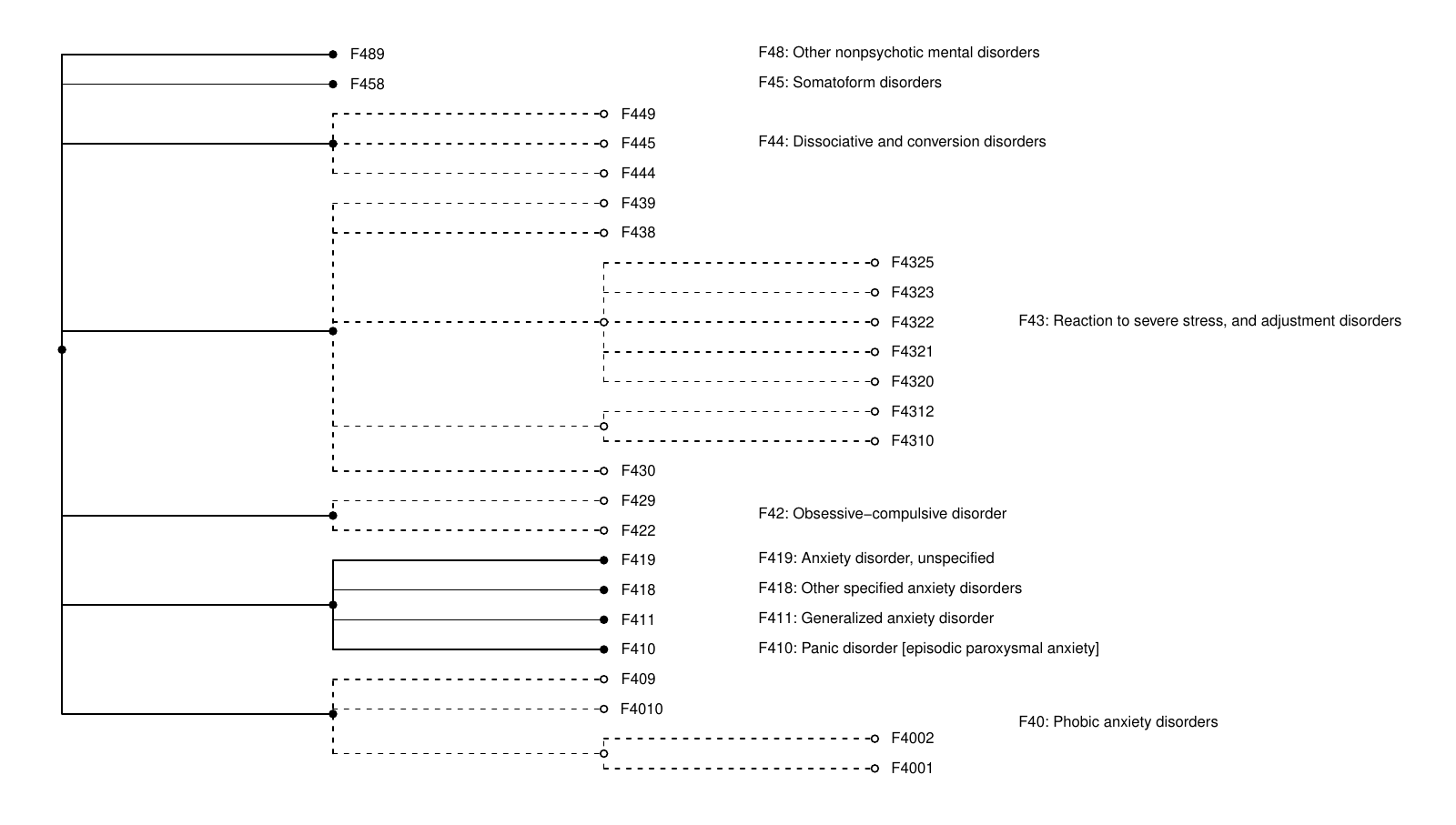}
  \caption{Suicide risk study: selection and aggregation of F4 codes (Anxiety, dissociative, stress-related, somatoform and other nonpsychotic mental disorders) based on TSLA-screen.}
  \label{fig:fig24}
\end{figure}

\begin{figure}[H]
  \centering
  \includegraphics[width=\linewidth, height=0.6\textwidth]{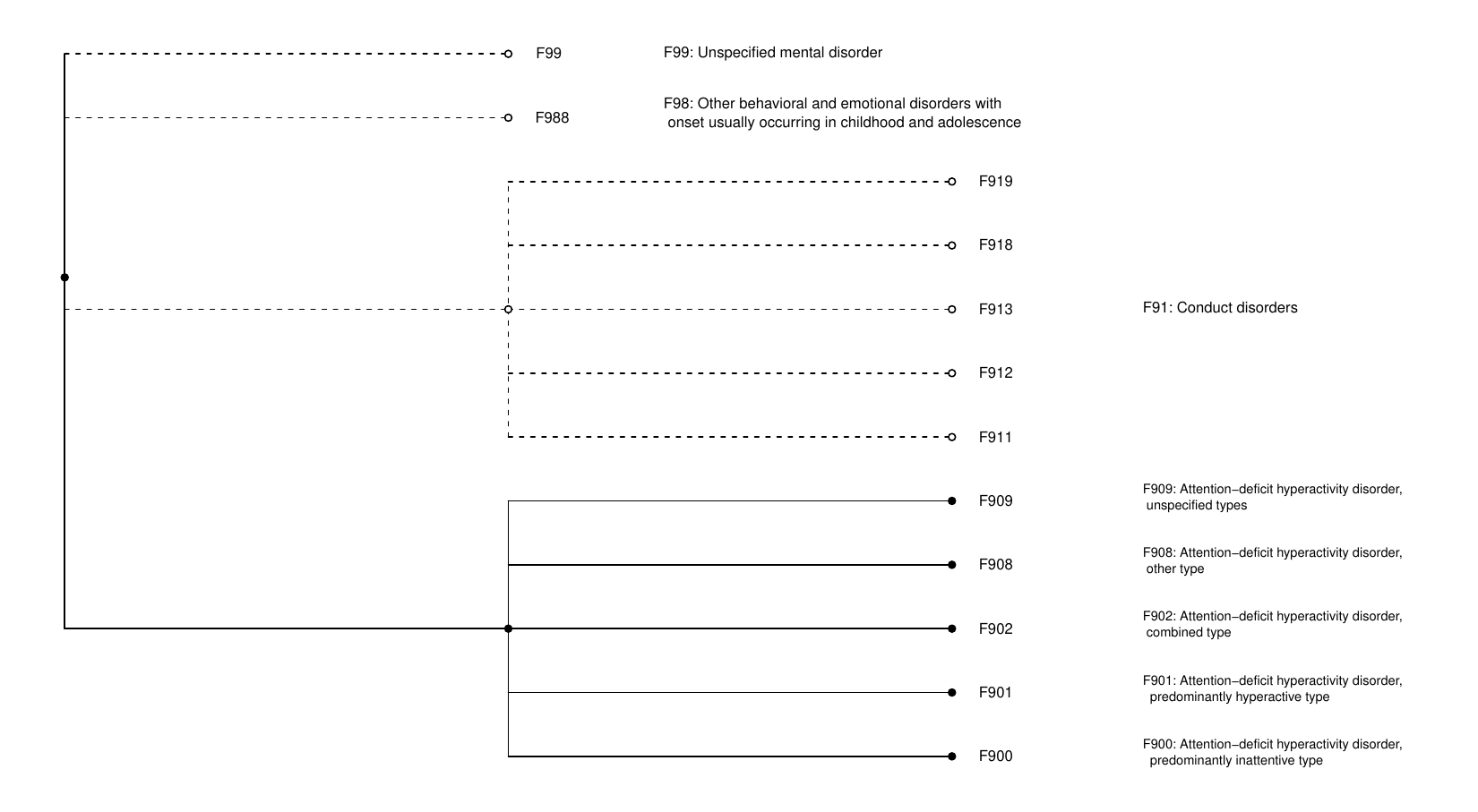}
  \caption{Suicide risk study: selection and aggregation of F9 codes (Behavioral and emotional disorders with onset usually occurring in childhood and adolescence \& unspecified mental disorder) based on TSLA-screen.}
  \label{fig:fig25}
\end{figure}

The aggregation patterns produced by TSLA with the prevalence-based pre-screening (TSLA-screen) are showed in Figures \ref{fig:fig21}--\ref{fig:fig25}.
  
 \newpage
 \bibliographystyle{chicago}

\end{document}